\documentclass{article}

\usepackage{microtype}
\usepackage{graphicx}
\usepackage{booktabs} %

\usepackage{hyperref}

\usepackage[ruled,vlined]{algorithm2e}

\usepackage[accepted]{icml2024}

\usepackage{amsmath}
\usepackage{amssymb}
\usepackage{mathtools}
\usepackage{amsthm}

\usepackage[capitalize,noabbrev]{cleveref}

\usepackage{url}
\usepackage{makecell}
\usepackage{array}
\usepackage{caption}
\usepackage{subcaption}
\usepackage{wrapfig}
\usepackage{graphicx}
\usepackage{multirow}
\usepackage{xfrac}
\usepackage{enumitem}
\usepackage{scalerel}

\makeatletter
\newcommand*{\rom}[1]{\expandafter\@slowromancap\romannumeral #1@}
\makeatother

\makeatletter
\newcommand*{\inlineequation}[2][]{%
	\begingroup
	\refstepcounter{equation}%
	\ifx\\#1\\%
	\else
	\label{#1}%
	\fi
	\relpenalty=10000 %
	\binoppenalty=10000 %
	\ensuremath{%
		#2%
	}%
	~\@eqnnum
	\endgroup
}
\makeatother

\theoremstyle{plain}
\newtheorem{theorem}{Theorem}[section]

\theoremstyle{definition}

\theoremstyle{remark}

\usepackage[textsize=tiny]{todonotes}

\newcommand*{\defeq}{\stackrel{{\normalfont \text{def}}}{=}}

\newcommand{\bbE}{\mathbb{E}} 
\newcommand{\bbP}{\mathbb{P}} 
\newcommand{\bbQ}{\mathbb{Q}} 
\newcommand{\bbR}{\mathbb{R}} 
 
\newcommand{\bbW}{\mathbb{W}}
\newcommand{\bbS}{\mathbb{S}}

\newcommand{\cL}{\mathcal{L}} 
\newcommand{\cN}{\mathcal{N}} 
\newcommand{\cS}{\mathcal{S}} 
 
\newcommand{\cX}{\mathcal{X}}

\newcommand{\cY}{\mathcal{Y}}
\newcommand{\cP}{\mathcal{P}}

\newcommand{\cC}{\mathcal{C}}
\newcommand{\cR}{\mathcal{R}}
\newcommand{\cZ}{\mathcal{Z}}
\newcommand{\cG}{\mathcal{G}}
\newcommand{\cV}{\mathcal{V}}
\newcommand{\cE}{\mathcal{E}}
\newcommand{\cM}{\mathcal{M}}
\newcommand{\E}[1]{{\underset{#1}{\bbE}}}

 \newcommand{\KL}{\text{KL}}
 
\newcommand{\OT}{\text{OT}}

\allowdisplaybreaks

\newcommand{\dd}{\mathrm{d}}
\newcommand{\dv}[2]{\frac{\dd #1}{\dd #2}}

\newcolumntype{x}[1]{>{\centering\arraybackslash\hspace{0pt}}p{#1}}

\DeclareMathOperator*{\arginf}{arg\,inf}

\usepackage[normalem]{ulem}

\everypar{\looseness=-1}

\definecolor{mygreen}{RGB}{0, 160, 0}

\definecolor{myyellow}{RGB}{255, 200, 0}

\definecolor{myfiol}{RGB}{100, 41, 204}

\begin{document}

\twocolumn[
\icmltitle{Estimating Barycenters of Distributions with Neural Optimal Transport}

\icmlsetsymbol{equal}{*}

\begin{icmlauthorlist}
\icmlauthor{Alexander Kolesov$\,^*$}{skoltech}
\icmlauthor{Petr Mokrov$\,^*$}{skoltech}
\icmlauthor{Igor Udovichenko}{skoltech}
\icmlauthor{Milena Gazdieva}{skoltech}
\icmlauthor{Gudmund Pammer}{eth}
\icmlauthor{Evgeny Burnaev}{skoltech,airi}
\icmlauthor{Alexander Korotin}{skoltech,airi}
\end{icmlauthorlist}

\icmlaffiliation{skoltech}{Skolkovo Institute of Science and Technology, Moscow, Russia}
\icmlaffiliation{eth}{Department of Mathematics, ETH Z\"urich, Z\"urich, Switzerland}
\icmlaffiliation{airi}{Artificial Intelligence Research Institute, Moscow, Russia}

\icmlcorrespondingauthor{Alexander Kolesov}{a.kolesov@skoltech.ru}
\icmlcorrespondingauthor{Petr Mokrov}{petr.mokrov@skoltech.ru}

\icmlkeywords{Machine Learning, ICML}

\vskip 0.3in
]

\printAffiliationsAndNotice{\icmlEqualContribution} %

\begin{abstract}
Given a collection of probability measures, a practitioner sometimes needs to find an ``average'' distribution which adequately aggregates 
reference distributions. A theoretically appealing notion of such an average is the Wasserstein barycenter, which is the primal focus of our work. By building upon the dual formulation of Optimal Transport (OT), we propose a new scalable approach for solving the Wasserstein barycenter problem. Our methodology is based on the recent Neural OT solver: it has bi-level adversarial learning objective and works for general cost functions. These are key advantages of our method since the typical adversarial algorithms leveraging barycenter tasks utilize tri-level optimization and focus mostly on quadratic cost. 
We also establish theoretical error bounds for our proposed approach and showcase its applicability and effectiveness in illustrative scenarios and image data setups. Our source code is available at \url{https://github.com/justkolesov/NOTBarycenters}.
\end{abstract}

\vspace{-7 mm}

\section{Introduction}\label{sec:intro}

Generative Modeling encompasses a set of tools designed for manipulating probability distributions. Among them, a special place is occupied by the Wasserstein barycenter problem. This problem consists in finding a distribution which minimizes the average of specific divergences to some pre-defined distributions. In the context of Wasserstein barycenters, the divergences are chosen to be Optimal Transport (OT) costs due to attractive geometrical properties and clear intuition behind them. 
Starting with the seminal work \citep{agueh2011barycenters}, the Wasserstein barycenter problem has been consistently gaining significant attention in the research community. This is partially due to the compelling mathematical theory and an extensive list of applications: Bayesian inference \citep{srivastava2018scalable}, Geometric modelling \citep{solomon2015convolutional}, Style Transfer \citep{mroueh2020wasserstein}, Texture mixing \citep{lacombe2023learning}, Reinforcement Learning \cite{likmeta2023wasserstein}, Federated Learning \cite{singh2020model}, etc.
Reflecting the practitioners' interest, the computational OT literature is also inspired by the Wasserstein barycenter problem and proposes several methods for solving this task. 
Early works, e.g., \citep{cuturi2014fast, anderes2016discrete}, deal with a discrete learning setup, i.e., they assume that the distributions of interest are discrete.
Unfortunately, the resulting solvers mostly lack certain beneficial properties a practitioner may require, see \S\ref{subsec:learningsetup}. Our work is in line with the alternative continuous learning setup. 
To tackle this challenge, existing barycenter solvers for the continuous learning setup typically
\begin{itemize}[leftmargin=*]

\vspace{-3 mm}

\item[\textbf{a)}] resort to complex algorithmic procedures, e.g.: tri-level adversarial learning objectives \citep{pmlr-v139-fan21d, korotin2022wasserstein}, Langevin sampling \citep{kolesov2023energy}.
\item[\textbf{b)}] consider only specific formulations of the Wasserstein barycenter problem to make it more tractable, e.g.: deal with quadratic cost \citep{noble2023treebased}, utilize entropic/quadratic regularization  \citep{li2020continuous}. %
\end{itemize}

\vspace{-3 mm}
\textbf{Contribution.} We take a step forward and propose a pioneering approach for solving the Wasserstein barycenter problem which \textbf{a)} permits conventional bi-level ($\max$-$\min$) adversarial objective and \textbf{b)} can be adapted to \textit{various} formulations of the barycenter problem, e.g., with general cost functions, w/wo regularizations. In particular: 
\begin{itemize}[leftmargin=*]
\vspace{-2 mm}
\item[1.] We combine recent Neural OT method \citep{korotin2023neural} with the congruence condition (\S\ref{subsec:method_derivation}) and derive a novel practical barycenter algorithm (\S\ref{subsec:stoch_maps}, \S\ref{subsec:algorithm}). 
\item[2.] We consider several formulations of the barycenter problem, and for each particular formulation, we obtain quality bounds for the recovered solutions (\S\ref{subsec:method_derivation}).
\item[3.] We showcase the performance of our method on moderate- and high dimensional data (\S\ref{sec:experiments}), e.g., in image space and the latent space of a pre-trained StyleGAN.

\end{itemize}

\vspace{-3 mm}
We acknowledge that our proposed approach borrows a significant number of theoretical and technical ideas from \citep{kolesov2023energy}.
Compared to the majority of other works, in practice, we do not limit ourselves to the quadratic cost functions and demonstrate an intriguing \textit{shape}-\textit{color} setup (\S\ref{subsec:colorshape}). 
To the best of our knowledge, this setup paves a new principled way for creating generative distributions which aggregate some practically desired features gathered from multi-source data. 
We believe that the development of this idea will be crucial for solving real-world tasks \citep[\S B.2]{kolesov2023energy}.

\textbf{Notations.} We write $\overline{K} = \{1, 2, \dots, K\}$ for $K \in \mathbb N$. For objects $o_1, o_2, \dots$ indexed by natural numbers, we use $o_{1:K}$ to denote the tuplet $(o_1, o_{2}, \dots, o_K)$.
Throughout the paper, $\cX\subset \bbR^{D'}, \cY \subset \bbR^D$ and $\cX_k \subset \bbR^{D_k}$ are compact subsets of Euclidean spaces. 
The set of real-valued continuous functions on $\cX$ is denoted by $\cC(\cX)$. 
We use $\cP(\cX)$ for the set of probability distributions on $\cX$ and ${\Pi(\bbP) \subset \cP(\cX \times \cY)}$ for the set of probability distributions on $\cX \!\times\! \cY$ with fixed first marginal $\bbP \!\in\! \cP(\cX)$.
All probability distributions on $\cX\! \times\! \cY$ with the marginals $\bbP$ and $\bbQ \in  \cP(\cY)$ are denoted by $\Pi(\bbP, \bbQ) \subset \Pi(\bbP)$ (a.k.a.\ transport plans). For $\pi \in \bbP(\cX \times \cY)$, $\pi(\cdot \vert x)$ denotes the conditional distribution given the $\cX$-coordinate, and $\pi^{\cY} \in \cP(\cY)$ denotes the second marginal distribution of $\pi$. For a measurable map $T$, we write $T_\#$ for the corresponding push-forward operator.

\section{Background}\label{sec:background}

To begin with, we recall the OT problem and its semi-dual formulation (\S\ref{subsec:weakot}). Then we proceed to formulate the corresponding OT barycenter problem (\S\ref{subsec:weakbary}). Finally, we specify our learning setup (\S\ref{subsec:learningsetup}). Throughout the paper, when we refer to the OT problem we mean both the \textit{classical} (strong) formulation, and its \textit{weak} generalization.

\textbf{Entropies and energy distances.} 
We use $H$ and $\KL$ to denote the differential entropy and the Kullback-Leibler divergence, respectively, see \citep[Def. 1.1]{nutz2021introduction}. 
Let $\ell \in \cC(\cY \times \cY)$ be a \textit{semimetric of negative type} \citep[\S2.1]{sejdinovic2013equivalence}, e.g., $\ell(y, y') = \Vert y - y' \Vert_2$. For $\mu_1, \mu_2 \in \cP(\cY)$, the square of the energy distance w.r.t. $\ell$  \citep[Eq.2.5]{sejdinovic2013equivalence} is given by:
\begin{eqnarray*}
	\cE_{\ell}^2(\mu_1, \mu_2) \defeq 2\!\!\!\!\E{{\scriptsize
    \makecell{y_1 \!\sim\! \mu_1 \\y_2 \!\sim\! \mu_2}
    }}\!\!\!\ell(y_1, y_2) - \!\!\!\E{{\scriptsize \makecell{y_1 \!\sim\! \mu_1 \\y_1' \!\sim\! \mu_1}}} \!\!\!\ell(y_1, y_1') - \!\!\!\E{{\scriptsize \makecell{y_2 \!\sim\! \mu_2 \\y_2' \!\sim\! \mu_2}}}\!\!\! \ell(y_2, y_2'),
\end{eqnarray*}
where $y_1,y_1',y_2,y_2'$ are pairwise independent.
The energy distance $\cE_\ell$ is a metric on $\cP(\cY)$ \citep{klebanov2005n}. It is  deeply connected to MMDs \citep{sejdinovic2013equivalence} and induces a distance on $\Pi(\bbP)$ via
\begin{eqnarray}
    \rho_\ell(\pi_1, \pi_2) \defeq \sqrt{\E{x \sim \bbP} \cE_\ell^2(\pi_1(\cdot\vert x), \pi_2(\cdot\vert x ) )}, \label{eq:rho_ell_metric}
\end{eqnarray}
where $\pi_1, \pi_2 \in \Pi(\bbP)$.
Indeed, $\rho_\ell$ defines a metric and was introduced in \citep[Appendix D]{asadulaev2024neural}.

\subsection{Classical and Weak Optimal Transport}\label{subsec:weakot}

Let $\bbP \in \cP(\cX), \bbQ \in \cP(\cY)$ and consider $c\in \cC(\cX\times \cY)$, called the \textit{ground} cost function. The classical OT problem \citep{kantorovitch1958translocation} between $\bbP$ and $\bbQ$ consists in:
\begin{eqnarray}
    \text{OT}_c(\bbP, \bbQ) \defeq \inf\limits_{\pi \in \Pi(\bbP, \bbQ)}\Big\{\E{(x, y) \sim \pi} \, c(x, y)\Big\}. \label{eq:strongot_primal}
\end{eqnarray}
The specific choice $c(x, y) = \frac{1}{2}\Vert x - y \Vert_2^2$ in \eqref{eq:strongot_primal} yields  $\bbW_2^2(\bbP, \bbQ) \defeq \inf_{\pi \in \Pi(\bbP, \bbQ)} \bbE_{(x, y) \sim \pi} \frac{1}{2}\Vert x - y \Vert_2^2$, commonly known as the (squared) Wasserstein-2 distance. 

Weak OT is a generalization of classical OT \eqref{eq:strongot_primal}, %
where \textit{weak} cost functions ${C \!: \cX \!\times\! \cP(\cY) \!\rightarrow\! \bbR}$ are employed. The weak OT problem \citep{gozlan2017kantorovich} then is defined as follows:
\begin{eqnarray}
    \text{OT}_C(\bbP, \bbQ) \defeq \inf\limits_{\pi \in \Pi(\bbP, \bbQ)}\E{x \sim \bbP} \, C\big(x, \pi(\cdot \vert x)\big). \label{eq:weakot_primal}
\end{eqnarray}
A typical example of a weak cost function $C$ is:
\begin{eqnarray*}
    C(x, \mu) = \E{y\sim \mu}c(x, y) + \gamma \cR(\mu),\label{eq:weak_cost}
\end{eqnarray*}
where $c(x, y)$ is a ground cost function, $\gamma \!>\! 0$ is the regularization parameter, and $\cR \!:\! \cP(\cY) \!\rightarrow\! \bbR$ is a \textit{regularizer}.
Introducing regularization into the classical OT formulation \eqref{eq:strongot_primal} has proven to be beneficial, leading to the development of efficient optimization algorithms  \citep{cuturi2013sinkhorn, blondel2018smooth} or the improvement of theoretical properties \citep{korotin2023kernel, asadulaev2024neural}. 
Table \ref{table-weakcost-examples} presents some popular examples explored in recent OT literature.
If a weak cost function $C$ is \textit{appropriate}\footnote{Lower-boundedness, convexity in the second argument and joint lower semicontinuity on $\cX \times \cP(\cY)$.}, then weak OT \eqref{eq:weakot_primal} admits a minimizer $\pi^*$ called the \textit{OT plan} \citep[Th. 1.2]{backhoff2019existence}. Given that $\cX, \cY$ are compact and $c$ is continuous, the weak cost functions from Table \ref{table-weakcost-examples} are \textit{appropriate}, see \citep{backhoff2019existence, korotin2023kernel}.

\textbf{Weak OT duality.} For an appropriate weak cost function $C$, the weak OT problem \eqref{eq:weakot_primal} satisfies the duality
\vspace{-1mm}
\begin{eqnarray}
    \text{OT}_C(\bbP, \bbQ) = \sup\limits_{f \in \cC(\cY)} \bigg\{ \E{x \sim \bbP} f^C(x) + \E{y \sim \bbQ} f(y) \bigg\}, \label{eq:weakot_dual}
\end{eqnarray}
\vspace{-1mm}\newline
where $f^C(x) \defeq \inf_{\mu \in \cP(\cY)} \big\{C(x, \mu) - \bbE_{y \sim \mu} f(y)\big\}$ is the \textit{weak $C$-transform}, 
see \citep{gozlan2017kantorovich}, \citep[Th. 1.3]{backhoff2019existence} for further details.

\begin{table}
\vspace{-2mm}
\footnotesize
\begin{tabular}{x{40mm}}
  \hline
    \rule{0pt}{9pt}\textbf{Cost function}\\
    \hline
  classical \citep{fan2023neural, rout2022generative} \\
  \rule{0pt}{7pt}\vspace*{5pt}$\epsilon$-entropic \citep{mokrov2024energyguided} \\
  $\gamma$-weak (kernel) quadratic \citep{korotin2023neural, korotin2023kernel}\vspace{2pt} \\
  \hline
\end{tabular}\hspace*{-6mm}
\begin{tabular}{x{40mm}}
  \hline
    \rule{0pt}{9pt}$C(x, \mu) =$\\
    \hline
  \rule{0pt}{10.5pt}\vspace*{5pt}$\bbE_{y \sim \mu} c(x, y)$ \\
   \rule{0pt}{8.7pt}\vspace*{3pt}$\bbE_{y \sim \mu} c(x, y) - \epsilon H(\mu)$\\
   \rule{0pt}{8.7pt}$\bbE_{y \sim \mu} \Vert x - y \Vert^\alpha - \frac{\gamma}{2} \bbE_{y, y' \sim \mu} \Vert y - y'\Vert^\alpha \!, \alpha \in [1, 2]$\vspace{2pt}\\
  \hline
\end{tabular}
\vspace{-2mm}
\caption{Popular instances of weak cost functions.}
\vspace{-4mm}
\label{table-weakcost-examples}
\end{table}
\vspace{-1mm}
\subsection{Classical and Weak OT Barycenter}\label{subsec:weakbary}

Let $\bbP_k \in \cP(\cX_k)$ be given distributions and $C_k\!:\! \cX_k \!\times\! \cP(\cY) \rightarrow \bbR$ be appropriate weak cost functions, $k \in \overline{K}$. For positive weights $\lambda_k, \sum_{k = 1}^K \lambda_k = 1$, the \textit{weak OT barycenter} problem consists in finding a distribution that minimizes the (weighted) sum of OT problems with fixed first marginals in $\bbP_{1:K}$:
\vspace{-3mm}
\begin{eqnarray}
    \cL^* \defeq \inf\limits_{\bbQ \in \cP(\cY)} \sum_{k = 1}^{K} \lambda_k \text{OT}_{C_k}(\bbP_k, \bbQ).\label{eq:weakbary_primal}
\end{eqnarray}
\vspace{-3mm}\newline
This equation encompasses various OT barycenter formulations explored in computational OT literature \citep{pmlr-v139-fan21d, li2020continuous, cazelles2021novel}. The utilization of certain weak OT problems in \eqref{eq:weakbary_primal} instead of classical OT \citep{agueh2011barycenters} permits us to derive theoretical guarantees for the recovered barycenter solutions, see Theorem \ref{thm:duality_gaps} as well as \citep{li2020continuous, kolesov2023energy}. 
Problem \eqref{eq:weakbary_primal} admits a minimizer $\bbQ^* \!\in\! \cP(\cY)$, which is unique provided that the weak cost functions $C_k$ are strictly convex w.r.t.\ the second argument. 
These assertions follow from standard measure-theoretic arguments, cf.\ \citep{backhoff2019existence}, and have been observed in previous works, e.g., \citep[\S 2.2]{kolesov2023energy}.

\textbf{Considered weak cost functions.} 
The weak OT barycenter task \eqref{eq:weakbary_primal} is overly general and thus has to be instantiated. 
To demonstrate the versatility of our approach, we stick in our experiments (\S\ref{sec:experiments}) to the following families of cost functions, adapted from Table \ref{table-weakcost-examples}:
\vspace{-2mm}
\begin{itemize}[leftmargin=5mm]
	\item Classical: \inlineequation[eq:strongcost]{C(x, \mu) = \bbE_{y\sim\mu}c(x, y).\hspace*{24.5mm}}
	\item $\epsilon$-KL: \inlineequation[eq:klcost]{C(x, \mu) = \bbE_{y\sim\mu}c(x, y) + \epsilon \KL(\mu \Vert \mu_0).\hspace*{10mm}} \newline Distribution $\mu_0$ is a given prior, e.g., Normal; $\epsilon > 0$.
	\item $\gamma$-Energy: \inlineequation[eq:energycost]{C(x, \mu) = \bbE_{y\sim\mu}c(x, y) + \gamma\cE_\ell^2(\mu, \mu_0).\hspace*{5mm}} \newline Distribution $\mu_0$ is a given prior; $\gamma > 0$.
\end{itemize}
\vspace{-1mm}
Here, $\mu_0 \in \cP(\cY)$ can be seen as a prior distribution, reflecting the prior knowledge about the barycenter distribution $\bbQ^*$, which is precisely the way we use it in our experiments in the latent space.

The weak cost functions introduced above are \textit{appropriate}. Indeed, they are lower-bounded (due to the compactness of $\cX, \cY$). 
Convexity and lower semicontinuity of $\KL$ are well-known \citep{nutz2021introduction}, and the same properties also hold true for $\cE_\ell^2$, see, e.g., \citep[Appendix A.2]{asadulaev2024neural}. 
For the classical cost function, these properties follow directly from its linearity (integrated against measures) and continuity of ground cost function $c$ on the compact $\cX\times\cY$. Meanwhile, the cost functions \eqref{eq:strongcost}, \eqref{eq:klcost}, \eqref{eq:energycost} have different theoretical properties which affect the analysis, see, e.g., Theorem \ref{thm:duality_gaps}.
\vspace{-2mm}
\subsection{Computational setup}\label{subsec:learningsetup} 
\vspace{-1mm}
In practice, the distributions $\bbP_k$ are not explicitly available, which causes ambiguity on how to adopt \eqref{eq:weakbary_primal} for real-world problems. To avoid confusion, we describe below in detail our \textbf{learning setup}. Let $\pi_k^*$, $k \in \overline{K}$, be the (unknown) weak OT plans between the distributions $\bbP_k$ and the weak OT barycenter $\bbQ^*$ for given (known) cost functions $C_k$. 
We assume empirical samples (datasets) $X_k \!=\! \{x_k^i\}_{i = 1}^{N_k} \sim \bbP_k$ are given. Our goal is to find approximations $\widehat{\pi}_k \!\in\! \Pi(\bbP_k)$ of the true OT plans $\pi_k^*$. These approximations are assumed to realize the \textit{conditional} sampling procedure, i.e., taking a sample $x_k \!\sim\! \bbP_k$ as input and producing samples from $\widehat{\pi}_k(\cdot \vert x_k)$ as output. We stress that the input samples $x_k$ are not necessarily from the training datasets $X_k$.
The setup described above is called \textbf{continuous} \citep{li2020continuous, kolesov2023energy}. 
Alternatively, the discrete setup \citep{peyre2019computational} aims at solving the OT barycenter problem between empirical measures $\widehat{\bbP}_k = \sfrac{1}{N_k}\sum_{n = 1}^{N_k}\delta_{x_k^n}$. The resulting OT plan approximations operate exclusively with samples presented in the datasets $X_k$, which then requires extra effort to adapt for new samples \citep{de2021consistent}.

\vspace{-4 mm}
\section{Related works}\label{sec:related_works}

In this section, we give an overview of related works. 
We start with adversarial Neural OT methods, which form the basis of our approach. 
Then we discuss competitive OT barycenter methods that follow the continuous setup (\S\ref{subsec:learningsetup}).

\textbf{Maximin Neural OT solvers} leverage the dual problem \eqref{eq:weakot_dual} through optimization of adversarial-like $\max$-$\min$ objectives. Among them, \citep{henry2019martingale, rout2022generative, gazdieva2022optimal, fan2023neural} focus on classical OT \eqref{eq:strongot_primal}. On the other hand, \citep{korotin2023kernel, korotin2023neural} address specifically regularized cost functions, see Table \ref{table-weakcost-examples}. Recent works consider more exotic OT formulations, e.g., general OT cost \textit{functionals} \citep{asadulaev2024neural}, diffusion-based Entropic OT \citep{gushchin2023entropic}, etc.

\textbf{Continuous OT barycenter methods} differentiate by the two key characteristics: which problem they solve (general/particular form of the barycenter problem) and which computational algorithm they use. 

In terms of the first characteristic, the majority of methods \citep{korotin2021continuous, pmlr-v139-fan21d, korotin2022wasserstein, noble2023treebased} deal with quadratic (Euclidean) ground cost functions, i.e., $c_k(x_k, y) = \sfrac{1}{2}\Vert x_k - y \Vert_2^2$. 
A limited number of works consider general $c_k$ \citep{li2020continuous, chi2023variational, kolesov2023energy}, but require additional entropic (quadratic) regularization. 
In contrast, our proposed approach provides greater flexibility, as it permits a wide range of admissible weak OT cost functions, including those introduced in \S\ref{subsec:weakbary}, but \underline{not limited} to them.

Now we discuss existing computational procedures for solving the OT barycenter task. A branch of works \citep{pmlr-v139-fan21d, korotin2022wasserstein} employs tri-level ($\min$-$\max$-$\min$) adversarial objectives, which may be hard to optimize.
Other works manage to develop non-adversarial minimization objectives, but with limitations: 
\citep{li2020continuous} \textit{requires} a fixed prior distribution, \citep{noble2023treebased, kolesov2023energy} utilize time-consuming sampling procedures, \citep{korotin2021continuous} relies on Input Convex NNs \citep{amos2017input} which have scalability issues \citep{korotin2021neural}. 
In contrast, our proposed approach uses a bi-level ($\max$-$\min$) computational algorithm (\S\ref{subsec:algorithm}) widely embraced in generative modeling, specifically, in the field of Neural OT. 
Noteworthy, \citep{chi2023variational} also proposes a bi-level objective but utilizes the \textit{reinforce} procedure. The latter is tricky, e.g., it requires variance reduction techniques. A \underline{comprehensive comparison} of our proposed method and \citep{chi2023variational} can be found in Appendix~\ref{app:ext-chi-discussion}.

\vspace{-4 mm}
\section{Proposed Method}\label{sec:method} 

In \S\ref{subsec:method_derivation}, we derive our novel $\max$-$\min$ optimization objective for learning weak OT barycenters and establish error bounds for approximate solutions. 
After that, in \S\ref{subsec:stoch_maps}, we describe the parametrization of plans $\Pi(\bbP)$ as (stochastic) maps. The described techniques are crucial when adapting our derived objective to practice. In \S\ref{subsec:algorithm}, we present the resulting computational algorithm. All the \underline{proofs} can be found in Appendix~\ref{app:proofs}. An \underline{additional conceptual derivation} of our main result (Theorem \ref{thm:our_maxmin_solves_weakbary}) is in Appendix~\ref{app:intuitive-derivation}.
\vspace{-2mm}
\subsection{Maximin Dual Weak OT Barycenter formulation}\label{subsec:method_derivation}

Let $f_k \in \cC(\cY)$, $\pi_k \in \Pi(\bbP_k)$. We introduce the functionals:
\vspace{-4mm}
\begin{eqnarray}
    \cV(f_{1:K}, \pi_{1:K}) \defeq \sum_{k = 1}^{K} \lambda_k \bigg\{\E{x_k \sim \bbP_k} C_k\big(x_k, \pi(\cdot\vert x_k)\big)& \nonumber \\
    \hspace{20mm}- \E{x_k \sim \bbP_k}\Big(\, \E{y \sim \pi_k(\cdot \vert x_k)} \, f_k(y)\Big)\bigg\};& \label{eq:V_func} \\
    \cL(f_{1:K}) \defeq \inf_{{\scriptsize \makecell{\pi_k \in \Pi(\bbP_k), \\ (k \in \overline{K})}}} \cV(f_{1:K}, \pi_{1:K}).\hspace{15mm}& \label{eq:L_func}
\end{eqnarray}
\vspace{-4mm}\newline
Theorem \ref{thm:our_maxmin_solves_weakbary} presents our main theoretical result. It states that if the potentials $f_{1:K}$ satisfy the \textbf{congruence} condition $\sum_{k=1}^K \!\lambda_k f_k \!\equiv\! 0$, then the optimization of \eqref{eq:L_func} yields the optimal value to the OT barycenter objective \eqref{eq:weakbary_primal}.
\begin{theorem}[\normalfont{$\max$-$\min$ formulation for OT barycenter}]
The optimal value $\cL^*$ of the OT barycenter problem \eqref{eq:weakbary_primal} is given by the following $\max$-$\min$ objective:
\begin{eqnarray}
    \mathcal{L}^{*}= \!\!\!\!\!\!\sup\limits_{\sum\! \lambda_k f_k = 0}\!\!\!\!\!\cL(f_{1:K}) = \!\!\!\!\!\!\sup\limits_{\sum \!\lambda_k f_k = 0}\,\inf_{{\scriptsize \makecell{\pi_k \in \Pi(\bbP_k), \\ (k \in \overline{K})}}}\!\!\!\!\!\cV(f_{1:K}, \pi_{1:K}). \label{eq:weakbary_maxmin}
\end{eqnarray}
\label{thm:our_maxmin_solves_weakbary}
\vspace{-8 mm}
\end{theorem}
Comparing with the literature, we establish in Theorem \ref{thm:our_maxmin_solves_weakbary} duality for the barycenter problems with \textbf{general weak} cost functions. 
Notably, existing alternatives in the literature primarily address specific cases, such as squared Euclidean ground cost \citep{agueh2011barycenters}, entropic and quadratic regularization \citep{li2020continuous, kolesov2023energy}. At the same time, the proof of our Theorem \ref{thm:our_maxmin_solves_weakbary} is an easy generalization of \citep[Thm. 4.1]{kolesov2023energy}.

\textbf{Quality bounds.} 
Recall that our ultimate goal is to approximate true OT plans $\pi_k^*$ between reference distributions $\bbP_k$ and the barycenter $\bbQ^*$, see \S\ref{subsec:learningsetup}. 
Let $(\widehat{f}_{1:K}, \widehat{\pi}_{1:K})$ be a tuple which approximately solves \eqref{eq:weakbary_maxmin}.
Theorem \ref{thm:duality_gaps} below characterizes proximity of the recovered plans $\widehat{\pi}_{1:K}$ to the true plans $\pi_{1:K}^*$, covering the cases of classical \eqref{eq:strongcost}, $\epsilon-\KL$ \eqref{eq:klcost} and $\gamma$-energy \eqref{eq:energycost}. 
The reported quality bounds are based on duality gaps, i.e., errors for solving outer ($\sup$) and internal ($\inf$) optimization problems w.r.t.\ the functional $\cV$.
\begin{theorem}[Quality bounds for recovered plans]
	Let $\widehat{f}_{1:K}$ be congruent potentials and $\widehat{\pi}_k \in \Pi(\bbP_k)$, $k \in \overline{K}$ be OT plan approximations. Consider the duality gaps:
	\begin{align}
		\delta_1(\widehat{f}_{1:K}, \widehat{\pi}_{1:K}) &\defeq \cV(\widehat{f}_{1:K}, \widehat{\pi}_{1:K}) - \cL(\widehat{f}_{1:K}); \label{eq:deltagap1} \\
		\delta_2(\widehat{f}_{1:K}) &\defeq \cL^* - \cL(\widehat{f}_{1:K}). \label{eq:deltagap2}
	\end{align}
	The following statements hold true:
	\begin{enumerate}[leftmargin=*]
           \vspace{-2 mm}
		\item Let $C_k$ be classical \eqref{eq:strongcost} cost functions, and $\cY$ be a convex set. Assume that the maps $y \!\mapsto\! c_k(x_k, y) \!-\! \widehat{f}_k(y)$ are $\beta$-strongly convex for each $k \!\in\! \overline{K}$, $x_k \!\in\! \cX_k$. Then, 
        \vspace*{-6mm}\newline
        \begin{eqnarray*}
            \sum_{k = 1}^{K} \lambda_k \E{x \sim \bbP_k} \bbW_2^2\big(\widehat{\pi}_k(\cdot \vert x), \pi_k^*(\cdot \vert x)\big) \leq \frac{2}{\beta}\Big(\delta_1 + \delta_2\Big),
        \end{eqnarray*}
		\item Let $C_k$ be $\epsilon$-$\KL$ \eqref{eq:klcost} cost functions, $\epsilon > 0$, and the prior $\mu_0$ has positive continuous Lebesgue density on $\cY$. Then,
        \vspace*{-9mm}\newline
        \begin{eqnarray*}
            \sum_{k = 1}^{K} \lambda_k \rho_{\text{TV}}(\widehat{\pi}_k, \pi_k^*)^2 \leq \frac{1}{\epsilon}\Big(\delta_1 + \delta_2\Big),
        \end{eqnarray*}
        where $\rho_{\text{TV}}$ is the total variation distance.
		\item Let $C_k$ be $\gamma$-Energy cost functions, $\gamma > 0$. Then,
        \vspace*{-6mm}\newline
        \begin{eqnarray*}
            \sum_{k = 1}^{K} \lambda_k \rho_{\ell}(\widehat{\pi}_k, \pi_k^*)^2 \leq \frac{2}{\gamma} \Big(\delta_1 + \delta_2\Big).
        \end{eqnarray*}
	\end{enumerate}
	\label{thm:duality_gaps}
 \vspace{-4 mm}
\end{theorem}
From the quality bounds, we deduce that when the pair $(\widehat{f}_{1:K}, \widehat{\pi}_{1:K})$ solves the inner and outer optimization in \eqref{eq:weakbary_maxmin} relatively well, e.g., our Algorithm \ref{algorithm:otbary} below which optimizes \eqref{eq:weakbary_maxmin} converged nearly to an optimum,
then the recovered plans $\widehat{\pi}_{1:K}$ are \textbf{close} to the true OT plans.
One problem may occur with the strong convexity assumptions for the maps $y \mapsto\! c_k(x_k, y) \!-\! \widehat{f}_k(y)$ in the case of classical cost functions. 
Due to the difficulties of imposing (strong) convexity constraints \citep{korotin2021neural}, we can not guarantee them in practice. Note that previous works \citep{fan2023neural, rout2022generative} also did not care much about them. Nevertheless, the experiments (\S\ref{sec:experiments}) demonstrate a good performance of our OT barycenter solver with classical cost functions
in various setups.
At the same time, we stress that the usage of regularized cost functions \eqref{eq:klcost}, \eqref{eq:energycost} completely eliminates the need for imposing strong convexity constraints in Theorem \ref{thm:duality_gaps}. This is an important motivation for considering weak OT \eqref{eq:weakot_primal} in addition to the classical one \eqref{eq:strongot_primal} \citep{korotin2023kernel} when solving OT barycenter problems.

\textbf{Relation to prior works.} Theorem \ref{thm:duality_gaps} encompasses and \underline{generalizes} existing error analysis results for recovered OT plans, and makes them applicable for the \underline{OT barycenter problem}. 
Prior works which establish quality bounds when solving OT-related problems from dual perspectives are: \citep[Th. 3.6]{makkuva2020optimal}, \citep[Th. 4]{fan2023neural}, \citep[Th. 4.3]{rout2022generative} (classical cost); \citep[Th. 4.3]{gushchin2023entropic}, \citep[Th. 2]{mokrov2024energyguided}, \citep[Th. 2]{kolesov2023energy} (entropic); \citep[Th. 3]{asadulaev2024neural} (general strongly convex cost). 

\vspace{-2 mm}
\subsection{Parameterizing transport plans via stochastic maps}\label{subsec:stoch_maps}

Objective \eqref{eq:weakbary_maxmin} prescribes optimization over plans ${\pi_k \!\in\! \Pi(\bbP_k)}$. Direct optimization over probability distributions is non-trivial, and we adopt the parameterization of plans with functions. We pursue the following approaches.

\underline{Stochastic maps.} The approach is similar to \citep[\S 4.1]{korotin2023neural}. We introduce an auxiliary space $\cS \subset \bbR^{D_s}$, an atomless distribution $\bbS \in \cP(\cS)$, and consider measurable maps $T : \cX \times \cS \rightarrow \cY$. Every plan $\pi \in \Pi(\bbP)$ can be implicitly represented by $T_\pi$ s.t.\ $\pi(\cdot \vert x) = T_\pi(x, \cdot)_\# \bbS$. Particularly, given $x\sim\bbP$ and $s \sim \bbS$, the pair $(x, T_\pi(x, s))$ is distributed as $\pi$. In turn, every measurable $T : \cX \times \cS \rightarrow \cY$ implicitly specifies a plan $\pi_T \in \Pi(\bbP)$. Taking the advantage of stochastic maps parameterization, \eqref{eq:weakbary_maxmin} allows for an alternative objective:
\begin{align}
     &\hspace*{10mm}\cL^* = \!\!\!\sup\limits_{\sum \!\lambda_k f_k = 0}\,\inf_{T_{1:K}}\widetilde{\cV}(f_{1:K}, T_{1:K}); \label{eq:weakbary_maxmin_maps}\\
     &\widetilde{\cV}(f_{1:K}, T_{1:K}) \defeq \sum_{k = 1}^{K} \lambda_k \bigg\{\E{x_k \sim \bbP_k} C_k\big(x_k, T_k(x_k, \cdot)_\# \bbS \big) \nonumber \\
     &\hspace*{23mm}- \E{x_k \sim \bbP_k}\,\E{s \sim \bbS} \,f_k\big(T_k(x_k, s)\big)\bigg\}.\label{eq:V_func_T}
\end{align}
Optimization \eqref{eq:weakbary_maxmin_maps} is generic and can be implemented in practice as far as it is possible to estimate the weak cost function by samples. This is the case for classical \eqref{eq:strongcost} and $\gamma$-energy \eqref{eq:energycost} cost functions.

\underline{Gaussian model.} We parameterize conditional plans $\pi(\cdot \vert x)$ as Gaussian distributions $\cN(\cdot \vert \mu(x), \text{diag}(\sigma^2(x)))$. The method is widely used, e.g., in the celebrated VAEs \citep{kingma2014auto}. Technically, the optimization w.r.t.\ transport plans boils down to optimization of functions $\mu(x), \sigma(x)$
given by, e.g., encoding NNs. The proposed Gaussian model is the particular instance of stochastic maps model with $\bbS \!=\! \cN(0, I_{D_s})$, $T(x, s) \!=\! \mu(x) + \sigma(x)\cdot s$. Note that the Gaussian parameterization allows explicitly computing the $\KL$ regularization term in \eqref{eq:klcost}. Theoretically, the substitution of $\pi(\cdot\vert x) \in \cP(\cY)$ with Gaussians is not very accurate and should be seen as a technical trick. To alleviate the imprecision of this approximation, we utilize Gaussian model exclusively in the latent space, see \S\ref{sec:experiments}. 

\underline{Deterministic maps.} We consider measurable maps $T : \cX \rightarrow \cY$ and model conditional plans as deterministic distributions with delta functions $\pi(\cdot \vert x) = \delta_{T(x)}(\cdot)$. The method is a particular case of stochastic maps parameterization with removed stochasticity. This can be naturally applied to classical OT cost. 
In this case, modelling with deterministic maps has a direct relation to the seminal formulation of OT problem due to Monge \citep{monge1781memoire} and has found its application in a number of works, e.g., \citep{makkuva2020optimal, korotin2021wasserstein, rout2022generative, fan2023neural}. Such an extensive practical utilization combined with our encouraging numerical validation (\S\ref{sec:experiments}) makes the deterministic maps model meaningful for solving the OT barycenter problem.

\vspace{-3 mm}
\subsection{Computational barycenter algorithm}\label{subsec:algorithm}

In this subsection, we develop a practical optimization procedure for solving \eqref{eq:weakbary_maxmin_maps}.
We parameterize (stochastic) maps $T_{1:K}$ and potentials $f_{1:K}$ as neural networks. The parameter space for the maps is $\Phi = \Phi_1 \times \Phi_2 \dots \times \Phi_K$; the NN maps are $T_{k, \phi} : \bbR^{D_k}\! \times\! \bbR^{D_s} \rightarrow \bbR^D$, $\phi = (\phi_1, \dots \phi_K) \in \Phi$. When modeling deterministic maps, we omit stochastic dimensions $\bbR^{D_s}$. The parameter space for the potentials is $\Theta = \Theta_1 \!\times\! \Theta_2\! \dots \!\times\! \Theta_K$. To ensure the congruence condition, we parameterize the potentials $f_{k, \theta}$, $\theta \!=\! (\theta_1, \dots, \theta_K) \in \Theta$ with help of auxiliary NNs $g_{\theta_k} \!: \bbR^D \!\rightarrow \bbR$ as follows: $f_{k, \theta} \!\defeq\! g_{\theta_k}\! - \!\sum_{k' = 1}^{K} \lambda_{k'} g_{\theta_{k'}}$. This trick is used in \citep{li2020continuous, kolesov2023energy}. Below, $T_{1:K, \phi}$ and $f_{1:K, \theta}$ denote the tuplets of the NN maps and potentials. 

\textbf{Training.} To optimize $\max$-$\min$ objective \eqref{eq:weakbary_maxmin_maps} with NNs we utilize stochastic gradient ascent-descent algorithm by performing several minimization steps w.r.t.\ $T_{1:K, \phi}$ per each maximization step w.r.t.\ congruent potentials $f_{1:K, \theta}$. The functional $\widetilde{\cV}$ in \eqref{eq:weakbary_maxmin_maps} is estimated from samples via Monte-Carlo. We derive random batches from reference distributions $\bbP_k$ and (optionally) random batches from auxiliary $\bbS$. This allows us to straightforwardly estimate the second term $\bbE_{x \sim \bbP_k} \bbE_{s \sim \bbS} f_{k, \theta} (T_{k, \phi}( x_k, s))$ in \eqref{eq:V_func_T}. The details on approximating weak cost functions $\widehat{C}$ are below. These approximations are averaged over batches derived from $\bbP_k$ when computing the estimator for the 1st term in \eqref{eq:V_func_T}.
\begin{itemize}[leftmargin=5mm]
\vspace{-2 mm}
	\item \textbf{Classical}: Sample auxiliary batch $S \!\sim\! \bbS$, use mean: \newline\vspace{-2mm}\newline\hspace*{14mm}
    $\widehat{C}(x, T(x, S)) = \sum\limits_{s \in S} \frac{c\left(x, T(x, s)\right)}{\vert S \vert}.$
    
    For deterministic maps model: $\widehat{C}(x, \!T(x)) \!=\! c(x, \!T(x))$.
	\item \textbf{$\epsilon$-KL}: We utilize \textit{Gaussian model} for stochastic maps, i.e, $T$ is defined by means and deviations $\mu$, $\sigma$, see \S\ref{subsec:stoch_maps}.  Distributions $\bbS, \mu_0$ are assumed to be Gaussian. Then,
    \newline\vspace{-2mm}\newline\hspace*{3mm} $\widehat{C}(x, T(x, S)) = \sum\limits_{s \in S} \frac{c(x, T(x, s))}{\vert S \vert},\, S \sim \bbS \,\, \text{{\scriptsize \# sample mean}}$
    \newline\vspace{-1mm}\newline\hspace*{12mm}$+ \epsilon \KL(\cN(\mu(x), \sigma(x))\Vert \mu_0).\,\,\,\text{{\scriptsize \# computed analytically}}$
 
	\item \textbf{$\gamma$-Energy}: Sample auxiliary ($S \sim \bbS$) and prior \newline ($Y_0 \sim \mu_0$) batches. Then compute:
    \newline\vspace{-2mm}\newline\hspace*{8mm}$\widehat{C}(x, T(x, S), Y_0) = \sum\limits_{s \in S} \frac{c(x, T(x, s))}{\vert S \vert} +$
    \newline\vspace{-1mm}\newline\hspace*{1mm}$\gamma \!\sum\limits_{s \in S}\!\bigg( 2\!\!\sum\limits_{y \in Y_0}\! \frac{\ell(T(x, s), y)}{\vert S \vert \vert Y_0 \vert} - \!\!\!\!\sum\limits_{s' \in S \setminus \{ s \}} \!\frac{\ell(T(x, s), T(x, s'))}{\vert S \vert (\vert S \vert - 1)} \!\bigg),$
    \newline\vspace{3mm}\newline which is the estimator of \eqref{eq:energycost} up to $T$-independent constant \citep[Lemma 6]{gretton2012kernel}.
\end{itemize}
\vspace{-5 mm}
All ingredients for computing the sample estimator of $\widetilde{\cV}$ are ready, and we proceed to our main barycenter Algorithm~\ref{algorithm:otbary}. \vspace{-2 mm}

\begin{algorithm}[h!]
        \SetKwProg{Empty}{}{:}{}
        \SetAlgorithmName{Algorithm}{empty}{Empty}
        \SetKwInOut{Output}{Output}
        \textbf{Input:} Distributions $\bbP_{1:K}, \bbS$ accessible by samples;\\ 
        \hspace*{5mm}NN maps $T_{k, \phi}:\mathbb{R}^{D_k}\times\mathbb{R}^{D_s}\rightarrow\mathbb{R}^{D}$ and NN congruent\\ 
        \hspace*{5mm}potentials $f_{k, \theta}: \bbR^{D} \rightarrow \bbR$, $k \in \overline{K}$; number of  inner \\
        \hspace*{5mm}iterations $M_T$; weak cost function estimator $\widehat{C}$;\\
        \hspace*{5mm}batch sizes; prior distribution $\mu_0$.\\
        \textbf{Output:} Learned (stochastic) maps $T_{1:K, \phi^*}$ representing\\
        \hspace*{5mm}OT plans between $\bbP_k$ and barycenter $\bbQ^*$.
        
        \Repeat{not converged}{
            Sample batches $X_k \sim \bbP_k$, $k \in \overline{K}$;\\ For each $x_k \in X_k$
            sample auxiliary batch $S[x_k] \sim \bbS$;\\
            $\cV_{f}\leftarrow \sum\limits_{k \in \overline{K}} \lambda_k \Big\{\sum\limits_{x_k\in X_k} \sum\limits_{s_k\in S[x_k]} \frac{f_{k, \theta} \left(T_{k, \phi}(x_k, s_k)\right)}{\vert X_k \vert \vert S[x_k]\vert}\Big\}$;\;
            
            Update $\theta$ by using $\frac{\partial \cV_{f}}{\partial \theta}$;\;
            
            \For{$m_T = 1,2, \dots, M_{T}$}{
                \Empty{{\normalfont Sample batches $X_k \sim \bbP_k$; For each $x_k \in X_k$}}{
                Sample auxiliary batch $S[x_k]\sim \bbS$;\;\\
                (Optionally) sample prior batch $Y_0[x_k] \sim \mu_0$;\;}
                $\cV_{T_k} \leftarrow \sum\limits_{x \in X_k}\!\!\Big\{ \frac{\widehat{C}(x, T_{k, \phi}(x, S[x]), Y_0[x])}{\vert X_k \vert} \\
                - \sum\limits_{s\in S[x]}\!\!\! \frac{f_{k, \theta} \left(T_{k, \phi}(x, s)\right)}{\vert X_k \vert \vert S[x]\vert} \!\Big\}$;\;
                $\quad \cV_{T}\leftarrow \sum_{k \in \overline{K}} \lambda_k \cV_{T_k}$;\;\\
                Update $\phi$ by using $\frac{\partial \cV_{T}}{\partial \phi}$;\;
            }
        }
        \caption{OT Barycenter via Neural Optimal Transport}
        \label{algorithm:otbary}
\end{algorithm}
\vspace{-5 mm}

\section{Experimental Illustrations}\label{sec:experiments}

In this section, we showcase the performance of our proposed method. We have an illustrative experiment in 2D (\S\ref{subsec:twister}), a benchmark experiment with CelebA images (\S\ref{subsec:ave_celeba}) and an impressive experiment with shape- and color-preserving cost functions (\S\ref{subsec:colorshape}). Additional practical test cases with \underline{multidimensional Gaussians} and \underline{MNIST digits} are in Appendix \ref{app:experimens_extended}; the \underline{experimental details} are in Appendix \ref{app:experiments_details}. Our source code is available at \url{https://github.com/justkolesov/NOTBarycenters}. 

\textbf{OT barycenters constrained to image manifolds.} Following \citep{kolesov2023energy}, in a number of our experiments, we restrict the support of the desired barycenter distribution to some image manifold $\cM$. In our experiments, these manifolds are given by pre-trained StyleGAN generators $G$, i.e., $\cM = G(\cZ)$, where $\cZ = \bbR^{D_z}$ is the latent space of StyleGAN. In order to make our Algorithm \ref{algorithm:otbary} learn the barycenter on $\cM$ we use a specific parameterization of NN maps $T_{k, \phi}$. At first, they push input points to latent space and then recover target points with $G$.
The details of the specific StyleGAN models in use can be found in the respective subsections. Note that the manifold-constrained barycenter task is equivalent to \textit{unconstrained} OT barycenter problem in the latent space with cost functions $c_{k, G}(x_k, z) \defeq c_k(x_k, G(z))$, which are \textit{principially} non-quadratic. The latter makes the manifold case impractical for the majority of existing barycenter solvers.
\vspace{-2mm}
\subsection{Illustrative 2D barycenters (Twister)}\label{subsec:twister}

\begin{figure*}[t!]
    \begin{subfigure}[b]{1\linewidth}
        \centering
        \hspace{42mm}
        \includegraphics[width=0.5\linewidth]{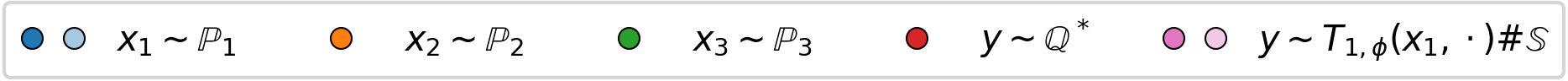}
        \label{fig:twister-legend}
    \end{subfigure}
    
    \centering
    \begin{subfigure}[b]{0.2515\linewidth}
        \centering
        \includegraphics[width=0.995\linewidth]{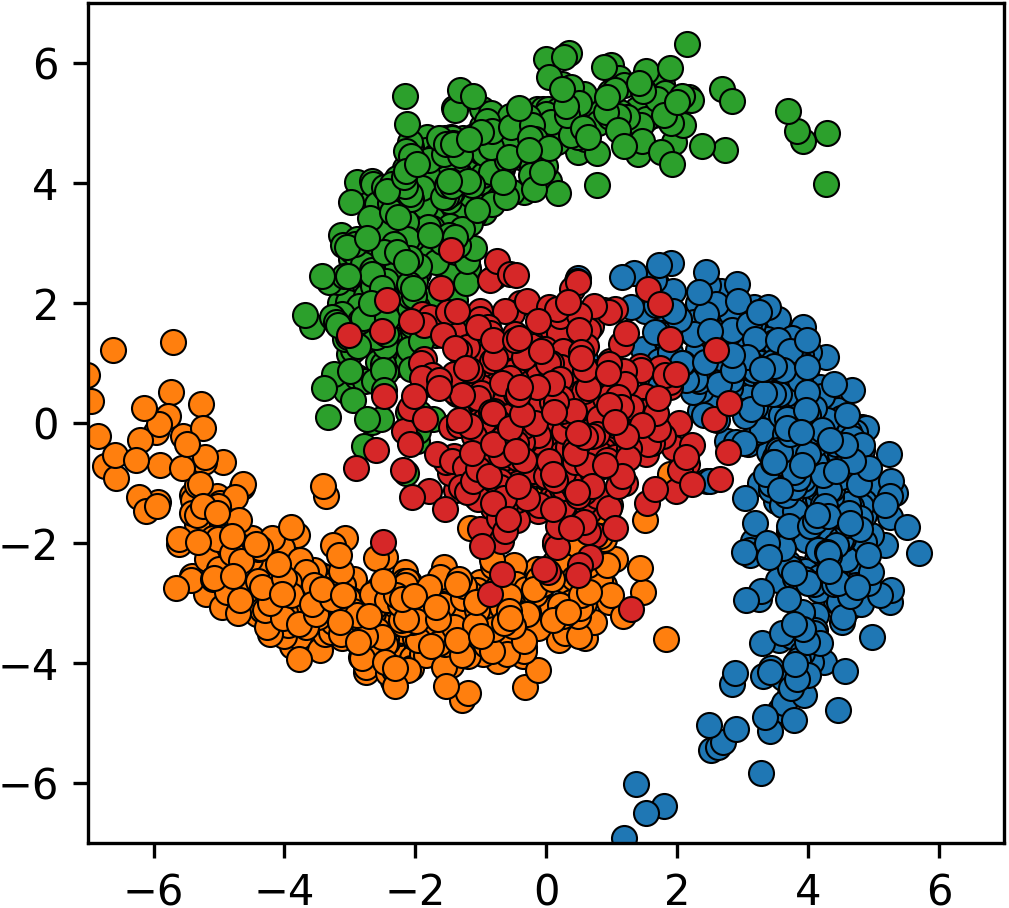}
        \caption{\centering\scriptsize Input distributions $\bbP_{1:3}$ and GT barycenter $\bbQ^*$ for the twisted cost.}
        \vspace{2.8mm}
        \label{fig:twister:inputs}
    \end{subfigure}
    \hfill
    \begin{subfigure}[b]{0.235\linewidth}
        \centering
        \includegraphics[width=0.995\linewidth]{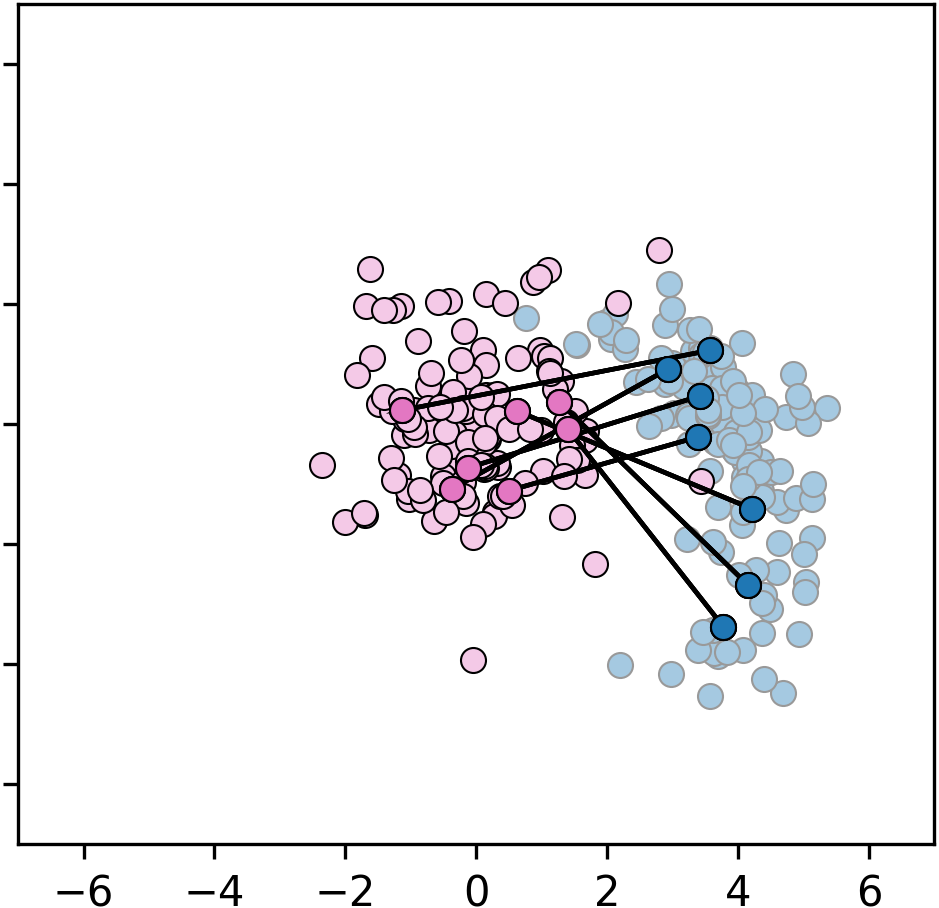}
        \caption{\centering\scriptsize \textbf{Our} unregularized (classical cost functions) barycenter, map from $\bbP_1$.}
        \vspace{2.8mm}
        \label{fig:twister:unregularized}
    \end{subfigure}
    \hfill
    \begin{subfigure}[b]{0.235\linewidth}
        \centering
        \includegraphics[width=0.995\linewidth]{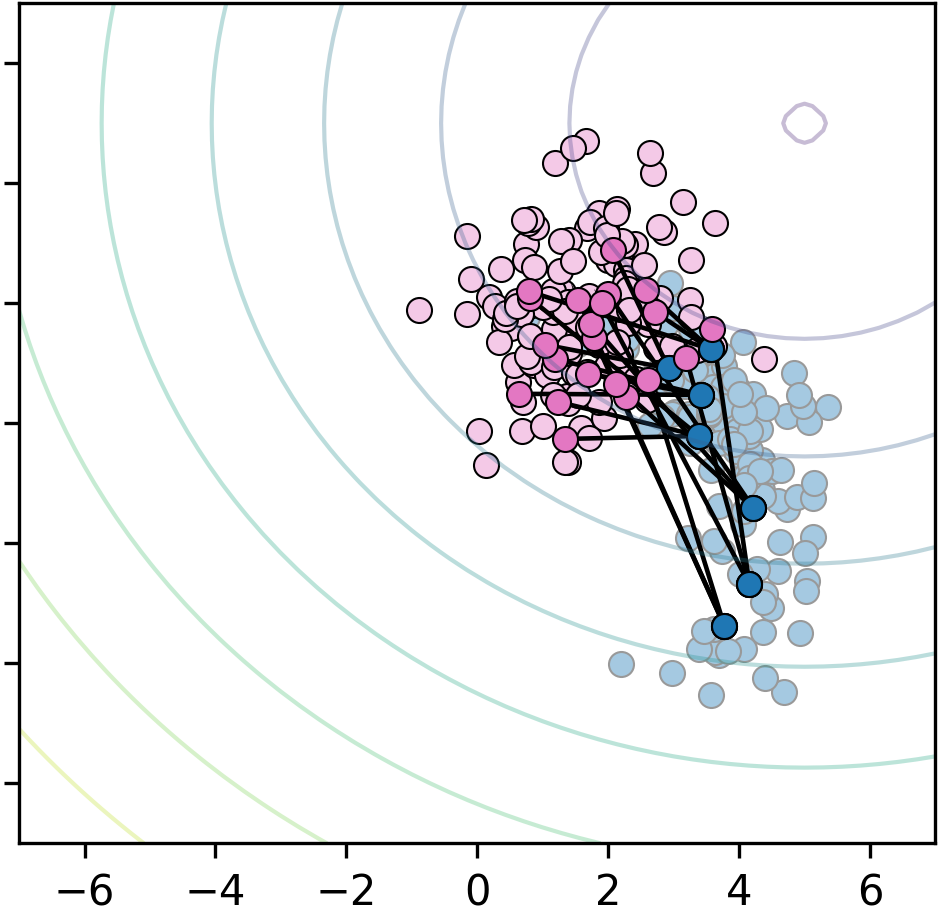}
        \caption{\centering\scriptsize \textbf{Our} regularized barycenter, $\epsilon\!=\!1$-$\KL$ cost functions, map from $\bbP_1$. The learned barycenter tends to prior $\mu_0$.}
        \label{fig:twister:kl}
    \end{subfigure}
    \hfill
    \begin{subfigure}[b]{0.235\linewidth}
        \centering
        \includegraphics[width=0.995\linewidth]{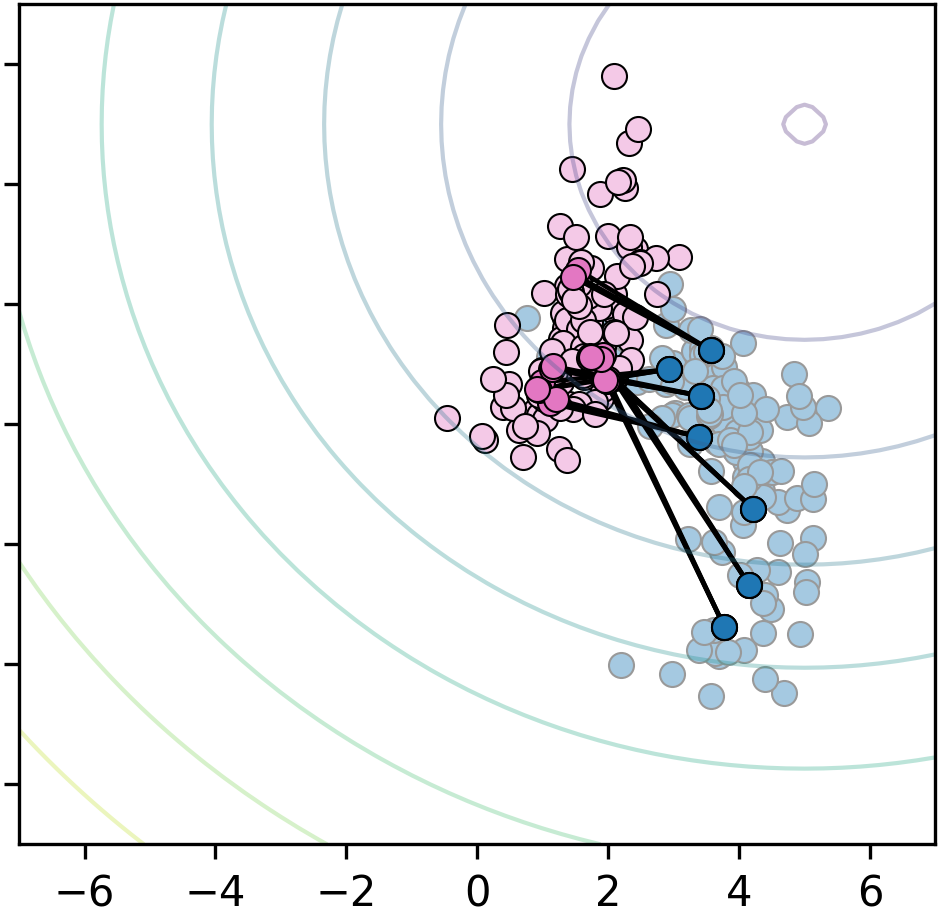}
        \caption{\centering\scriptsize \textbf{Our} regularized barycenter, $\gamma\!=\!1$-Energy costs, map from $\bbP_1$. The learned barycenter tends to prior $\mu_0$.}
        \label{fig:twister:energy}
    \end{subfigure}
    \vspace*{-2mm}
    \caption{\centering 2D Twister experiment (\S\ref{subsec:twister}). Contours represent the prior distribution $\mu_0$.} %
    \label{fig:twister}
    \vspace*{-2mm}
\end{figure*}

The experiment follows \textbf{2D Twister} setup from \citep[\S 5.1]{kolesov2023energy}. In particular, we introduce a map $u : \bbR^2 \rightarrow \bbR^2$ which rotates input points $x$ by angles proportional to $\Vert x \Vert$. Then for distributions $\bbP_{1:3}$ depicted in Fig.\ \ref{fig:twister:inputs}, classical cost functions $c_k(x_k, y) = \sfrac{1}{2}\Vert u(x_k) - u(y) \Vert_2^2$ and weights $\lambda_k \!= \!\sfrac{1}{3}$ the (unregularized) ground truth barycenter $\bbQ^*$ is known, see \citep[Appendix C.1]{kolesov2023energy}. It is the zero-centered Gaussian, see Fig. \ref{fig:twister:inputs}. Our approach without regularizations, i.e., with classical costs \eqref{eq:strongcost}, successfully recovers $\bbQ^*$, see Fig. \ref{fig:twister:unregularized}. Since our setup is symmetric, we demonstrate stochastic OT mapping only from $\bbP_1$. In the other runs (Figs. \ref{fig:twister:kl}, \ref{fig:twister:energy}) we demonstrate how the obtained barycenters are influenced by regularizations. Specifically, we use $\epsilon$-$\KL$ \eqref{eq:klcost} and $\gamma$-Energy \eqref{eq:energycost} costs  with Gaussian prior $\mu_0 = \cN\big((5, 5), I_2\big)$. As expected, the recovered barycenters do not coincide with $\bbQ^*$ and `tend' to $\mu_0$.

 \begin{figure*}[!t]
     \centering
    \begin{subfigure}[b]{0.49\linewidth} 
          \centering
          \includegraphics[width=0.995\linewidth]{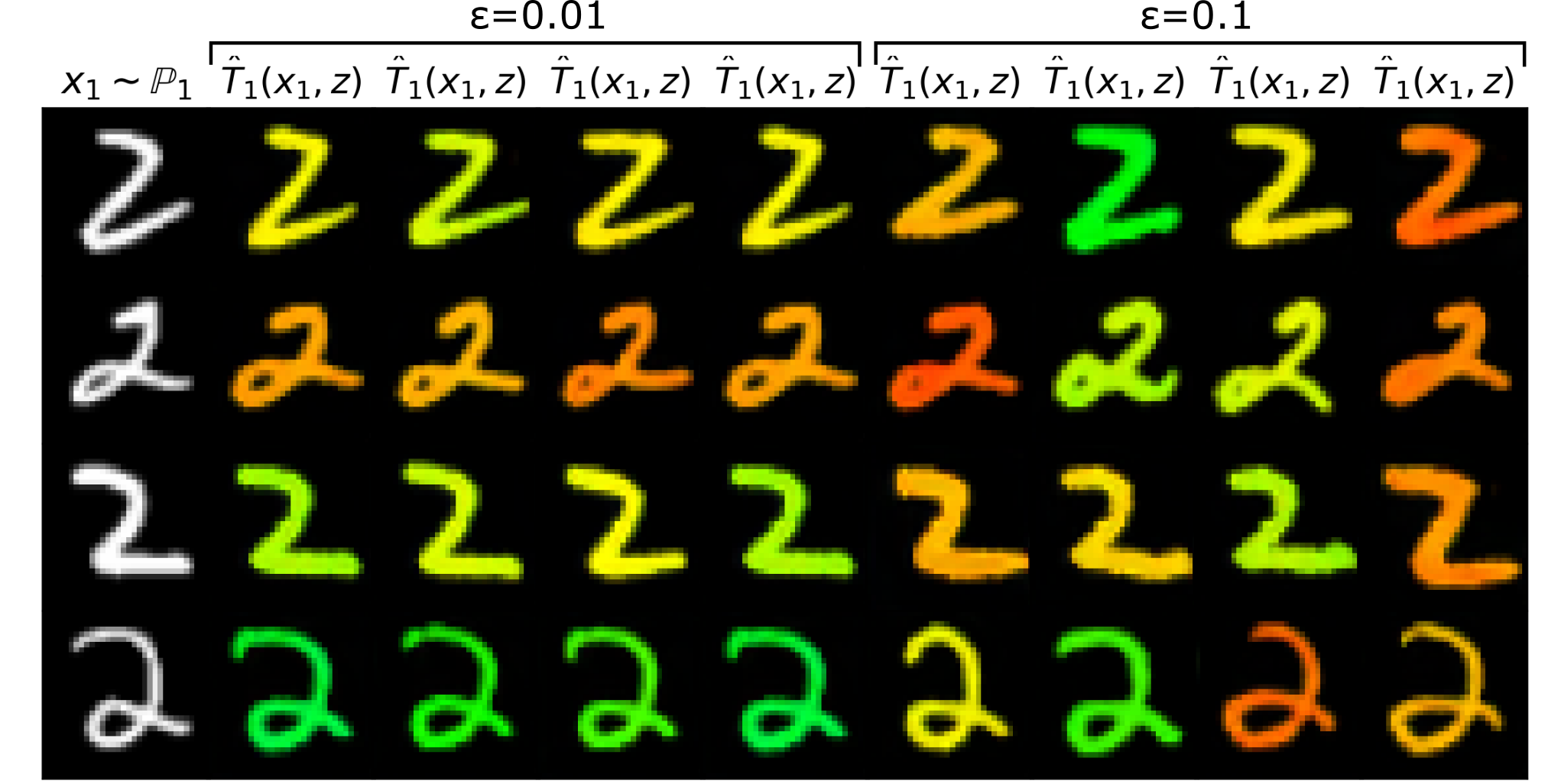}
          \caption{\centering Source ($x_1 \sim \bbP_1$) and transformed ($y = T_1(x_1, z)$) samples \newline for input $\bbP_1$ (first reference distribution).}
         \label{fig:shape}
     \end{subfigure}\hfill
      \begin{subfigure}[b]{0.49\linewidth}  
         \centering
         \includegraphics[width=0.995\linewidth]{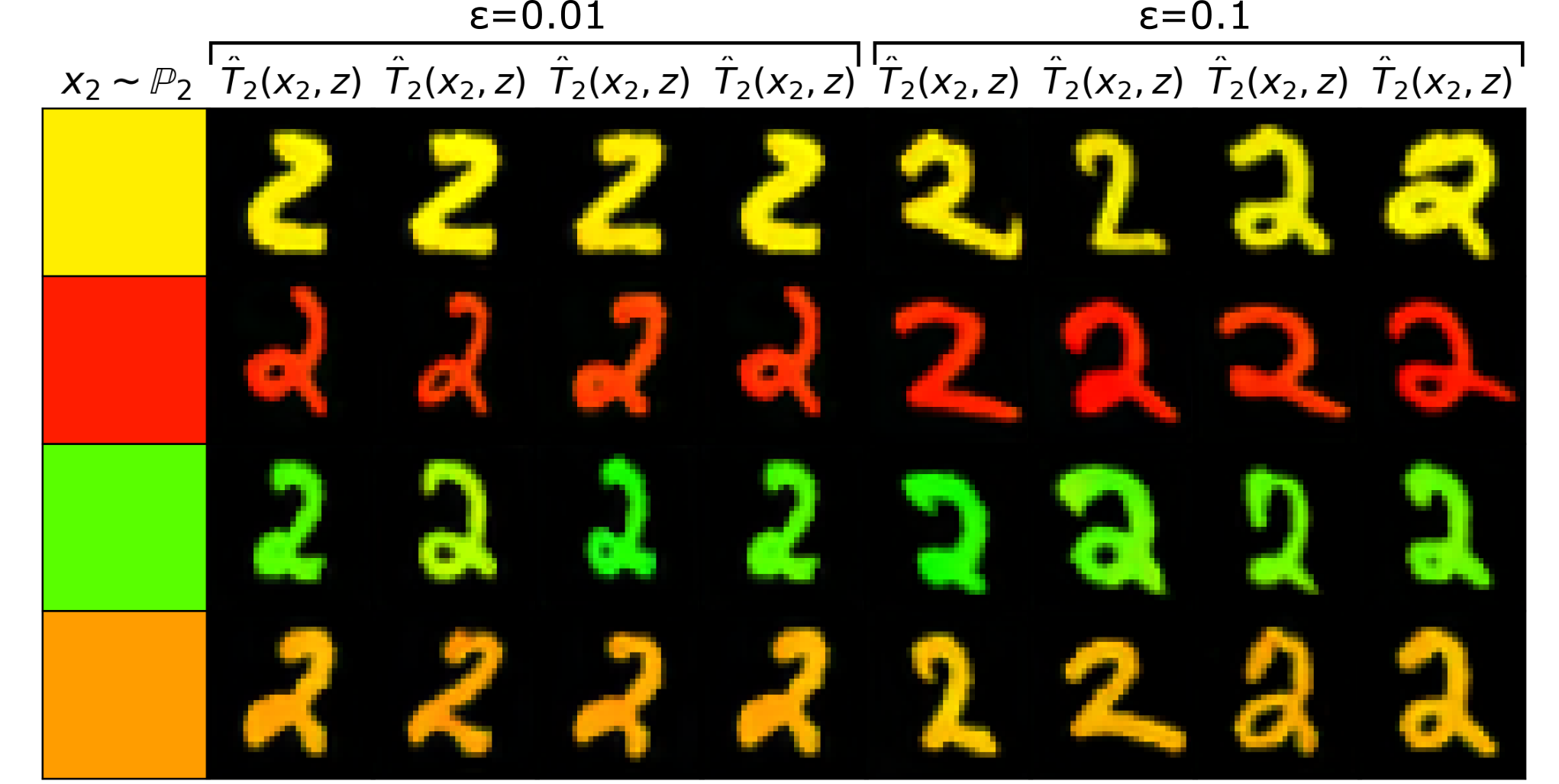}
         \caption{\centering Source ($x_2 \sim \bbP_2$) and transformed ($y = T_2(x_2, z)$) samples \newline for input $\bbP_2$ (second reference distribution).}
          \label{fig:color}
      \end{subfigure}
 \vspace{-1.6mm}
     \caption{\textbf{Our} learned stochastic maps to the OT barycenter in the Shape-Color experiment (\S\ref{subsec:colorshape}).}
     \label{fig:shape_color}
     \vspace{-5mm}
 \end{figure*}

\vspace{-2 mm}
\subsection{Shape-Color Experiment}\label{subsec:colorshape} 
\vspace{-1 mm}
The majority of continuous barycenter solvers \citep{korotin2022wasserstein, pmlr-v139-fan21d, noble2023treebased} work only with quadratic cost functions in data space, i.e., fit barycenters that are direct pixel-wise averages of samples from inputs $\mathbb{P}_{k}$, e.g., see Fig. \ref{fig:MNIST-0-1} in Appendix \ref{app:experimens_extended}. However, such barycenters are poorly useful in practice. Our proposed approach works for general cost functions, and we conduct an experiment that demonstrates learning \textit{more reasonable} barycenters. Below we introduce the components of our setup.

We consider $\epsilon$-KL OT barycenter problem (\ref{eq:klcost}) with weights $(\frac{1}{2}, \frac{1}{2})$ for distributions $\mathbb{P}_{1}$ and $\mathbb{P}_{2}$. We test $\epsilon=0.01, 0.1$.

\textbf{Shape distribution }($\mathbb{P}_{1}$). This distribution is composed of {\color{gray}{grayscale}} (non-colored) images of MNIST digits '2'. This is a distribution on space $\mathcal{X}_{1} \defeq [0,1]^{32 \times 32}$ of images.

\textbf{Color distribution }($\mathbb{P}_{2}$). This is a distribution of HSV (hue, saturation, value) vectors (colors) on $\mathcal{X}_{2} = [0,1]^{3}$. Saturation and value are set to 1 while hue $\sim$ $U[0, 0.5]$. Thus, $\mathbb{P}_{2}$ is composed of {\color{red}red}, {\color{yellow} yellow} and {\color{green} green} colors. 

\begin{figure}[!t]
    \includegraphics[width=0.5\textwidth]{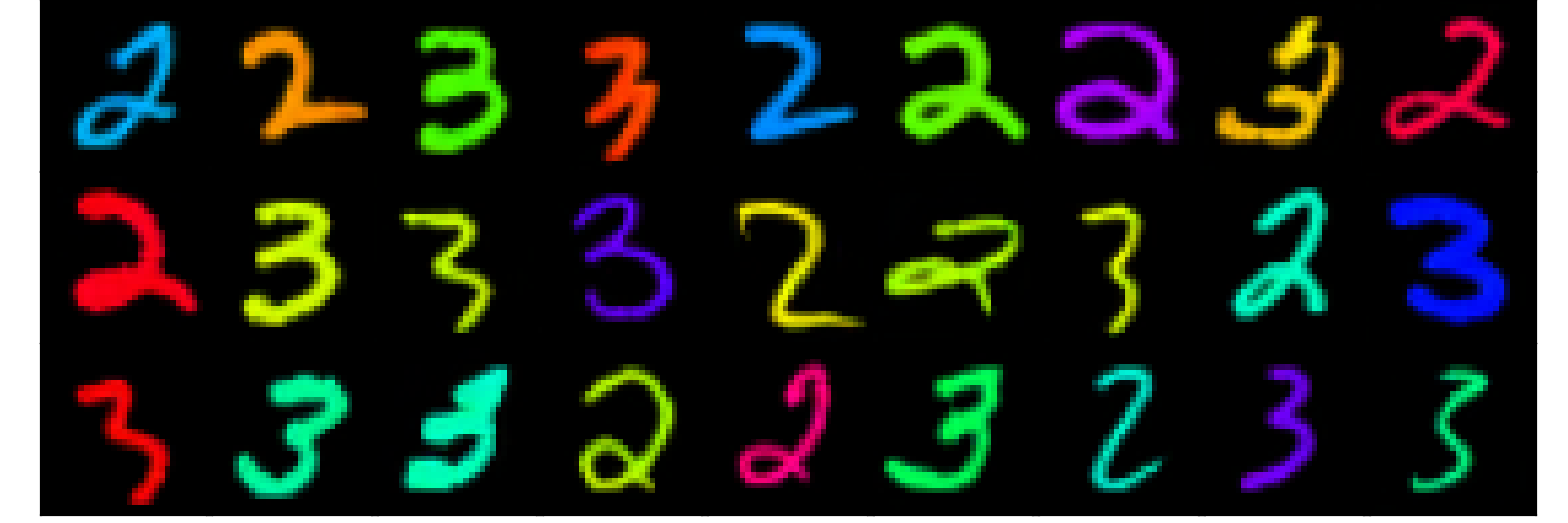}
    \caption{Samples from the StyleGAN $G$ (which represents manifold $\mathcal{M}$) trained on colored MNIST digits ``2'' \&``3''.}
    \label{fig:style_gan}
\vspace{-8 mm}
\end{figure}

\textbf{Manifold $\mathcal{M}$.} We search for the barycenter on the manifold $\mathcal{M}$ of digits "2" and "3" of \textbf{all} colors ({\color{red}red}, {\color{orange}orange}, {\color{yellow}yellow}, {\color{green}green}, {\color{cyan}cyan}, {\color{blue}blue}, {\color{purple}purple}). It is represented by the StyleGAN $G$ trained on colored images '2' and '3' (Figure \ref{fig:style_gan}).

\vspace{-2 mm}

\textbf{Transport costs.} 
The transport cost for samples from distribution $\mathbb{P}_{1}$ is $c_{1}(x_{1}, z) \defeq \frac{1}{2}||x_{1} - H_{g}(G(z))||^{2}$, where $H_{g}: \mathbb{R}^{3 \times 32 \times 32} \to \mathbb{R}^{32 \times 32}$ is {\color{gray}decolorization} operator. This cost compares shapes and does not take the color into account. The other transport cost is  $c_{2}(x_{2}, z) \defeq \frac{1}{2}||x_{2} - H_{c}(G(z))||^{2}$, where $H_{c} : \mathbb{R}^{3 \times 32 \times 32} \to \mathbb{R}^{3}$ transforms generated digit to three-dimensional HSV vector of its color, see Appendix \ref{app:experiments_details}.

\textbf{Setup summary.} Manifold $\mathcal{M}$ contains both digits "2" and "3" of all colors. Meanwhile, the barycenter of $\bbP_1$ and $\bbP_2$ is not expected to be supported by the \textit{whole} $\cM$. Indeed, it should contain only shapes as in $\bbP_1$ and only colors as in $\bbP_2$, i.e., \textit{only} {\color{red}red}, {\color{yellow} yellow} and {\color{green} green} colors.

\textbf{Results.} \textit{Our method indeed recovers the expected barycenter.} The learned map from $\mathbb{P}_{1}$ to the barycenter preserves the shape of input digit "2" but paints it with colors from  $\mathbb{P}_{2}$, see Fig. \ref{fig:shape_color}.
At the same time, the map from $\mathbb{P}_{2}$ generates images of digit "2" with a given color. Furthermore, the diversity of generated images can be controlled by the parameter $\epsilon$.

\vspace{-1 mm}
\subsection{Ave, celeba! Barycenter Benchmark Dataset}\label{subsec:ave_celeba} 

\begin{figure*}[t]
     \centering
    \begin{subfigure}[b]{0.3205\linewidth}
          \centering
          \includegraphics[width=1.07\textwidth,height=0.5027\textheight]{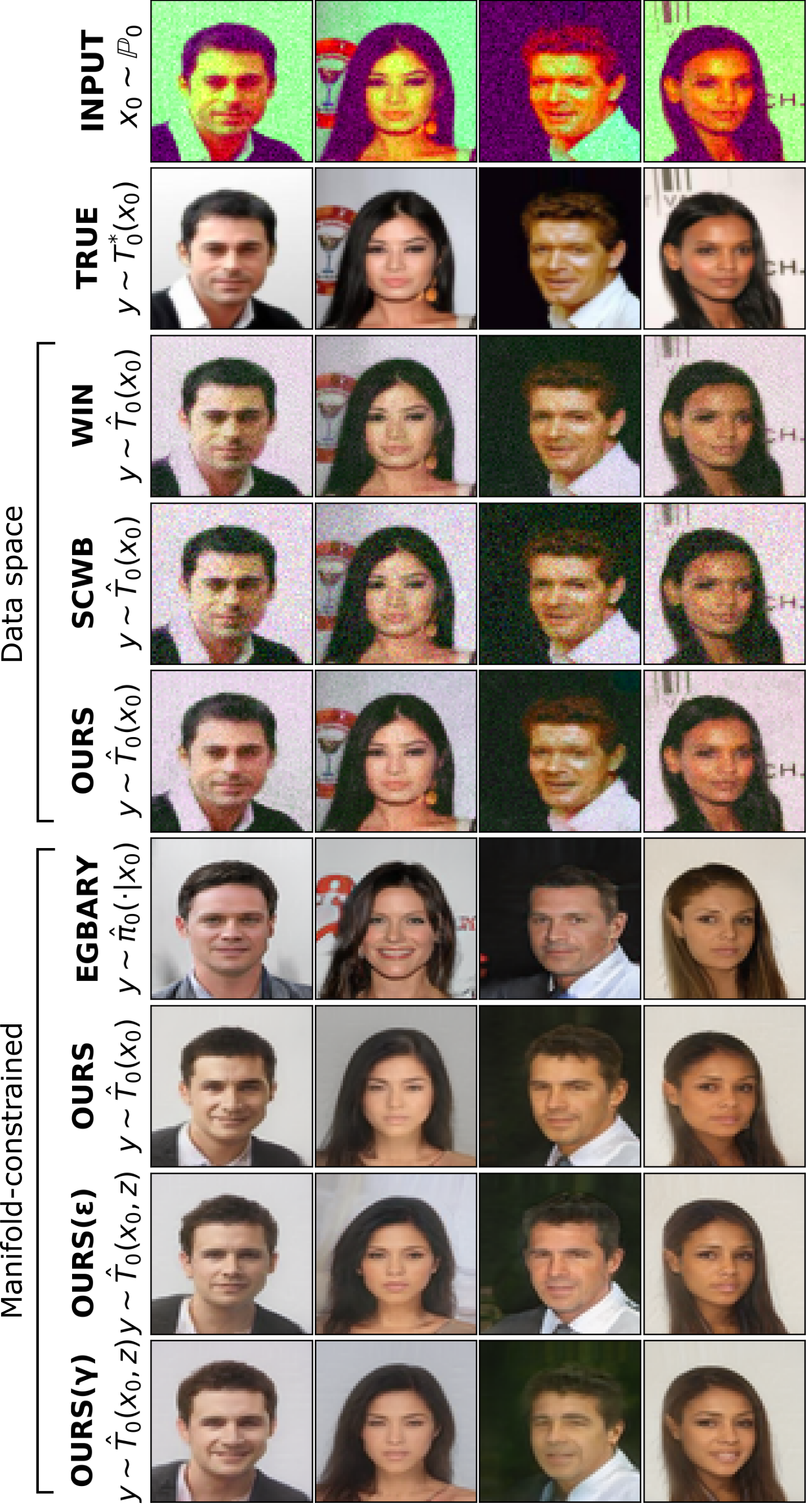}
          \caption{\centering input, transformed samples ($\bbP_0$).\newline \scaleto{\textbf{TRUE}}{5pt} shows GT mapping $\bbP_0\!\rightarrow\!\bbQ^*$.}
          \label{fig:celeba-p0}
     \end{subfigure}\hfill
      \begin{subfigure}[b]{0.3205\linewidth}   
         \centering
         \includegraphics[width=0.995\textwidth]{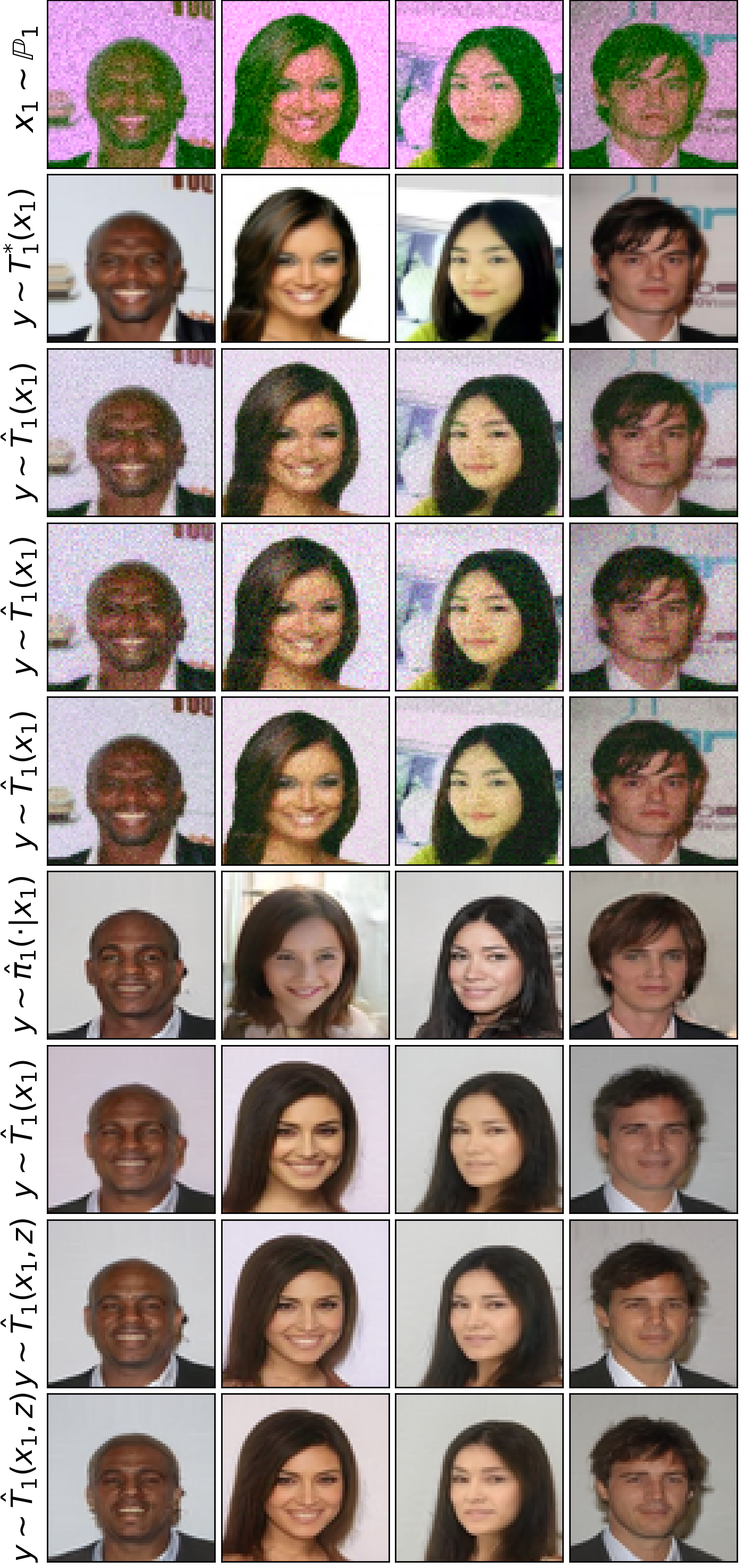}
         \caption{\centering input, transformed samples ($\bbP_1$).\newline \scaleto{\textbf{TRUE}}{5pt} shows GT mapping $\bbP_1\!\rightarrow\!\bbQ^*$.}
         \label{fig:celeba-p1}
      \end{subfigure}
      \begin{subfigure}[b]{0.3205\linewidth} 
         \centering
         \includegraphics[width=0.995\textwidth]{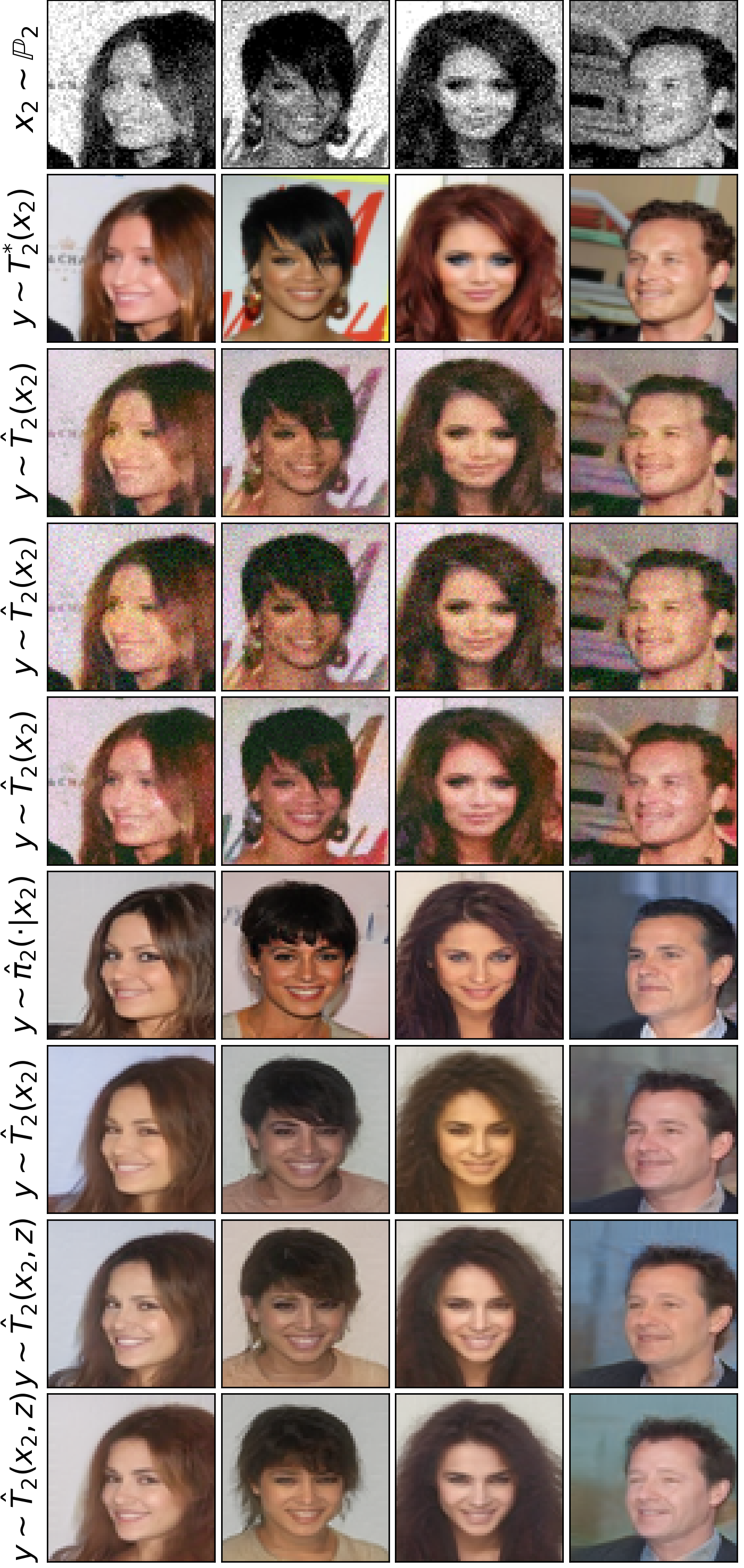}
         \caption{\centering input, transformed samples ($\bbP_2$).\newline \scaleto{\textbf{TRUE}}{5pt} shows GT mapping $\bbP_2\!\rightarrow\!\bbQ^*$.}
         \label{fig:celeba-p2}
      \end{subfigure}
 \vspace{-2mm}
     \caption{Learned (stochastic) maps to the OT barycenter by different solvers; Ave, Celeba! experiment (\S\ref{subsec:ave_celeba}).}
     \label{fig:ave-celeba-main}
 \vspace{-4mm}
 \end{figure*}

To quantitatively evaluate our proposed method, we utilize Ave, celeba! barycenter benchmark as proposed in \citep{korotin2022wasserstein}. It contains 3 distributions ($\bbP_{0}, \bbP_{1}, \bbP_{2}$) of transformed CelebA faces. These transformations are such that the barycenter $\bbQ^*$ of the reference distributions with weights $(\sfrac{1}{4}, \sfrac{1}{2}, \sfrac{1}{4})$ and classical quadratic cost functions coincides with CelebA distribution. Moreover, the ground truth quadratic OT maps ($T_0^*, T_1^*, T_2^*$) from the reference distributions to $\bbQ^*$ are also known by construction. All that remains is to compare the recovered barycenter and mappings with the true ones.

In our experiments, we consider both conventional data-space setup and \textit{manifold-constrained} setup. In the latter case, similar to \citep{kolesov2023energy}, we utilize StyleGAN generator trained on CelebA dataset. The \underline{extended information} on the training/performance of our method and baselines on Ave, celeba! benchmark can be found in Appendix~\ref{app:ave-celeba-ext}.

\begin{table}[]
\centering \scriptsize
\begin{tabular}{|x{10mm}|x{22mm}|x{2mm}|x{2mm}|x{2mm}|}
\hline
\multirow{2}{*}{\textbf{Space}}      & \multirow{2}{*}{\textbf{Solver}} & \multicolumn{3}{l|}{\textbf{FID}$\downarrow$ }                                      \\ \cline{3-5} 
                            &                         & \multicolumn{1}{l|}{$k\!=\!1$} & \multicolumn{1}{l|}{$k\!=\!2$} &\multicolumn{1}{l|} {$k\!=\!3$}  \\ 
                            
                            \hline
\multirow{3}{*}{\makecell{Data\\space}} & SCWB & \multicolumn{1}{l|}{56.7}    & \multicolumn{1}{l|}{53.2}    & \multicolumn{1}{l|}{58.8}    \\ \cline{2-5} 
                            & WIN & \multicolumn{1}{l|}{49.3}    & \multicolumn{1}{l|}{46.9}    & \multicolumn{1}{l|}{61.5}   \\ \cline{2-5} 
                            & OURS & \multicolumn{1}{l|}{\textbf{39.0}}    & \multicolumn{1}{l|}{\textbf{38.6}}    & \multicolumn{1}{l|}{\textbf{39.8}}    \\ 
                            
                            \hline
\multirow{4}{*}{\makecell{Mani-\\fold}}    & EgBary  & \multicolumn{1}{l|}{\textbf{8.4}}   & \multicolumn{1}{l|}{\textbf{8.7}}    & \multicolumn{1}{l|}{\textbf{10.2}}    \\ \cline{2-5} 
                            & OURS                   & \multicolumn{1}{l|} 
                            {30.7}    & \multicolumn{1}{l|}{31.0}    & \multicolumn{1}{l|}{31.7}   \\ \cline{2-5} 
                            & OURS ($\epsilon$)                   & \multicolumn{1}{l|}{34.5}    & \multicolumn{1}{l|}{34.9}    & \multicolumn{1}{l|}{35.7}    \\ \cline{2-5} 
                            & OURS ($\gamma$)                   & \multicolumn{1}{l|}{38.3}    & \multicolumn{1}{l|}{37.8}    & \multicolumn{1}{l|}{37.6}    \\ \hline
\end{tabular}
\caption{Quantitative comparison of barycenter solvers on the Ave, celeba! benchmark dataset.}
\label{table-fid-ave-celeba}
\vspace{-7mm}
\end{table}

\textbf{Data-space.} Our competitors are: WIN \citep{korotin2022wasserstein}, SCWB \citep{pmlr-v139-fan21d}. These methods are good baselines for continuous OT barycenter setup.
Similar to the competitors, we use unregularized cost functions. From Table \ref{table-fid-ave-celeba} we see that our method achieves \textbf{the best} FID. 

\textbf{Manifold-constrained.} We run our method with both classic \eqref{eq:strongcost} and regularized \eqref{eq:klcost}, \eqref{eq:energycost} cost functions, see the samples from recovered barycenters in Fig. \ref{fig:ave-celeba-main}. Additional samples for different regularization strengths $\gamma$, $\epsilon$ are in the Appendix, Figs. \ref{fig:ave-celeba-canvas-ent}, \ref{fig:ave-celeba-canvas-ker}. The competitive method is EgBary \citep{kolesov2023energy}, it demonstrates better results than ours, see Table \ref{table-fid-ave-celeba}. We leave the detailed analysis of this fact to future research, but note that EgBary utilizes MCMC (namely, Langevin sampling) at training and inference. Probably, this procedure better suits latent-space setups, but is rather time-consuming. For completeness, we place a \underline{detailed experimental comparison} of our method and EgBary in Appendix~\ref{app:celeba-ext:our-vs_egbary}.

\vspace{- 3mm}
\section{Discussion}\label{sec:discussion}
\vspace{-1 mm}

\textbf{Potential impact.} 
The main feature of our new OT barycenter solver is flexibility. Specifically, it is adjustable to different variants of the barycenter problem and can recover both deterministic and stochastic transport plans between data distributions. At the same time, we do not use exotic computational procedures: the core of our algorithm is a conventional bi-level adversarial game. Due to these properties, our method seems to be a good candidate to become a \textit{standard} tool for solving the OT barycenter problem. 

\vspace{-1 mm}
Another impactful byproduct of our paper is a new promising practical setup (\S\ref{subsec:colorshape}). It provides the ability to create barycenter distributions with desired characteristics by proper selection of \textit{non-Euclidean} cost functions. We believe that this finding will be useful in industrial tasks. 

\vspace{-1 mm}
\textbf{Limitations.}
\textbf{(a)} When dealing with classical cost functions, our approach is not guaranteed to recover OT plans $\pi_k^*$. In particular, the quality bounds for the recovered solutions in this case were established only under specific assumptions, see our Theorem~\ref{thm:duality_gaps}, Statement 1. \textbf{(b)} The utilization of Gaussian models (\S\ref{subsec:stoch_maps}) for transport plan parameterization is a simplification and may result in biased solutions. \textbf{(c)} Adversarial optimization procedures like those used in our paper may be prone to instabilities and necessitate hyperparameter tuning. Nevertheless, our method seems to work well for different costs and parameterizations (\S\ref{sec:experiments}). We conjecture that the concerns from above are avoidable or not so important in practice.

\section*{Acknowledgements} Skoltech was supported by the Analytical center under the RF Government (subsidy agreement 000000D730321P5Q0002, Grant No. 70-2021-00145 02.11.2021).

\section*{Impact statement} 

This paper presents work whose goal is to advance the field of Machine Learning, namely, generative modeling and computational optimal transport. Potential broader impact of our research is the same as that of most of the other generative modeling researches. In fact, there are many potential societal consequences of our work, none of which we feel must be specifically highlighted here.

\bibliography{bibliography}

\begin{thebibliography}{52}
\providecommand{\natexlab}[1]{#1}
\providecommand{\url}[1]{\texttt{#1}}
\expandafter\ifx\csname urlstyle\endcsname\relax
  \providecommand{\doi}[1]{doi: #1}\else
  \providecommand{\doi}{doi: \begingroup \urlstyle{rm}\Url}\fi

\bibitem[Agueh \& Carlier(2011)Agueh and Carlier]{agueh2011barycenters}
Agueh, M. and Carlier, G.
\newblock Barycenters in the wasserstein space.
\newblock \emph{SIAM Journal on Mathematical Analysis}, 43\penalty0 (2):\penalty0 904--924, 2011.

\bibitem[{\'A}lvarez-Esteban et~al.(2016){\'A}lvarez-Esteban, Del~Barrio, Cuesta-Albertos, and Matr{\'a}n]{alvarez2016fixed}
{\'A}lvarez-Esteban, P.~C., Del~Barrio, E., Cuesta-Albertos, J., and Matr{\'a}n, C.
\newblock A fixed-point approach to barycenters in wasserstein space.
\newblock \emph{Journal of Mathematical Analysis and Applications}, 441\penalty0 (2):\penalty0 744--762, 2016.

\bibitem[Amos et~al.(2017)Amos, Xu, and Kolter]{amos2017input}
Amos, B., Xu, L., and Kolter, J.~Z.
\newblock Input convex neural networks.
\newblock In \emph{International Conference on Machine Learning}, pp.\  146--155. PMLR, 2017.

\bibitem[Anderes et~al.(2016)Anderes, Borgwardt, and Miller]{anderes2016discrete}
Anderes, E., Borgwardt, S., and Miller, J.
\newblock Discrete wasserstein barycenters: Optimal transport for discrete data.
\newblock \emph{Mathematical Methods of Operations Research}, 84:\penalty0 389--409, 2016.

\bibitem[Asadulaev et~al.(2024)Asadulaev, Korotin, Egiazarian, Mokrov, and Burnaev]{asadulaev2024neural}
Asadulaev, A., Korotin, A., Egiazarian, V., Mokrov, P., and Burnaev, E.
\newblock Neural optimal transport with general cost functionals.
\newblock In \emph{The Twelfth International Conference on Learning Representations}, 2024.
\newblock URL \url{https://openreview.net/forum?id=gIiz7tBtYZ}.

\bibitem[Backhoff-Veraguas et~al.(2019)Backhoff-Veraguas, Beiglb{\"o}ck, and Pammer]{backhoff2019existence}
Backhoff-Veraguas, J., Beiglb{\"o}ck, M., and Pammer, G.
\newblock Existence, duality, and cyclical monotonicity for weak transport costs.
\newblock \emph{Calculus of Variations and Partial Differential Equations}, 58\penalty0 (6):\penalty0 203, 2019.

\bibitem[Bertsekas \& Shreve(1996)Bertsekas and Shreve]{bertsekas1996stochastic}
Bertsekas, D. and Shreve, S.~E.
\newblock \emph{Stochastic optimal control: the discrete-time case}, volume~5.
\newblock Athena Scientific, 1996.

\bibitem[Blondel et~al.(2018)Blondel, Seguy, and Rolet]{blondel2018smooth}
Blondel, M., Seguy, V., and Rolet, A.
\newblock Smooth and sparse optimal transport.
\newblock In \emph{International conference on artificial intelligence and statistics}, pp.\  880--889. PMLR, 2018.

\bibitem[Cazelles et~al.(2021)Cazelles, Tobar, and Fontbona]{cazelles2021novel}
Cazelles, E., Tobar, F., and Fontbona, J.
\newblock A novel notion of barycenter for probability distributions based on optimal weak mass transport.
\newblock \emph{Advances in Neural Information Processing Systems}, 34:\penalty0 13575--13586, 2021.

\bibitem[Chi et~al.(2023)Chi, Yang, Li, Ouyang, and Guan]{chi2023variational}
Chi, J., Yang, Z., Li, X., Ouyang, J., and Guan, R.
\newblock Variational wasserstein barycenters with c-cyclical monotonicity regularization.
\newblock In \emph{Proceedings of the AAAI Conference on Artificial Intelligence}, volume~37, pp.\  7157--7165, 2023.

\bibitem[Choi et~al.(2023)Choi, Choi, and Kang]{choi2023generative}
Choi, J., Choi, J., and Kang, M.
\newblock Generative modeling through the semi-dual formulation of unbalanced optimal transport.
\newblock In \emph{Thirty-seventh Conference on Neural Information Processing Systems}, 2023.
\newblock URL \url{https://openreview.net/forum?id=7WQt1J13ex}.

\bibitem[Cuturi(2013)]{cuturi2013sinkhorn}
Cuturi, M.
\newblock Sinkhorn distances: Lightspeed computation of optimal transport.
\newblock \emph{Advances in neural information processing systems}, 26, 2013.

\bibitem[Cuturi \& Doucet(2014)Cuturi and Doucet]{cuturi2014fast}
Cuturi, M. and Doucet, A.
\newblock Fast computation of wasserstein barycenters.
\newblock In \emph{International conference on machine learning}, pp.\  685--693. PMLR, 2014.

\bibitem[Daniels et~al.(2021)Daniels, Maunu, and Hand]{daniels2021score}
Daniels, M., Maunu, T., and Hand, P.
\newblock Score-based generative neural networks for large-scale optimal transport.
\newblock \emph{Advances in neural information processing systems}, 34:\penalty0 12955--12965, 2021.

\bibitem[De~Lara et~al.(2021)De~Lara, Gonz{\'a}lez-Sanz, and Loubes]{de2021consistent}
De~Lara, L., Gonz{\'a}lez-Sanz, A., and Loubes, J.-M.
\newblock A consistent extension of discrete optimal transport maps for machine learning applications.
\newblock \emph{arXiv preprint arXiv:2102.08644}, 2021.

\bibitem[Fan et~al.(2021)Fan, Taghvaei, and Chen]{pmlr-v139-fan21d}
Fan, J., Taghvaei, A., and Chen, Y.
\newblock Scalable computations of wasserstein barycenter via input convex neural networks.
\newblock In Meila, M. and Zhang, T. (eds.), \emph{Proceedings of the 38th International Conference on Machine Learning}, volume 139 of \emph{Proceedings of Machine Learning Research}, pp.\  1571--1581. PMLR, 18--24 Jul 2021.
\newblock URL \url{https://proceedings.mlr.press/v139/fan21d.html}.

\bibitem[Fan et~al.(2023)Fan, Liu, Ma, Zhou, and Chen]{fan2023neural}
Fan, J., Liu, S., Ma, S., Zhou, H.-M., and Chen, Y.
\newblock Neural monge map estimation and its applications.
\newblock \emph{Transactions on Machine Learning Research}, 2023.
\newblock ISSN 2835-8856.
\newblock URL \url{https://openreview.net/forum?id=2mZSlQscj3}.
\newblock Featured Certification.

\bibitem[Gazdieva et~al.(2022)Gazdieva, Rout, Korotin, Kravchenko, Filippov, and Burnaev]{gazdieva2022optimal}
Gazdieva, M., Rout, L., Korotin, A., Kravchenko, A., Filippov, A., and Burnaev, E.
\newblock An optimal transport perspective on unpaired image super-resolution.
\newblock \emph{arXiv preprint arXiv:2202.01116}, 2022.

\bibitem[Gozlan et~al.(2017)Gozlan, Roberto, Samson, and Tetali]{gozlan2017kantorovich}
Gozlan, N., Roberto, C., Samson, P.-M., and Tetali, P.
\newblock Kantorovich duality for general transport costs and applications.
\newblock \emph{Journal of Functional Analysis}, 273\penalty0 (11):\penalty0 3327--3405, 2017.

\bibitem[Gretton et~al.(2012)Gretton, Borgwardt, Rasch, Sch{\"o}lkopf, and Smola]{gretton2012kernel}
Gretton, A., Borgwardt, K.~M., Rasch, M.~J., Sch{\"o}lkopf, B., and Smola, A.
\newblock A kernel two-sample test.
\newblock \emph{The Journal of Machine Learning Research}, 13\penalty0 (1):\penalty0 723--773, 2012.

\bibitem[Gushchin et~al.(2023)Gushchin, Kolesov, Korotin, Vetrov, and Burnaev]{gushchin2023entropic}
Gushchin, N., Kolesov, A., Korotin, A., Vetrov, D., and Burnaev, E.
\newblock Entropic neural optimal transport via diffusion processes.
\newblock In \emph{Advances in Neural Information Processing Systems}, 2023.

\bibitem[Henry-Labordere(2019)]{henry2019martingale}
Henry-Labordere, P.
\newblock (martingale) optimal transport and anomaly detection with neural networks: A primal-dual algorithm.
\newblock \emph{arXiv preprint arXiv:1904.04546}, 2019.

\bibitem[Kantorovitch(1958)]{kantorovitch1958translocation}
Kantorovitch, L.
\newblock On the translocation of masses.
\newblock \emph{Management science}, 5\penalty0 (1):\penalty0 1--4, 1958.

\bibitem[Kechris(2012)]{kechris2012classical}
Kechris, A.
\newblock \emph{Classical descriptive set theory}, volume 156.
\newblock Springer Science \& Business Media, 2012.

\bibitem[Kingma \& Welling(2014)Kingma and Welling]{kingma2014auto}
Kingma, D.~P. and Welling, M.
\newblock Auto-encoding variational bayes.
\newblock In Bengio, Y. and LeCun, Y. (eds.), \emph{2nd International Conference on Learning Representations, {ICLR} 2014, Banff, AB, Canada, April 14-16, 2014, Conference Track Proceedings}, 2014.
\newblock URL \url{http://arxiv.org/abs/1312.6114}.

\bibitem[Klebanov et~al.(2005)Klebanov, Bene{\v{s}}, and Saxl]{klebanov2005n}
Klebanov, L.~B., Bene{\v{s}}, V., and Saxl, I.
\newblock \emph{N-distances and their applications}.
\newblock Charles University in Prague, the Karolinum Press Prague, Czech Republic, 2005.

\bibitem[Kolesov et~al.(2023)Kolesov, Mokrov, Udovichenko, Gazdieva, Pammer, Burnaev, and Korotin]{kolesov2023energy}
Kolesov, A., Mokrov, P., Udovichenko, I., Gazdieva, M., Pammer, G., Burnaev, E., and Korotin, A.
\newblock Energy-guided continuous entropic barycenter estimation for general costs.
\newblock \emph{ArXiv}, abs/2310.01105, 2023.
\newblock URL \url{https://api.semanticscholar.org/CorpusID:263605771}.

\bibitem[Korotin et~al.(2021{\natexlab{a}})Korotin, Egiazarian, Asadulaev, Safin, and Burnaev]{korotin2021wasserstein}
Korotin, A., Egiazarian, V., Asadulaev, A., Safin, A., and Burnaev, E.
\newblock Wasserstein-2 generative networks.
\newblock In \emph{International Conference on Learning Representations}, 2021{\natexlab{a}}.
\newblock URL \url{https://openreview.net/forum?id=bEoxzW_EXsa}.

\bibitem[Korotin et~al.(2021{\natexlab{b}})Korotin, Li, Genevay, Solomon, Filippov, and Burnaev]{korotin2021neural}
Korotin, A., Li, L., Genevay, A., Solomon, J.~M., Filippov, A., and Burnaev, E.
\newblock Do neural optimal transport solvers work? a continuous wasserstein-2 benchmark.
\newblock \emph{Advances in Neural Information Processing Systems}, 34:\penalty0 14593--14605, 2021{\natexlab{b}}.

\bibitem[Korotin et~al.(2021{\natexlab{c}})Korotin, Li, Solomon, and Burnaev]{korotin2021continuous}
Korotin, A., Li, L., Solomon, J., and Burnaev, E.
\newblock Continuous wasserstein-2 barycenter estimation without minimax optimization.
\newblock In \emph{International Conference on Learning Representations}, 2021{\natexlab{c}}.
\newblock URL \url{https://openreview.net/forum?id=3tFAs5E-Pe}.

\bibitem[Korotin et~al.(2022)Korotin, Egiazarian, Li, and Burnaev]{korotin2022wasserstein}
Korotin, A., Egiazarian, V., Li, L., and Burnaev, E.
\newblock Wasserstein iterative networks for barycenter estimation.
\newblock In Oh, A.~H., Agarwal, A., Belgrave, D., and Cho, K. (eds.), \emph{Advances in Neural Information Processing Systems}, 2022.
\newblock URL \url{https://openreview.net/forum?id=GiEnzxTnaMN}.

\bibitem[Korotin et~al.(2023{\natexlab{a}})Korotin, Selikhanovych, and Burnaev]{korotin2023kernel}
Korotin, A., Selikhanovych, D., and Burnaev, E.
\newblock Kernel neural optimal transport.
\newblock In \emph{The Eleventh International Conference on Learning Representations}, 2023{\natexlab{a}}.
\newblock URL \url{https://openreview.net/forum?id=Zuc_MHtUma4}.

\bibitem[Korotin et~al.(2023{\natexlab{b}})Korotin, Selikhanovych, and Burnaev]{korotin2023neural}
Korotin, A., Selikhanovych, D., and Burnaev, E.
\newblock Neural optimal transport.
\newblock In \emph{The Eleventh International Conference on Learning Representations}, 2023{\natexlab{b}}.
\newblock URL \url{https://openreview.net/forum?id=d8CBRlWNkqH}.

\bibitem[Lacombe et~al.(2023)Lacombe, Digne, Courty, and Bonneel]{lacombe2023learning}
Lacombe, J., Digne, J., Courty, N., and Bonneel, N.
\newblock Learning to generate wasserstein barycenters.
\newblock \emph{Journal of Mathematical Imaging and Vision}, 65\penalty0 (2):\penalty0 354--370, 2023.

\bibitem[Li et~al.(2020)Li, Genevay, Yurochkin, and Solomon]{li2020continuous}
Li, L., Genevay, A., Yurochkin, M., and Solomon, J.~M.
\newblock Continuous regularized wasserstein barycenters.
\newblock \emph{Advances in Neural Information Processing Systems}, 33:\penalty0 17755--17765, 2020.

\bibitem[Likmeta et~al.(2023)Likmeta, Sacco, Metelli, and Restelli]{likmeta2023wasserstein}
Likmeta, A., Sacco, M., Metelli, A.~M., and Restelli, M.
\newblock Wasserstein actor-critic: Directed exploration via optimism for continuous-actions control.
\newblock \emph{Proceedings of the AAAI Conference on Artificial Intelligence}, 37\penalty0 (7):\penalty0 8782--8790, Jun. 2023.
\newblock \doi{10.1609/aaai.v37i7.26056}.
\newblock URL \url{https://ojs.aaai.org/index.php/AAAI/article/view/26056}.

\bibitem[Makkuva et~al.(2020)Makkuva, Taghvaei, Oh, and Lee]{makkuva2020optimal}
Makkuva, A., Taghvaei, A., Oh, S., and Lee, J.
\newblock Optimal transport mapping via input convex neural networks.
\newblock In \emph{International Conference on Machine Learning}, pp.\  6672--6681. PMLR, 2020.

\bibitem[Mokrov et~al.(2024)Mokrov, Korotin, Kolesov, Gushchin, and Burnaev]{mokrov2024energyguided}
Mokrov, P., Korotin, A., Kolesov, A., Gushchin, N., and Burnaev, E.
\newblock Energy-guided entropic neural optimal transport.
\newblock In \emph{The Twelfth International Conference on Learning Representations}, 2024.
\newblock URL \url{https://openreview.net/forum?id=d6tUsZeVs7}.

\bibitem[Monge(1781)]{monge1781memoire}
Monge, G.
\newblock M{\'e}moire sur la th{\'e}orie des d{\'e}blais et des remblais.
\newblock \emph{Mem. Math. Phys. Acad. Royale Sci.}, pp.\  666--704, 1781.

\bibitem[Mroueh(2020)]{mroueh2020wasserstein}
Mroueh, Y.
\newblock Wasserstein style transfer.
\newblock In Chiappa, S. and Calandra, R. (eds.), \emph{Proceedings of the Twenty Third International Conference on Artificial Intelligence and Statistics}, volume 108 of \emph{Proceedings of Machine Learning Research}, pp.\  842--852. PMLR, 26--28 Aug 2020.
\newblock URL \url{https://proceedings.mlr.press/v108/mroueh20a.html}.

\bibitem[Noble et~al.(2023)Noble, Bortoli, Doucet, and Durmus]{noble2023treebased}
Noble, M., Bortoli, V.~D., Doucet, A., and Durmus, A.
\newblock Tree-based diffusion schr\"odinger bridge with applications to wasserstein barycenters.
\newblock In \emph{Thirty-seventh Conference on Neural Information Processing Systems}, 2023.
\newblock URL \url{https://openreview.net/forum?id=H2SuXHbFIn}.

\bibitem[Nutz(2021)]{nutz2021introduction}
Nutz, M.
\newblock Introduction to entropic optimal transport.
\newblock \emph{Lecture notes, Columbia University}, 2021.

\bibitem[Peyr{\'e} et~al.(2019)Peyr{\'e}, Cuturi, et~al.]{peyre2019computational}
Peyr{\'e}, G., Cuturi, M., et~al.
\newblock Computational optimal transport: With applications to data science.
\newblock \emph{Foundations and Trends{\textregistered} in Machine Learning}, 11\penalty0 (5-6):\penalty0 355--607, 2019.

\bibitem[Rout et~al.(2022)Rout, Korotin, and Burnaev]{rout2022generative}
Rout, L., Korotin, A., and Burnaev, E.
\newblock Generative modeling with optimal transport maps.
\newblock In \emph{International Conference on Learning Representations}, 2022.
\newblock URL \url{https://openreview.net/forum?id=5JdLZg346Lw}.

\bibitem[Santambrogio(2015)]{santambrogio2015optimal}
Santambrogio, F.
\newblock Optimal transport for applied mathematicians.
\newblock \emph{Birk{\"a}user, NY}, 55\penalty0 (58-63):\penalty0 94, 2015.

\bibitem[Seguy et~al.(2018)Seguy, Damodaran, Flamary, Courty, Rolet, and Blondel]{seguy2018large}
Seguy, V., Damodaran, B.~B., Flamary, R., Courty, N., Rolet, A., and Blondel, M.
\newblock Large scale optimal transport and mapping estimation.
\newblock In \emph{International Conference on Learning Representations}, 2018.
\newblock URL \url{https://openreview.net/forum?id=B1zlp1bRW}.

\bibitem[Sejdinovic et~al.(2013)Sejdinovic, Sriperumbudur, Gretton, and Fukumizu]{sejdinovic2013equivalence}
Sejdinovic, D., Sriperumbudur, B., Gretton, A., and Fukumizu, K.
\newblock Equivalence of distance-based and rkhs-based statistics in hypothesis testing.
\newblock \emph{The annals of statistics}, pp.\  2263--2291, 2013.

\bibitem[Singh \& Jaggi(2020)Singh and Jaggi]{singh2020model}
Singh, S.~P. and Jaggi, M.
\newblock Model fusion via optimal transport.
\newblock \emph{Advances in Neural Information Processing Systems}, 33:\penalty0 22045--22055, 2020.

\bibitem[Sion(1958)]{sion58general}
Sion, M.
\newblock On general minimax theorems.
\newblock \emph{Pacific Journal of Mathematics}, 8\penalty0 (1):\penalty0 171--176, 1958.

\bibitem[Solomon et~al.(2015)Solomon, De~Goes, Peyr{\'e}, Cuturi, Butscher, Nguyen, Du, and Guibas]{solomon2015convolutional}
Solomon, J., De~Goes, F., Peyr{\'e}, G., Cuturi, M., Butscher, A., Nguyen, A., Du, T., and Guibas, L.
\newblock Convolutional wasserstein distances: Efficient optimal transportation on geometric domains.
\newblock \emph{ACM Transactions on Graphics (ToG)}, 34\penalty0 (4):\penalty0 1--11, 2015.

\bibitem[Srivastava et~al.(2018)Srivastava, Li, and Dunson]{srivastava2018scalable}
Srivastava, S., Li, C., and Dunson, D.~B.
\newblock Scalable bayes via barycenter in wasserstein space.
\newblock \emph{The Journal of Machine Learning Research}, 19\penalty0 (1):\penalty0 312--346, 2018.

\bibitem[Tsybakov(2009)]{tsybakov2009introduction}
Tsybakov, A.~B.
\newblock \emph{Introduction to Nonparametric Estimation}.
\newblock Springer New York, NY, 2009.
\newblock \doi{https://doi.org/10.1007/b13794}.

\end{thebibliography}
\bibliographystyle{icml2024}

\newpage
\appendix
\onecolumn

\section{Proofs}\label{app:proofs}

In this section, we give the proofs for our Theorems in \S\ref{sec:method}. In our calculations and formulas below, we sometimes substitute the expectations written as $\mathbb E_{{\cdot \sim \mu}} (*)$ with their integral representations $\int (*) d \mu(\cdot)$. This is done for ease of comprehension. In the last subsection (\S \ref{app:intuitive-derivation}), we give a high-level conceptual derivation of our bi-level objective \eqref{eq:weakbary_maxmin}.

\subsection{Proof of Theorem \ref{thm:our_maxmin_solves_weakbary}}
\begin{proof}
    Before jumping into the proof, observe that any coupling in $\pi \in \Pi(\mathbb P_k)$ can be written, by the disintegration theorem, as $\pi(dx,dy) = \mathbb P_k(dx) \, K(x;dy)$, where $K : \mathcal X_k \to \mathcal P(\mathcal Y)$ is a measurable function.
    Therefore, we have
    \begin{align} \nonumber
        \inf_{\pi_k \in \Pi(\mathbb P_k)} \!\!\int\limits_{\cX_k \times \cY} \!\!\!\!\big\{C_k(x_k, \pi_k(\cdot | x_k )) - f_k(y)\big\} \, d\pi_k(x_k,y) &=
        \inf_{ \substack{K : \mathcal X_k \to \mathcal P(\mathcal Y) \\ \text{measurable} } } \int\limits_{\cX_k} \bigg\{ C_k(x_k,K(x_k)) - \int\limits_{\cY} f_k(y) \, K(x_k;dy) \bigg\} \, d \mathbb P_k(x_k) 
        \\ \label{eq:swap}
        &= \int\limits_{\cX_k} \inf_{\mu \in \mathcal P(\mathcal Y)} \left\{ C_k(x_k,\mu) - \int f_k(y) \, d\mu(y) \right\} \, d\mathbb P_k(x_k) 
        \\ \nonumber
        &= \mathbb E_{x_k \sim \mathbb P_k} f_k^{C_k}(x_k),
    \end{align}
    where the second equality is justified by the following reasoning:
    the map $(x_k,\mu) \mapsto C(x_k,\mu) - \int f_k(y) \, d\mu(y)$ is measurable and bounded from below, thus, we can invoke \citep[Proposition 7.27 and Proposition 7.50]{bertsekas1996stochastic} to find, for every $\epsilon > 0$, a measurable function $K^\epsilon : \mathcal X_k \to \mathcal P(\mathcal Y)$ such that
    \[
        C_k(x_k,K^\epsilon(x_k)) - \int_{\cY} f_k(y) \, dK^\epsilon(x_k;dy) \le \inf_{\mu \in \mathcal P(\mathcal Y)} C_k(x_k,\mu) - \int_{\cY} f_k(y) \, d\mu(y) + \epsilon = f^{C_k}_k(x_k) + \epsilon \quad
        \mathbb P_k\text{-a.e.\ }x_k.
    \]
    From this inequality follows readily the equality in \eqref{eq:swap}, as $\epsilon$ was arbitrary.

    Next, we turn our attention towards the barycenter problem.
    Using the weak optimal transport duality \eqref{eq:weakot_dual}, we have for every $\mathbb Q \in \mathcal P(\mathcal Y)$ that
    \begin{align*}
        \sum_{k = 1}^K \lambda_k \OT_{C_k}(\mathbb P_k, \mathbb Q) &=
        \sup_{f_{1:K} \in C(\mathcal Y)^K} \sum_{k = 1}^K \lambda_k \left( \mathbb E_{x_k \sim \mathbb P_k} f_k^{C_k}(x_k) +  \mathbb E_{y \sim \mathbb Q} f_k(y) \right).
    \end{align*}
    Notice that the map
    \[
        \mathcal P(\mathcal Y) \times \cC(\mathcal Y)^K \ni (\mathbb Q, f_{1:K}) \mapsto \sum_{k = 1}^K \lambda_k \left\{  \mathbb E_{x_k \sim \mathbb P_k} f^{C_k}_k(x_k) + \mathbb E_{y \sim \mathbb Q} f_k(y)  \right\}
    \]
    is continuous and convex in $\mathbb Q$ and, thanks to \citep[Proposition 1]{kolesov2023energy}, concave in $f_{1:K}$.
    As $\mathcal Y$ is compact, the same holds true for $\mathcal P(\mathcal Y)$, hence, we can apply Sion's minimax theorem \citep[Theorem 3.4]{sion58general}, which permits us to swap $\inf$ with $\sup$ and get
    \begin{align*}
        \mathcal L^\ast &= \inf_{\mathbb Q \in \mathcal P(\mathcal Y)} \sup_{f_{1:K} \in \cC(\mathcal Y)^K} \sum_{k = 1}^K \lambda_k \left( \mathbb E_{x_k \sim \mathbb P_k} f^{C_k}_k(x_k) + \mathbb E_{y \sim \mathbb Q} f_k(y) \right) \\
        &= \sup_{f_{1:K} \in \cC(\mathcal Y)^K} \inf_{\mathbb Q \in \mathcal P(\mathcal Y)} \sum_{k = 1}^K \lambda_k \left( \mathbb E_{x_k \sim \mathbb P_k} f_k^{C_k}(x_k) + \mathbb E_{y \sim \mathbb Q} f_k(y)  \right) \\
        &= \sup_{f_{1:K} \in \cC(\mathcal Y)^K} \left\{ m(f_{1:K}) + \sum_{k = 1}^K \lambda_k\mathbb E_{x_k \sim \mathbb P_k} f_k^{C_k}(x_k) \right\},
    \end{align*}
    where $m(f_{1:K}) \defeq \inf_{y \in \mathcal Y} \left\{ \sum_{k = 1}^K \lambda_k f_k(y) \right\}$.
    Since $(f_1 - a)^{C_1} = f_1^{C_1} + a$ for all $a \in \mathbb R$, cf.\ \citep[Proposition 1 (ii)]{kolesov2023energy}, we can translate $f_1$ by $-\frac{m(f_{1:K})}{\lambda_1}$ without changing the value.
    Consequently, 
    \begin{align*}
        \mathcal L^\ast &= \sup_{ \substack{ f_{1:K} \in \cC(\mathcal Y)^K \\ m(f_{1:K}) = 0 } } \sum_{k = 1}^K \lambda_k \mathbb E_{x_k \sim \mathbb P_k} f_k^{C_k}(x_k).
    \end{align*}
    Pick a vector $f_{1:K} \in \cC(\mathcal Y)^K$ with $m(f_{1:K}) = 0$.
    Writing $\tilde f_1 := - \sum_{k = 2}^K \frac{\lambda_k}{\lambda_1} f_k$, we have
    \[
       0 = m(f_{1:K}) \le \sum_{k = 1}^K \lambda_k f_k = f_1 - \tilde f_1.
    \]
    Due to monotonicity of the $C$-transform (cf.\ \citep[Proposition 1 (i)]{kolesov2023energy}), we find that $f_1^{C_1} \le \tilde f_1^{C_1}$.
    Thus, by replacing $(f_1,\ldots,f_K)$ with $(\tilde f_1,f_2,\ldots,f_K) \in \cC(\mathcal Y)^K$, we improve the value.
    We obtain
    \begin{align*}
        \mathcal L^\ast &= \sup_{\substack{f_{1:K} \in \cC(\mathcal Y)^K, \\ \sum_{k = 1}^K \lambda_k f_k = 0}} \sum_{k = 1}^K \lambda_k \mathbb E_{x_k \sim \mathbb P_k} f_k^{C_k}(x_k) 
        \\
        &=
        \sup_{\substack{f_{1:K} \in \cC(\mathcal Y)^K, \\ \sum_{k = 1}^K \lambda_k f_k = 0}} 
        \sum_{k = 1}^K \lambda_k \left\{ \inf_{\pi_k \in \Pi(\mathbb P_k)} \int_{\cY} \big[C_k(x_k, \pi_k(\cdot | x_k )) - f_k(y)\big] \, d\pi_k(x_k,y) \right\} \\
        &=  \sup_{\substack{f_{1:K} \in \cC(\mathcal Y)^K, \\ \sum_{k = 1}^K \lambda_k f_k = 0}}
        \inf_{\substack{\pi_k \in \Pi(\mathbb P_k), \\ (k \in \overline K)}} \mathcal V(f_{1:K}, \pi_{1:K}),
    \end{align*}
    where the second equality follows from \eqref{eq:swap}, and conclude the proof.
\end{proof}

\subsection{Proof of Theorem \ref{thm:duality_gaps}}
\begin{proof} We subsequently prove our statements for classical \eqref{eq:strongcost}, $\epsilon$-$\KL$ \eqref{eq:klcost} and $\gamma$-Energy cost functions. Our analysis of specific cost cases will follow a generic template explained below. 
For a potential $f_k \in \cC(\cY)$ and a plan $\pi_k \in \Pi(\bbP_k)$ we define:
\begin{align}
    \cV_k(f_k, \pi_k) \defeq \E{x_k \sim \bbP_k} C_k\big(x_k, \pi_k(\cdot \vert x_k)\big) - \E{x_k \sim \bbP_k}\Big(\E{y \sim \pi_k(\cdot \vert x_k)} f_k(y)\Big). \label{eq:V_func_k}
\end{align}
In what follows, for given congruent potentials $\widehat{f}_{1:K}$ and plans $\widehat{\pi}_{1:K}$, we express \eqref{eq:V_func} and \eqref{eq:L_func} with help of \eqref{eq:V_func_k}. 
The functional $\cV$, cf.\ \eqref{eq:V_func}, can be expressed by the following weighted sum:
\begin{align}
    \cV(\widehat{f}_{1:K}, \widehat{\pi}_{1:K}) = \sum_{k = 1}^{K} \lambda_k \cV_k(\widehat{f}_k, \widehat{\pi}_k). \label{eq:V_through_V_k}
\end{align}
In turn, the inner minimization problem \eqref{eq:L_func} can be split into separate optimization problems over plans $\pi_k \in \Pi(\bbP_k)$:
\begin{align}
    \cL_k(\widehat{f}_k) \defeq \inf_{\pi_k \in \Pi(\bbP_k)} \cV_k(\widehat{f}_k, \pi_k).\label{eq:L_func_k}
\end{align}
When considering the specific cases for different cost functions, we will manage to explicitly or implicitly recover the minimizers of functionals $\cV_k$:
\begin{align*}
    \pi_k^f \in \arginf_{\pi_k \in \Pi(\bbP_k)} \cV_k(\widehat{f}_k, \pi_k).
\end{align*}
The arguments regarding the existence (and possible uniqueness) of plans $\pi_k^f$ will be given correspondingly. Note that it is better to write $\pi_k^{\widehat{f}_k}$ instead of $\pi_k^f$. Nevertheless, abusing the notations, we will use exactly $\pi_k^f$ in the text.

Given $\pi_k^f \in \Pi(\bbP_k)$, equation for functional $\cL$ \eqref{eq:L_func} can be written as follows:
\begin{align}
    \cL(\widehat{f}_{1:K}) = \sum_{k = 1}^K \lambda_k \cL_k(\widehat{f}_k) = \sum_{k = 1}^{K} \lambda_k \cV_k(\widehat{f}_k, \pi_k^f). \label{eq:L_through_V_k}
\end{align}

The difference between \eqref{eq:V_through_V_k} and \eqref{eq:L_through_V_k} is exactly the first gap $\delta_1$ \eqref{eq:deltagap1}:
\begin{align}
    \delta_1 = \cV(\widehat{f}_{1:K}, \widehat{\pi}_{1:K}) - \cL(\widehat{f}_{1:K}) &= \sum_{k = 1}^{K} \lambda_k \Big\{ \cV_k(\widehat{f}_k, \widehat{\pi}_k) - \cV_k(\widehat{f}_k, \pi_k^f)\Big\} = \sum_{k = 1}^K \lambda_k \delta_{1, k}(\widehat{f}_k, \widehat{\pi}_k)\,;
    \nonumber
    \\
    \delta_{1,k}(\widehat{f}_k, \widehat{\pi}_k) &\defeq \cV_k(\widehat{f}_k, \widehat{\pi}_k) - \cV_k(\widehat{f}_k, \pi_k^f). \label{eq:deltagap1_k}
\end{align}
The detailed analysis of quantities $\delta_{1, k}(\widehat{f}_k, \widehat{\pi}_k)$ will be provided for each particular cost function case. 

Now we move on the the analysis of the second gap $\delta_2$ \eqref{eq:deltagap2} and derive the similar factorization for this quantity. Recall that the optimal value $\cL^*$ of OT barycenter problem \eqref{eq:weakbary_primal} is:
\begin{gather}
    \cL^* = \sum_{k = 1}^{K} \lambda_k \OT_{C_k}(\bbP_k, \bbQ^*) = \sum_{k = 1}^{K} \lambda_k \int\limits_{\cX_k} C_k(x_k, \pi_k^*(\cdot \vert x_k)) d \bbP_k(x_k), \label{eq:proof:gaps:9}
\end{gather}
where $\bbQ^*$ is the barycenter distribution and $\pi_k^* \in \Pi(\bbP_k, \bbQ^*)$, $k \in \overline{K}$, are (weak) OT plans between $\bbP_k$ and $\bbQ^*$. Note that the second marginal distribution of $\pi_k^*$ for each $k$ is $\bbQ^*$. Thanks to the congruence condition for potentials $\widehat{f}_{1:K}$, we have:
\begin{align}
    \sum_{k = 1}^{K} \lambda_k \int\limits_{\cX_k} \int\limits_{\cY} \widehat{f}_k(y) d \pi_k^* (y \vert x_k) d \bbP_k(x_k) =  \sum_{k = 1}^{K} \lambda_k \int\limits_{\cY} \widehat{f}_k(y) \underbrace{d (\pi_k^{*})^\cY (y)}_{ = d \bbQ^*(y)} = \int\limits_{\cY} \underbrace{\Big\{\sum_{k = 1}^{K} \lambda_k \widehat{f}_k (y) \Big\}}_{= 0} d \bbQ^*(y) = 0. \label{eq:proof:gaps:10}
\end{align}
The combination of \eqref{eq:proof:gaps:9} and \eqref{eq:proof:gaps:10} yields the following:
\begin{align*}
    \cL^* &= \sum_{k = 1}^{K} \lambda_k \int\limits_{\cX_k} C_k(x_k, \pi_k^*(\cdot \vert x_k)) d \bbP_k(x_k) - \underbrace{\sum_{k = 1}^{K} \lambda_k \int\limits_{\cX_k} \int\limits_{\cY} \widehat{f}_k(y) d \pi_k^* (y \vert x_k) d \bbP_k(x_k)}_{ = 0 \text{ from \eqref{eq:proof:gaps:10}}} \\
    &= \sum_{k = 1}^{K} \lambda_k \bigg\{\int\limits_{\cX_k} C_k\big(x_k, \pi_k^* (\cdot \vert x_k)\big) - \int\limits_{\cX_k} \int\limits_{\cY} \widehat{f}_k(y) d \pi_k^* (y \vert x_k) d \bbP_k(x_k)\bigg\} = \sum_{k = 1}^{K} \lambda_k \cV_k(\widehat{f}_k, \pi_k^*). 
\end{align*}
By summarizing the expression for $\cL^*$ and \eqref{eq:L_through_V_k} we derive the factorization for $\delta_2$:
\begin{align}
    \delta_2 =\cL^* - \cL(\widehat{f}_{1:K}) &= \sum_{k = 1}^{K} \lambda_k \Big\{ \cV_k(\widehat{f}_k, \pi_k^*) - \cV_k(\widehat{f}_k, \pi_k^f)\Big\} = \sum_{k = 1}^K \lambda_k \delta_{2, k}(\widehat{f}_k); \nonumber
    \\
    \delta_{2, k}(\widehat{f}_k) &= \cV_k(\widehat{f}_k, \pi_k^*) - \cV_k(\widehat{f}_k, \pi_k^f). \label{eq:deltagap2_k}
\end{align}

Now we are ready to consider the specific cost functions. Following the outline above, for each case we will:
\begin{itemize}
    \item Establish the existence of plans $\pi_k^f$ which optimize functionals $\cL_k$. Derive some properties of these plans needed for further analysis.
    \item Analyze the gaps $\delta_{1, k} = \cV_k(\widehat{f}_k, \widehat{\pi}_k) - \cV_k(\widehat{f}_k, \pi_k^f)$ and $\delta_{2, k} = \cV_k(\widehat{f}_k, \pi_k^*) - \cV_k(\widehat{f}_k, \pi_k^f)$. Using $\delta_{1, k}$ and $\delta_{2, k}$, upper bound some discrepancies between $\widehat{\pi}_k, \pi_k^f$ and $\pi_k^*, \pi_k^f$.
    \item Aggregate the obtained upper bounds for each $k \in \overline{K}$. Derive the ultimate bound utilizing the identity: 
    \begin{align*}
        \delta_1 + \delta_2 = \sum_{k = 1}^{K}\lambda_k (\delta_{1, k} + \delta_{2, k}).
    \end{align*}
\end{itemize}

\textit{Proof of statement 1 (classical cost function).} 
We consider cost functions $C_k$, $k \in \overline{K}$ given by
\begin{align}\label{eq:classical_cost}
    C_k(x_k, \mu) = \int_{\cY} c_k(x_k, y) d \mu(y),
\end{align}
where $x_k \in \cX_k$, $\mu \in \cP(\cY)$.
Here, the functional $\cV_k$, cf.\ \eqref{eq:V_func_k}, takes the following form
\begin{align}
    \cV_k(\widehat{f}_k, \pi_k) = \int_{\cX_k} \int_{\cY} \big\{ c_k(x_k, y) - \widehat{f}_k(y) \big\} d \pi_k(y \vert x_k) d \bbP_k(x_k). \label{eq:V_func_k_classical}
\end{align}
For convenience, we introduce the function $g_k(x_k, y) \defeq c_k(x_k, y) - \widehat{f}_k(y)$, $x_k \in \cX_k$, $y \in \cY$. 
Note that $g_k$ is $\beta$-strongly convex in the second argument by the Theorem assumption. 
Define the maps $T_k^f : \cX_k \rightarrow \cY$ as follows: 
\begin{align*}
    T_k^f(x_k) \defeq \arginf_{y \in \cY} g_k(x_k, y).
\end{align*}
Note that the $\arginf$ above exists and is unique since $\cY$ is compact and $g_k$ is strongly convex in $y$.
Furthermore, since $x \mapsto \inf_{y \in \cY} g_k(x,y)$ and $g_k$ are both continuous, we have that the graph 
\[ {\rm graph}(T^f_k) = \left\{ (x_k, y) \in \mathcal X_k \times \mathcal Y : g_k(x_k,y) = \inf_{\tilde y \in \mathcal Y} g_k(x_k,\tilde y) \right\}
\]
is a closed subset of $\mathcal X_k \times \mathcal Y$ and therefore Borel.
The latter is by \citep[Theorem 4.12]{kechris2012classical} equivalent to $T^f_k$ being Borel.
Consequently, it induces a transport plan $\pi_k^f$ via
\begin{align*}
    (\text{Id}, T_k^f)_\# \bbP_k \defeq \pi_k^f \in \Pi(\bbP_k).
\end{align*}
We continue to study the functional $\cV_k$ as defined in \eqref{eq:V_func_k_classical}. 
For any $\pi_k \in \Pi(\bbP_k)$ we have
\begin{align*}
    \cV_k(\widehat{f}_k, \pi_k) \!&=\!\! \int\limits_{\cX_k}\!%
    {\int\limits_{\cY} g_k(x_k, y) d \pi_k(y \vert x_k)}
    d \bbP_k(x_k) \!\geq\! \int\limits_{\cX_k} \!g_k(x_k, T_k^f(x_k)) d \bbP_k(x_k) \!
    \\
    &=\!\! \int\limits_{\cX_k}\!\int\limits_{\cY}\!g_k(x_k, y)d \pi_k^f(y \vert x_k) d \bbP_k(x_k) \!=\! \cV_k(\widehat{f}_k, \pi_k^f),
\end{align*}
i.e., $\pi_k^f$ indeed minimizes \eqref{eq:V_func_k_classical} over $\Pi(\mathbb P_k)$ and solves \eqref{eq:L_func_k} for classical cost functions of the form \eqref{eq:classical_cost}. 

Now we take a closer look at the gaps $\delta_{1, k}$ defined in \eqref{eq:deltagap1_k}. 
For $k \in \overline{K}$ we compute
\begin{align}
    \delta_{1, k} &= \cV_k(\widehat{f}_k, \widehat{\pi}_k) - \cV_k(\widehat{f}_k, \pi_k^f) = \int\limits_{\cX_k} \bigg\{ \int\limits_{\cY} g_k(x_k, y) d \widehat{\pi}_k(y \vert x_k) - g_k\big(x_k, T_k^f(x_k)\big)\bigg\} d \bbP_k(x_k) \nonumber \\
    &=\int\limits_{\cX_k} \int\limits_{\cY} \Big\{ g_k(x_k, y) - g_k\big(x_k, T_k^f(x_k)\big)\Big\} d \widehat{\pi}_k(y \vert x_k) d \bbP_k(x_k) \label{eq:proof:gaps:4}\\ 
    &\geq \frac{\beta}{2}\int\limits_{\cX_k} \int\limits_{\cY} \Vert y - T_k^f(x_k) \Vert^2 \underbrace{d \widehat{\pi}_k(y \vert x_k) d \bbP_k(x_k)}_{= d \widehat{\pi}_k(x_k, y)} = \frac{\beta}{2}\!\!\int\limits_{\cX_k \times \cY} \!\!\Vert y - T_k^f(x_k) \Vert^2 d \widehat{\pi}_k(x_k, y).\label{eq:delta1_classical_ineq}
\end{align}
To transition from \eqref{eq:proof:gaps:4} to \eqref{eq:delta1_classical_ineq} we utilize the inequality $\frac{\beta}{2} \Vert T_k^f(x_k) - y \Vert^2 \leq g_k(x_k, y) - g_k\big(x_k, T_k^f(x_k)\big)$, which directly follows from $\beta$-strong convexity (recall that $T_k^f(x_k)$ is the minimizer of $y \mapsto g_k(x_k, y)$).

Concerning the gaps $\delta_{2, k}$, cf.\ \eqref{eq:deltagap2_k}, we can conduct the same analysis as for $\delta_{1, k}$ (eqs. \eqref{eq:proof:gaps:4}, \eqref{eq:delta1_classical_ineq}) and derive:
\begin{align}
    \delta_{2, k} \geq \frac{\beta}{2}\!\! \int\limits_{\cX_k \times \cY} \Vert y - T_k^f(x_k)\Vert^2 d \pi_k^*(x_k, y).\label{eq:delta2_classical_ineq}
\end{align}
We are left to summarize inequalities for $\delta_{1, k}$ \eqref{eq:delta1_classical_ineq} and $\delta_{2, k}$ \eqref{eq:delta2_classical_ineq}:
\begin{align}
    \delta_{1, k} + \delta_{2, k} \geq \frac{\beta}{2} \bigg\{ \!\int\limits_{\cX_k \times \cY}\!\!\! \Vert y - T_k^f(x_k)\Vert^2 \!\!\!\!\!\!\underbrace{d \widehat{\pi}_k(x_k, y)}_{= d \widehat{\pi}_k(y \vert x_k) d \bbP_k(x_k)} + \int\limits_{\cX_k \times \cY} \!\!\!\Vert y - T_k^f(x_k)\Vert^2 \!\!\!\!\!\!\underbrace{d \pi_k^*(x_k, y)}_{d \pi^*_k( y \vert x_k) d \bbP_k(x_k)} \bigg\} 
    = \nonumber \\
    \frac{\beta}{2} \int\limits_{\cX_k} \Big\{ \int\limits_{\cY}\Vert y - T_k^f(x_k)\Vert^2 d \widehat{\pi}_k(y \vert x_k) + \int\limits_{\cY} \Vert y' - T_k^f(x_k)\Vert^2 d \pi_k^*(y' \vert x_k) \Big\} d \bbP_k(x_k) 
    = \nonumber \\
    \frac{\beta}{2} \int\limits_{\cX_k} \!\Big\{ \!\int\limits_{\cY \times \cY}\!\!\Vert y - T_k^f(x_k)\Vert^2 d \widehat{\pi}_k(y \vert x_k) \!\underbrace{d \pi_k^*(y' \vert x_k)}_{\text{integrates to 1}} + \!\int\limits_{\cY \times \cY} \!\!\Vert y' - T_k^f(x_k)\Vert^2 \!\underbrace{d \widehat{\pi}_k(y \vert x_k)}_{\text{integrates to 1}}\! d \pi_k^*(y' \vert x_k) \Big\} d \bbP_k(x_k) 
    = \nonumber \\
    \frac{\beta}{2} \int\limits_{\cX_k} \Big\{\int\limits_{\cY \times \cY} \!\!\big( \Vert y - T_k^f(x_k) \Vert^2 + \Vert y' - T_k^f(x_k) \Vert^2 \big) d \widehat{\pi}_k(y \vert x_k) d \pi_k^*(y' \vert x_k) \Big\} d \bbP_k(x_k) 
    \geq \label{eq:proof:gaps:7} \\
    \frac{\beta}{2}\int\limits_{\cX_k} \Big\{\int\limits_{\cY \times \cY} \!\! \frac{\Vert y - y'\Vert^2}{2} d \widehat{\pi}_k(y \vert x_k) d \pi_k^*(y' \vert x_k) \Big\} d \bbP_k(x_k) 
    \geq \label{eq:proof:gaps:8} \\
    \frac{\beta}{2} \int\limits_{\cX_k} \Big\{ \inf\limits_{\gamma_{x_{\scaleto{k}{2.5pt}}} \in \scaleto{\Pi}{5pt} \text{$\big($}\widehat{\pi}_k(\cdot \vert x_k), \pi_k^*(\cdot \vert x_k)\text{$\big)$}} \int\limits_{\cY \times \cY} \!\!\!\frac{\Vert y - y'\Vert^2}{2}  d \gamma_{x_k}(y, y')\Big\} d \bbP_k(x_k) = \frac{\beta}{2} \int\limits_{\cX_k} \bbW_2^2\big(\widehat{\pi}_k(\cdot\vert x_k), \pi_k^*(\cdot\vert x_k)\big) d \bbP_k(x_k) \nonumber.
\end{align}

The transition from \eqref{eq:proof:gaps:7} to \eqref{eq:proof:gaps:8} follows from simple consideration: $$\Vert y - T_k^f(x_k) \Vert^2 + \Vert y' - T_k^f(x_k) \Vert^2 \geq \sfrac{1}{2}\big(\Vert y - T_k^f(x_k) \Vert + \Vert T_k^f(x_k) - y' \Vert\big)^2 \overset{\text{Triang. ineq.}}{\geq} \sfrac{1}{2}\Vert y - y'\Vert^2.$$

Aggregating the inequalities for $\delta_{1_k} + \delta_{2, k}$ with weights $\lambda_k$, $k \in \overline{K}$ finishes the proof of statement 1:
\begin{align*}
    \delta_1 + \delta_2 = \sum_{k = 1}^{K} \lambda_k (\delta_{1, k} + \delta_{2, k}) \geq \frac{\beta}{2} \sum_{k = 1}^K \lambda_k \int_{\cX_k} \bbW_2^2\big(\widehat{\pi}_k(\cdot \vert x), \pi_k^*(\cdot\vert x)\big) d \bbP_k(x).
\end{align*}

\textit{Proof of statement 2 ($\epsilon$-$\KL$ weak cost function).} 
Recall that the $\epsilon$-$\KL$ cost functions are given by
\begin{align*}
    C_k(x_k, \mu) = \int_{\cY} c_k(x_k, y) d \mu(y) + \epsilon \KL(\mu \Vert \mu_0) = \int_{\cY} c_k(x_k, y) d \mu(y) - \epsilon H(\mu) - \epsilon \! \int_{\cY} \log \dv{\mu_0(y)}{y} d \mu(y).
\end{align*}
The term $\dv{\mu_0(y)}{y}$ denotes the density function of the distribution $\mu_0$ w.r.t.\ the Lebesgue measure at point $y$. 
The functional $\cV_k$ \eqref{eq:V_func_k} takes for a given potential $\widehat{f}_k$ and an arbitrary $\pi_k \in \Pi(\bbP_k)$ the form:
\begin{align}
    \cV_k(\widehat{f}_k, \pi_k) 
    = \nonumber \\
    \int\limits_{\cX_k} \!\!\Big\{\! \int\limits_{\cY} \!\!c_k(x_k, y) d \pi_k(y \vert x_k) \!-\! \epsilon H\!\big(\pi_k(\cdot \vert x_k)\big) \!-\! \epsilon \!\! \int\limits_{\cY} \!\log \!\scaleto{\dv{\mu_0(y)}{y}}{20pt} d \pi_k(y\vert x_k) \!\Big\} d \bbP_k(x_k) \!-\!\! \int\limits_{\cX_k}\! \int\limits_{\cY} \!\!\widehat{f}_k(y) d \pi_k(y \vert x_k) d \bbP_k(x_k) 
    =\nonumber \\
    \int\limits_{\cX_k} \!\Big\{\! \int\limits_{\cY} \!c_k(x_k, y) d \pi_k(y \vert x_k) \!-\! \epsilon H\big(\pi_k(\cdot \vert x_k)\big) \!- \! \int\limits_{\cY} \underbrace{\big[\epsilon\log \scaleto{\dv{\mu_0(y)}{y}}{20pt} + \widehat{f}_k(y)\big]}_{\defeq \widetilde{f}_k(y)} d \pi_k(y\vert x_k) \Big\} d \bbP_k(x_k)
    =\nonumber \\
    \int\limits_{\cX_k} \!\Big\{\! \int\limits_{\cY} \!c_k(x_k, y) d \pi_k(y \vert x_k) \!-\! \epsilon H\big(\pi_k(\cdot \vert x_k)\big) \!- \! \int\limits_{\cY} \widetilde{f}_k(y) d \pi_k(y \vert x_k) \Big\} d \bbP_k(x_k). \label{eq:V_func_k_KL}
\end{align}

The expression under curly brackets in \eqref{eq:V_func_k_KL} appeared in previous works, see \citep[Equation 8]{mokrov2024energyguided}. Following \citep{mokrov2024energyguided}, we introduce $\cG_{x_k, \widetilde{f}_k} : \cP(\cY) \rightarrow \bbR$ as follows:
\begin{align}
    \cG_{x_k, \widetilde{f}_k}(\mu) \defeq \int\limits_{\cY} c_k(x_k, y) d \mu(y) - \epsilon H(\mu) - \int\limits_{\cY} \widetilde{f}_k(y) d \mu(y). \label{eq:G_func_energyguided}
\end{align}
The functional \eqref{eq:G_func_energyguided} is minimized by distribution $\mu_{x_k}^{\widetilde{f}_k} \in \cP(\cY)$ with density \citep[Theorem 1]{mokrov2024energyguided}:
\begin{align*}
    \dv{\mu_{x_k}^{\widetilde{f}_k}(y)}{y} = \frac{1}{Z(\widetilde{f}_k, x_k)} \exp\bigg(\frac{\widetilde{f}_k(y) - c_k(x_k, y)}{\epsilon}\bigg) = \frac{1}{Z(\widetilde{f}_k, x_k)} \dv{\mu_0(y)}{y} \exp\bigg(\frac{\widehat{f}_k(y) - c_k(x_k, y)}{\epsilon}\bigg),
\end{align*}
where $Z(\widetilde{f}_k, x_k) =  \int_{\cY} \exp\big(\frac{\widetilde{f}_k(y) - c_k(x_k, y)}{\epsilon}\big) d y  =  \int_{\cY} \exp\big(\frac{\widehat{f}_k(y) - c_k(x_k, y)}{\epsilon}\big) d \mu_0(y)$ is the normalizing constant (a.k.a.\ partition function). Define the plan:
\begin{align*}
    d \pi_k^f(x_k, y) \defeq d \bbP_k(x_k) d \mu_{x_k}^{\widetilde{f}_k}(y).
\end{align*}
Note that $\pi_k^f(\cdot \vert x_k) = \mu_{x_k}^{\widetilde{f}_k}$. For arbitrary $\pi_k \in \Pi(\bbP_k)$ we have:
\begin{align*}
    \cV_k(\widehat{f}_k, \pi_k) = \int_{\cX_k} \underbrace{\cG_{x_k, \widetilde{f}_k}(\pi_k(\cdot \vert x_k))}_{\geq \cG_{x_{\scaleto{k}{2.5pt}}, \widetilde{f}_{\scaleto{k}{2.5pt}}}(\mu_{x_{\scaleto{k}{2.5pt}}}^{\widetilde{f}_{\scaleto{k}{2.5pt}}})} d \bbP_k(x_k) \geq \int_{\cX_k} \cG_{x_k, \widetilde{f}_k} \big(\mu_{x_k}^{\widetilde{f}_k}\big) d \bbP_k(x_k) = \cV_k(\widehat{f}_k, \pi_k^f),
\end{align*}
i.e., distribution $\pi_k^f$ indeed minimizes $\cV_k$ for $\epsilon$-$\KL$ weak cost. Similar to \citep[Equation 14]{mokrov2024energyguided} we derive
\begin{align*}
    \cV_k(\widehat{f}_k, \pi_k^f) = \int_{\cX_k}\cG_{x_k, \widetilde{f}_k}\big(\pi_k^f(\cdot \vert x_k)\big) d \bbP_k(x_k) = - \epsilon \int_{\cX_k} \log Z(\widetilde{f}_k, x_k) \bbP_k(x_k).
\end{align*}
Now we are ready to analyze the gaps $\delta_{1, k}$ \eqref{eq:deltagap1_k} and $\delta_{2, k}$ \eqref{eq:deltagap2_k}. Note that our further derivations are similar to the proof of \citep[Theorem 2]{mokrov2024energyguided}.
\begin{align}
    \delta_{1, k} = \cV_k(\widehat{f}_k, \widehat{\pi}_k) - \cV_k(\widehat{f}_k, \pi_k^f) 
    = \nonumber \\
    \int\limits_{\cX_k} \!\Big\{\! \int\limits_{\cY} \!c_k(x_k, y) d \widehat{\pi}_k(y \vert x_k) \!-\! \epsilon H\big(\widehat{\pi}_k(\cdot \vert x_k)\big) \!- \! \int\limits_{\cY} \widetilde{f}_k(y) d \widehat{\pi}_k(y \vert x_k) \Big\} d \bbP_k(x_k) + \epsilon \int\limits_{\cX_k} \log Z(\widetilde{f}_k, x_k) d \bbP_k(x_k) 
    = \nonumber \\
    - \epsilon \int\limits_{\cX_k} \int\limits_{\cY} \!\!\!\!\!\!\underbrace{\frac{\widetilde{f}_k(y) - c_k(x_k, y)}{\epsilon}}_{= \log \dv{\pi_k^f(y \vert x_k)}{y} + \log Z(\widetilde{f}_k, x_k)} \!\!\!\!\!\! d \widehat{\pi}_k(y \vert x_k) d \bbP_k(x_k) + \epsilon \int\limits_{\cX_k} \log Z(\widetilde{f}_k, x_k) d \bbP_k(x_k) - \epsilon \int\limits_{\cX_k} H(\widehat{\pi}_k(\cdot \vert x_k)) d \bbP_k(x_k) 
    = \label{eq:proof:gaps:13} \\
    - \epsilon \int\limits_{\cX_k} \int\limits_{\cY} \log \dv{\pi_k^f(y \vert x_k)}{y} d \widehat{\pi}_k(y \vert x_k) d \bbP_k(x_k) + \epsilon \int\limits_{\cX_k}  \int\limits_{\cY} \log \dv{\widehat{\pi}_k(y \vert x_k)}{y} d \widehat{\pi}_k(y \vert x_k) d \bbP_k(x_k)  = \label{eq:proof:gaps:11} \\
    \epsilon \int\limits_{\cX_k} \int\limits_{\cY} \Big[ \log \dv{\widehat{\pi}_k(y \vert x_k)}{y} - \log \dv{\pi_k^f(y \vert x_k)}{y}\Big] d \widehat{\pi}_k(y \vert x_k) d \bbP_k(x_k) = \epsilon \int\limits_{\cX_k} \KL\big(\widehat{\pi}_k(\cdot \vert x_k)\Vert \pi_k^f(\cdot \vert x_k)\big) d \bbP_k(x_k) \label{eq:proof:gaps:12}.
\end{align}

In \eqref{eq:proof:gaps:11} and \eqref{eq:proof:gaps:12} we implicitly assume that $\widehat{\pi}_k(\cdot \vert x_k)$ is absolutely continuous. 
We remark that if this is not the case, then the derivations above still holds true, since in this case both negative entropy in \eqref{eq:proof:gaps:13} and $\KL$ divergence in \eqref{eq:proof:gaps:12} are equal to $+\infty$. Thanks to the fact that $\widehat{\pi}_k, \pi_k^f \in \Pi(\bbP_k)$, i.e., the first marginal of these distributions equals to $\bbP_k$, \eqref{eq:proof:gaps:12} can be further rewritten, see \citep[Appendix B.2]{mokrov2024energyguided}:
\begin{align}
    \delta_{1, k} = \epsilon \int\limits_{\cX_k} \KL\big(\widehat{\pi}_k(\cdot \vert x_k)\Vert \pi_k^f(\cdot \vert x_k)\big) d\bbP_k(x_k) = \epsilon \KL(\widehat{\pi}_k \Vert \pi_k^f). \label{eq:delta1_kl_ineq}
\end{align}

Regarding the gaps $\delta_{2, k}$ \eqref{eq:deltagap2_k}, a similar analysis as for $\delta_{1, k}$ yields:
\begin{align}
    \delta_{2, k} = \cV_k(\widehat{f}_k, \pi_k^*) - \cV_k(\widehat{f}_k, \pi_k^f) = \epsilon \KL(\pi_k^* \Vert \pi_k^f). \label{eq:delta2_kl_ineq}
\end{align}
To derive the final bounds, we recall \textit{Pinsker's} inequality \citep[Lemma 2.5]{tsybakov2009introduction}. Given distributions $\pi^a$ and $\pi^b$, we have:
\begin{align*}
    2 \rho_{\text{TV}}(\pi^a, \pi^b)^2 \leq \KL(\pi^a \Vert \pi^b),
\end{align*}
where $\rho_{\text{TV}}$ is the total variation distance, see \citep[Definition 2.4]{tsybakov2009introduction}. Using Pinsker's inequality, we find:
\begin{align*}
    \delta_1 + \delta_2 &= \sum_{k = 1}^{K} \lambda_k(\delta_{1, k} + \delta_{2, k}) \overset{\text{Eqs. }\eqref{eq:delta1_kl_ineq}, \eqref{eq:delta2_kl_ineq}}{=} \epsilon \sum_{k = 1}^{K} \lambda_k \big(\KL(\widehat{\pi}_k \Vert \pi_k^f) + \KL(\pi_k^* \Vert \pi_k^f)\big)
    \\
    &\hspace*{-6mm}\overset{\text{Pinsker's ineq.}}{\geq} \epsilon \sum_{k = 1}^{K} \lambda_k \big(2\rho_{\text{TV}}(\widehat{\pi}_k, \pi_k^f)^2 +2\rho_{\text{TV}}(\pi_k^*, \pi_k^f)^2\big) \geq \epsilon \sum_{k = 1}^{K} \lambda_k \big(\rho_{\text{TV}}(\widehat{\pi}_k, \pi_k^f) + \rho_{\text{TV}}(\pi_k^*, \pi_k^f)\big)^2 \\
    &\hspace*{-6mm}\overset{\text{Triangle ineq.}}{\geq}\epsilon \sum_{k = 1}^{K} \lambda_k \rho_{\text{TV}}(\widehat{\pi}_k, \pi_k^*)^2,
\end{align*}
which finishes the proof of statement 2.

\textit{Proof of statement 3 ($\gamma$-Energy weak cost function case).} Let $\pi_k \in \Pi(\bbP_k)$ be arbitrary and $\widehat{f}_k$ be a given potential. Below we take a closer look at functional $\cV_k$ \eqref{eq:V_func_k} for $\gamma$-Energy cost function:
\begin{align*}
\cV_k(\widehat{f}_k, \pi_k) &= \int\limits_{\cX_k} \Big\{ \int\limits_{\cY} c_k(x_k, y) d \pi_k(y \vert x_k)  + \gamma \cE_\ell^2\big(\pi_k(\cdot \vert x_k), \mu_0\big)\Big\} d \bbP_k(x_k) - \int\limits_{\cX_k} \int\limits_{\cY} \widehat{f}_k(y) 
d \pi_k(y \vert x_k) d \bbP_k(x_k)
\\
&= \int_{\cX_k} \hat C_k(x_k, \pi_k(y | x_k)) d \mathbb P_k(x_k),   
\end{align*}
where
\[
    \hat C_k(x_k,\mu) \defeq \int_{\cY} \big(c_k(x_k,y) - \hat f_k(y)\big) \, d\mu(y) + \gamma \cE_\ell^2\big( \mu, \mu_0 \big).
\]
To see the existence of a minimizer for $\cV_k$, observe that $\Pi(\bbP_k)$ is compact w.r.t.\ weak convergence of measures.
This follows easily from $\Pi(\bbP_k)$ being a closed subset of the compact $\mathcal P(\cX \times \cY)$ (where the latter is due to $\cX \times \cY$ being compact).
The function $\hat C_k$ is bounded from below (since $c_k, \hat f_k$ as well as $\cE_\ell^2$ are bounded from below), continuous w.r.t.\ weak convergence of measures (since $c_k, \hat f_k$ and $\ell$ are continuous), and convex in the second argument (as the integral is linear in $\mu$ and $\cE_\ell^2(\cdot,\mu_0)$ is convex).
Hence, there exists by \citep[Theorem 2.9]{backhoff2019existence} a minimizer $\pi^f_k \in \Pi(\bbP_k)$ for $\cV_k(\widehat{f}_k,\cdot)$.

Similarly, we have
\[
    \cV_k(\widehat{f}_k, \pi_k) =
    \int_{\cX_k \times \cY} \big(c_k(x_k,y) - \hat f_k(y)\big) \, d\pi_k(x_k,y) + \gamma \rho_\ell^2(\pi_k, \pi_0),
\]
where $\pi_0 \defeq \mathbb P_k \otimes \mu_0$,
from where it is evident that $\cV_k$ is $2 \gamma$-strongly convex on $\Pi(\bbP_k)$ w.r.t.\ the metric $\rho_\ell$ \eqref{eq:rho_ell_metric}. 
For the definition of strong convexity, see \citep[Appendix A, Definition 2]{asadulaev2024neural}.

Then the inequalities for the gaps $\delta_{1,k}$ \eqref{eq:deltagap1_k} and $\delta_{2, k}$ \eqref{eq:deltagap2_k} directly follow from strong convexity of $\cV_k$, see \citep[Appendix A, Lemma 1]{asadulaev2024neural}:
\begin{align*}
    \delta_{1, k} = \cV_k(\widehat{f}_k, \widehat{\pi}_k) - \cV_k(\widehat{f}_k, \pi_k^f) \overset{\text{{\scriptsize \makecell{Lemma 1,\\ Asadulaev et. al.}}}}{\geq} \gamma \rho_\ell(\pi_k^f, \widehat{\pi}_k)^2, \\
    \delta_{2, k} = \cV_k(\widehat{f}_k, \pi_k^*) - \cV_k(\widehat{f}_k, \pi_k^f) \hspace*{6.7mm}\geq\hspace*{6.7mm} \gamma\rho_\ell(\pi_k^f, \pi_k^*)^2.
\end{align*}

To finish the proof, we summarize the inequalities above over $k \in \overline{K}$ with weights $\lambda_k$:
\begin{align*}
    \delta_1 + \delta_2 &= \sum_{k = 1}^{K} \lambda_k (\delta_{1, k} + \delta_{2, k}) \geq \sum_{k = 1}^{K} \lambda_k \big(\gamma\rho_\ell(\pi_k^f, \widehat{\pi}_k)^2 + \gamma\rho_\ell(\pi_k^f, \pi_k^*)^2\big) \\ 
    &\geq \gamma\sum_{k = 1}^{K} \lambda_k\frac{\big(\rho_\ell(\pi_k^f, \widehat{\pi}_k) + \rho_\ell(\pi_k^f, \pi_k^*)\big)^2}{2} \overset{\text{Triang. ineq.}}{\geq} \frac{\gamma}{2}\sum_{k = 1}^{K} \lambda_k \rho_\ell(\widehat{\pi}_k, \pi_k^*)^2.\qedhere
\end{align*}

\end{proof}

\subsection{Intuitive derivation of max-min OT barycenter objective \eqref{eq:weakbary_maxmin}}\label{app:intuitive-derivation}

In this subsection, we give an intuition behind our proposed bi-level objective \eqref{eq:weakbary_maxmin}. In particular, we explain how we managed to avoid $\min$-$\max$-$\min$ optimization and whence the congruence condition arises. 

To derive our proposed formulation \eqref{eq:weakbary_maxmin} we borrow some ideas from the existing literature:

1. The first idea stems from the NOT paper \citep{korotin2023neural}. The authors take advantage of the dual formulation of Weak OT problem \eqref{eq:weakot_dual} and come up with the following $\max$-$\min$ objective:
\begin{align}
     \text{OT}_C(\mathbb{P}, \mathbb{Q}) = \sup_{f \in \mathcal{C}(\mathcal{Y})} \inf_{\pi \in \Pi(\mathbb{P})}  \bigg\{ \mathbb{E}_{x \sim \mathbb{P}} C(x, \pi(\cdot \vert x)) - \mathbb{E}_{y \sim \pi^{\mathcal{Y}}} f(y) + \mathbb{E}_{y \sim \mathbb{Q}} f(y)\bigg\}.  \label{eq-exp-not}\tag{NOT}
\end{align}

Recall that in the formula above $\pi^{\mathcal{Y}}$ is the second marginal of the plan $\pi$ (projection to $\mathcal{Y}$). Note that the \textbf{direct extension} of the \eqref{eq-exp-not} objective to the barycenter problem \textbf{will result in} a $\min$-$\max$-$\min$ \textbf{problem}, because we will have to optimize not only w.r.t.\ to $f_k$ and $\pi_k$, but also w.r.t. the barycenter distribution $\mathbb{Q}$. This is approximately how the method from WIN paper \citep{korotin2022wasserstein} is constructed (it is one of the baselines we consider in our paper). In contrast, we proceed in a smarter way and combine the \eqref{eq-exp-not} objective with another brilliant idea which ultimately allows us to avoid $\min$-$\max$-$\min$: 

2. \textbf{Congruent} potentials. Let $\mathbb{P}_{1:K}$ be the reference distributions for which we seek the barycenter. Let $\mathcal{H}_k(f, \pi, \mathbb{Q})$ be the functional under the $\sup_{f}$ $\inf_{\pi}$ in \eqref{eq-exp-not}, i.e.: 
\begin{align*}
\text{OT}_{C_k}(\mathbb{P}_k, \mathbb{Q}) = \sup_{f \in \mathcal{C}(\mathcal{Y})} \inf_{\pi \in \Pi(\mathbb{P}_k)} \mathcal{H}_k(f, \pi, \mathbb{Q}).
\end{align*}
The barycenter problem \eqref{eq:weakbary_primal} with weights $\lambda_{1:K}$ then can be formulated as follows:
\begin{gather}
\mathcal{L}^* = \inf_{\mathbb{Q} \in \mathcal{P}(\mathcal{Y})} \sum_{k = 1}^{K} \sup_{f_k \in \mathcal{C}(\mathcal{Y})} \inf_{\pi_k \in \Pi(\mathbb{P})} \lambda_k \mathcal{H}_k(f_k, \pi_k, \mathbb{Q}). \label{eq-exp-notb-naive}\tag{NOTB NAIVE}
\end{gather}
This is exactly what we called the \textit{direct $\min$-$\max$-$\min$ extension of the \eqref{eq-exp-not} objective to the barycenter problem} in the paragraph above. So, the question is, how to avoid tri-level adversariality? We propose to exploit the \textbf{optimality condition} of the \eqref{eq-exp-notb-naive} objective w.r.t.\ the learned barycenter distribution $\mathbb{Q}$. In simpler words, we equate \textbf{the (Frechet) derivative} of the \eqref{eq-exp-notb-naive} objective w.r.t.\ $\mathbb{Q}$ to zero at the optimum, i.e., where $\mathbb{Q} = \mathbb{Q}^*$. In the mathematical literature such ``derivative'' is called first variation, see \citep[\S 7.2]{santambrogio2015optimal}. Omitting some details, throwing away $\sup_{f}$, $\inf_{\pi}$ and being mathematically non-rigorous, the optimality condition reads as follows:
\begin{align}
\nabla_{\mathbb{Q}} \bigg\{\sum_{k = 1}^{K} \lambda_k \mathbb{E}_{y \sim \mathbb{Q}} f_k^*(y)\bigg\}\bigg|_{\mathbb{Q} = \mathbb{Q}^*} = 0, \label{eq-exp-optimality}\tag{OPTIMALITY}
\end{align}
where $f_k^*$ are the optimal potentials of the \eqref{eq-exp-not} objectives between $\mathbb{P}_k$ and $\mathbb{Q}^*$.
Indeed, each expression of the \eqref{eq-exp-not} objective consists of three terms, and only the last term depends on $\mathbb{Q}$. When we take the derivative $\nabla_{\mathbb{Q}}$, the first two terms vanish. The analysis of the \eqref{eq-exp-optimality} problem is already an easy task, since it is known that $\nabla_{\mathbb{Q}} \big[\mathbb{E}_{y \sim \mathbb{Q}} f(y) \big]= f$. Thanks to this observation, the \eqref{eq-exp-optimality} reads as $\sum_{k = 1}^{K} \lambda_k f_k = 0$, which is exactly the \textbf{congruence} condition that we utilize in our paper. 

In fact, eliminating the outer $\inf_{\mathbb{Q}}$ optimization and incorporating the \textbf{congruence} condition into the \eqref{eq-exp-notb-naive} objective is exactly the method we propose in our paper. Of course, our explanation above is far from being mathematically rigorous. But we hope that our ideological derivation in this subsection sheds some light on the magic behind our approach and makes it more intuitive.

\section{Extended experiments}\label{app:experimens_extended}

\subsection{Barycenters for Gaussians}
In this experiment, we consider the OT barycenter problem with \textit{Gaussian} input distributions $\bbP_k$. In the Gaussian case, it is known that the ground-truth OT barycenter for the classical OT cost functions $c_k(x_k, y)=\sfrac{1}{2}\|x_k-y\|^2$ is also Gaussian and can be computed using the fixed point iteration procedure \citep{alvarez2016fixed}. As the baselines, we take two recent solvers: EgBary \citep{kolesov2023energy} and WIN \citep{korotin2022wasserstein}. The authors of the first paper solve the Entropic OT (EOT) barycenter problem, see Table \ref{table-weakcost-examples}, and we consider their approach for small $\epsilon=0.01$ to reduce the bias. The second paper learns unregularized OT barycenters with quadratic cost functions through an iterative procedure.

For assessment of our solver and WIN, we use the unexplained variance percentage metrics defined by $\mathcal{L}_{2}\text{-UVP}(\hat{T})=100\cdot[\|\hat{T}-T^*\|_{\mathbb{P}}^2/\text{var}(\bbQ^*)] \%$, where $\bbQ^*$ is the ground truth OT barycenter, see \citep[\S 5.1]{korotin2021wasserstein}. For evaluation of EgBary which learns the EOT plans, we consider their barycentric projections, see \citep[Appendix C4]{kolesov2023energy}. Our results are presented in Table \ref{table-gaussians}, the reported values of $\mathcal{L}_{2}\text{-UVP}$ are averaged over input distributions $\bbP_k$ (w.r.t. barycenter weights $\lambda_k$).

\vspace{-4mm}
\begin{table}[!h]
\centering
\footnotesize
\begin{tabular}{c|c|c|c|c|c}\hline
\textbf{Method/Dim}    & 2 & 4 & 8 & 16 & 64 \\ \hline
\textbf{Ours} & \textbf{0.01} & \textbf{0.02} & \textbf{0.04} & \textbf{0.04} & \textbf{0.08} \\ \hline
EgBary &  0.02 & 0.05 & 0.06 & 0.09 & 0.84\\ \hline
WIN &  0.03 & 0.08 & 0.13 & 0.25 & 0.75 \\ \hline
\end{tabular}
\caption{$\mathcal{L}_{2}\text{-UVP}$ for our method, EgBary  ($\epsilon=0.01$) and WIN \citep{korotin2022wasserstein}, $D=2,4,8,16,64$.}
\label{table-gaussians}
\vspace{-2mm}
\end{table}
We see that our approach gives better results than its competitors for all the dimensions. The difference is especially visible for the biggest dimension $D=64$. These results are expected, since EgBary approach is designed to learn the regularized barycenter, while WIN algorithm utilizes the iterative procedure leading to accumulation of the error with the increase of dimension.

\subsection{Barycenters for MNIST digits (``0'' and ``1'')}\label{app:mnist_01}

In this experiment, we seek for the barycenter of two image distributions: grayscaled handwritten digits ``0'' ($\bbP_1$) and grayscaled handwritten digits ``1'' ($\bbP_2$). We use images from the MNIST dataset.  The weights for the barycenter problem are $\lambda_k = \sfrac{1}{2}$, the cost functions are classical quadratic Euclidean. The described task is popular in previous continuous OT barycenter research \citep{pmlr-v139-fan21d,korotin2022wasserstein, noble2023treebased, kolesov2023energy}. 

\textbf{Considered data setups.} It is known that the support of the recovered barycenter is contained in the manifold $\cM$ of weighted pixel-wise blends $\sfrac{1}{2}\cdot x_1 + \sfrac{1}{2}\cdot x_2$ of input images $x_1\sim \bbP_1$, $x_2\sim\bbP_2$. 
This allows us to use our idea with an auxiliary StyleGAN model trained on the pixel-wise combinations of digits. In this case, we learn OT barycenters in the latent space of this StyleGAN with \textit{non-quadratic} cost functions. At the same time, we do not limit ourselves to the latent space and also consider the conventional (ambient) data-space setup without auxiliary generative models.

\begin{figure*}[!t]
 \centering
\begin{subfigure}[b]{0.48\linewidth} 
      \centering
      \includegraphics[width=0.995\linewidth]{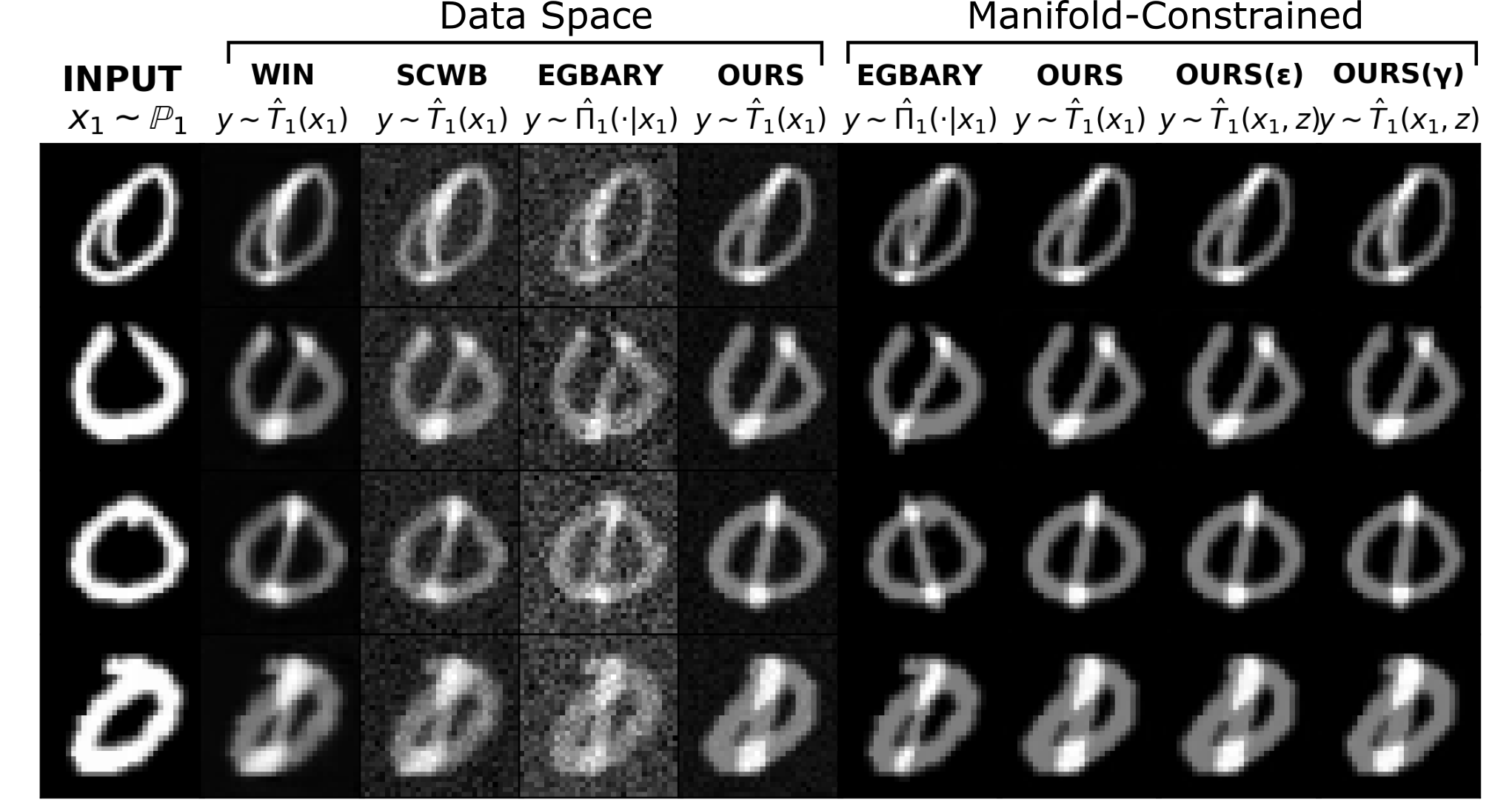}
      \caption{Source and transformed samples for input $\bbP_1$.}
     \label{fig:mnist-0-1-map_1}
 \end{subfigure}\hfill
  \begin{subfigure}[b]{0.48\linewidth}  
     \centering
     \includegraphics[width=0.995\linewidth]{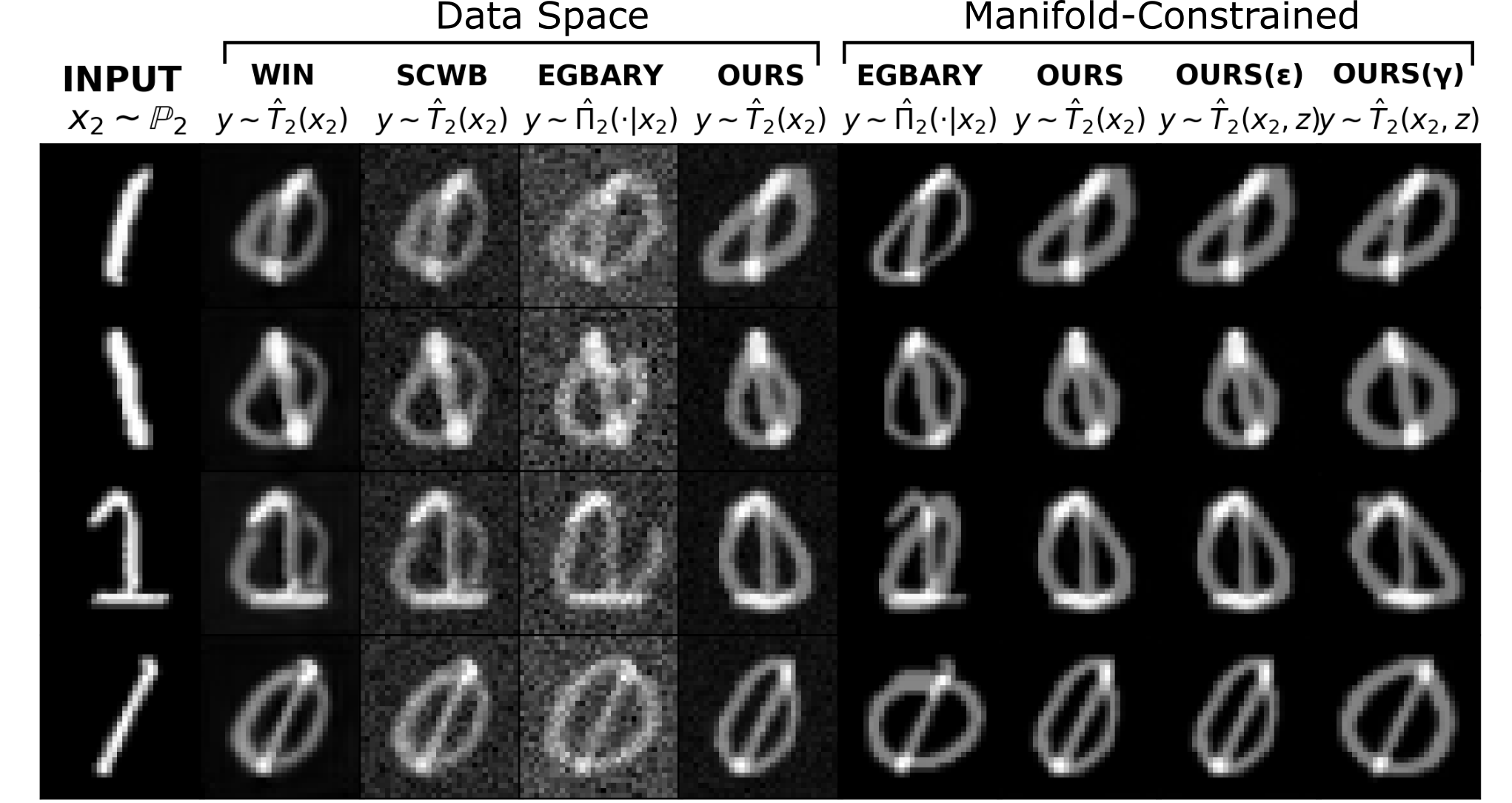}
     \caption{Source and transformed samples for input $\bbP_2$.}
      \label{fig:mnist-0-1-map_2}
  \end{subfigure}
\vspace{-1.6mm}
 \caption{Learned (stochastic) maps to the OT barycenter by different solvers; MNIST 0/1 experiment (\S\ref{app:mnist_01}).}
 \label{fig:MNIST-0-1}
 \vspace{-1mm}
\end{figure*}

\begin{figure*}[!t]
     \centering
    \begin{subfigure}[b]{0.48\linewidth}  
          \centering
          \includegraphics[width=0.995\linewidth]{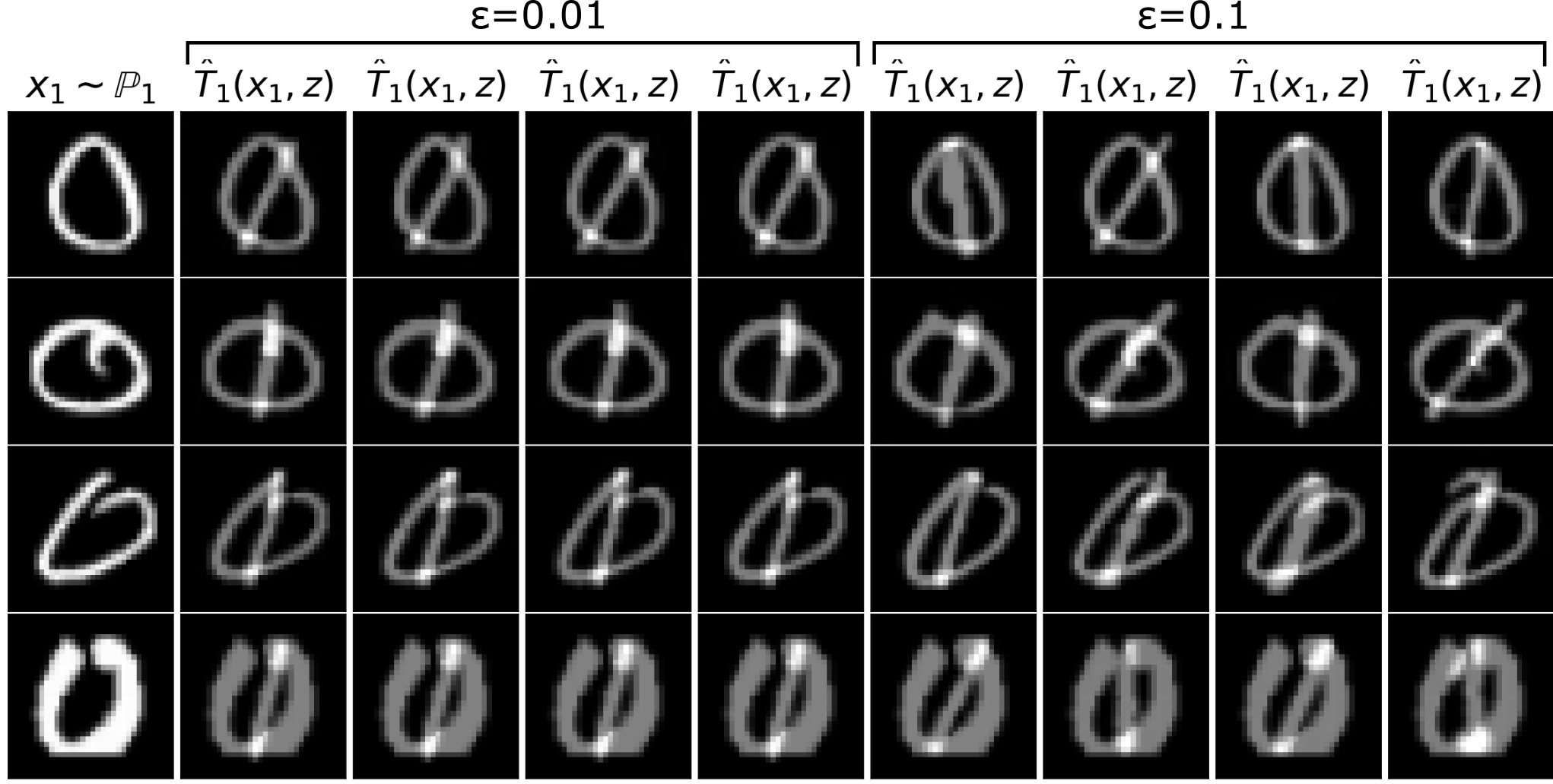}
          \caption{Source and transformed samples for input $\bbP_1$.}
         \label{fig:mnist-0-1-map_1_canvas_ent}
     \end{subfigure}\hfill
      \begin{subfigure}[b]{0.48\linewidth}  
         \centering
         \includegraphics[width=0.995\linewidth]{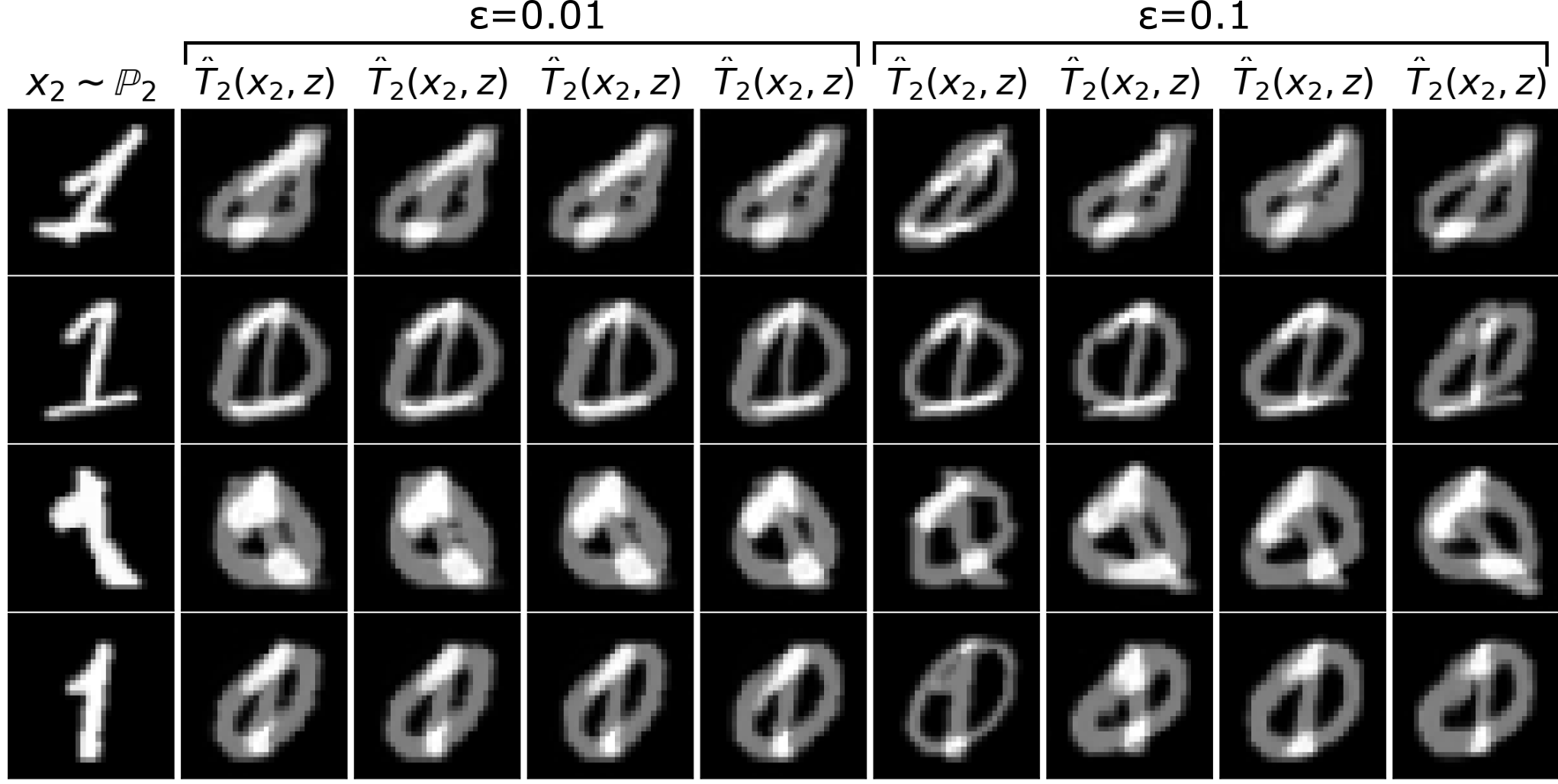}
         \caption{Source and transformed samples for input $\bbP_2$.}
          \label{fig:mnist-0-1-map_2_canvas}
      \end{subfigure}
 \vspace{-1.6mm}
     \caption{Additional samples from \textbf{our} learned stochastic OT barycenter maps. Manifold-constrained data setup; $\epsilon$-$\KL$ weak cost functions with different regularization strengths $\epsilon$; MNIST 0/1 experiment (\S\ref{app:mnist_01}).}
     \label{fig:MNIST-0-1_canvas_ent}
     \vspace{-1mm}
 \end{figure*}

\begin{figure*}[!t]
     \centering
    \begin{subfigure}[b]{0.48\linewidth}  
          \centering
          \includegraphics[width=0.995\linewidth]{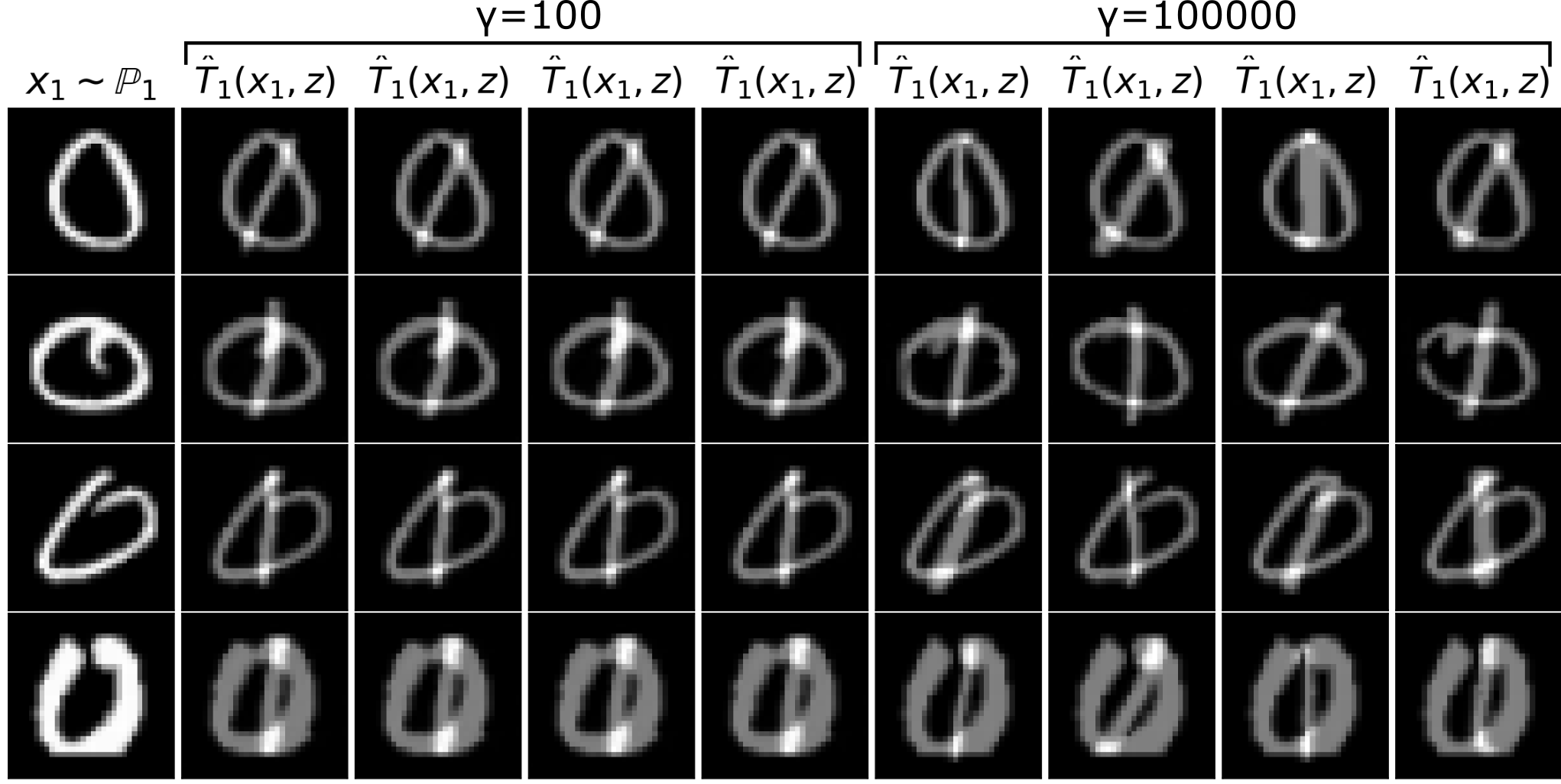}
          \caption{Source and transformed samples for input $\bbP_1$.}
         \label{fig:mnist-0-1-map_1_canvas_ker}
     \end{subfigure}\hfill
      \begin{subfigure}[b]{0.48\linewidth}  
         \centering
         \includegraphics[width=0.995\linewidth]{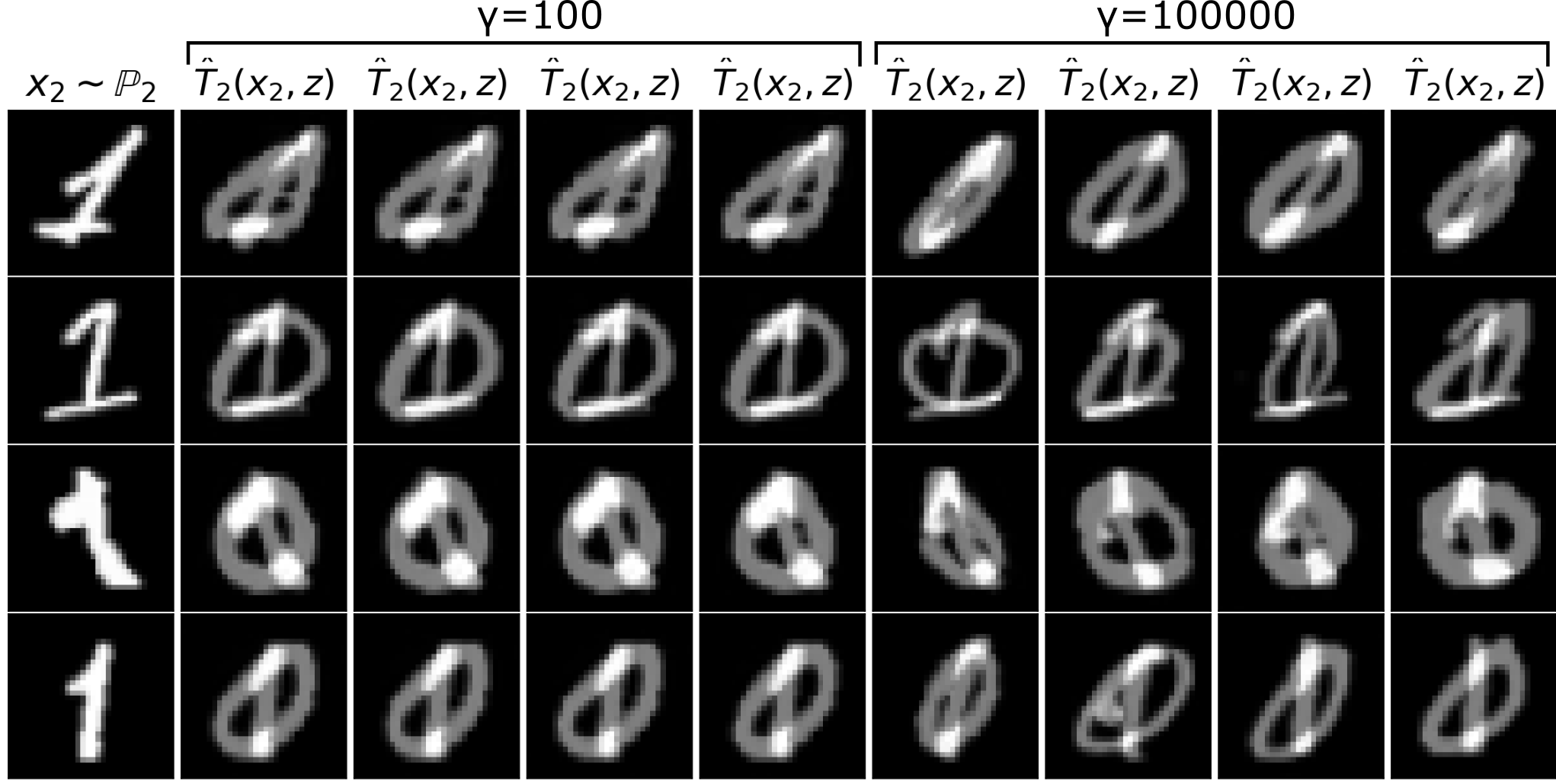}
         \caption{Source and transformed samples for input $\bbP_2$.}
          \label{fig:mnist-0-1-map_2_canvas_ker}
      \end{subfigure}
 \vspace{-1.6mm}
     \caption{Additional samples from \textbf{our} learned stochastic OT barycenter maps. Manifold-constrained data setup; $\gamma$-Energy weak cost functions with different regularization strengths $\gamma$; MNIST 0/1 experiment (\S\ref{app:mnist_01}).}
     \label{fig:MNIST-0-1_canvas_ker}
     \vspace{-1mm}
 \end{figure*}

\textbf{Evaluation.} Our results are presented in Fig. \ref{fig:MNIST-0-1}. In the \underline{data-space} setup, we launch our Algorithm \ref{algorithm:otbary} with classical quadratic cost functions and utilize deterministic maps parameterization, see \S\ref{subsec:stoch_maps}, i.e., we recover the \textit{unregularized} OT barycenter. As the baselines for visual comparison, we take WIN \citep{korotin2022wasserstein} and SCWB \citep{pmlr-v139-fan21d} which are specifically designed for solving the unregularized barycenter problem with quadratic cost functions. Qualitatively, our obtained samples from the barycenter are in concordance with these competitors. We also include EgBary \citep{kolesov2023energy} method as the baseline in the data-space setup. Its samples are noised due to Entropic regularization. This shows that the usage of regularized cost functions, e.g., \eqref{eq:klcost} and \eqref{eq:energycost}, is more reasonable in the \underline{manifold-constrained} setup. The corresponding samples can be found in Fig.\ \eqref{fig:MNIST-0-1}. For $\epsilon$-$\KL$ and $\gamma$-Energy cost function cases, we use $\epsilon=0.01$ and $\gamma = 100$. More samples from our recovered barycenters with different regularizations can be found in Fig. \ref{fig:MNIST-0-1_canvas_ent} and Fig. \ref{fig:MNIST-0-1_canvas_ker}. 
\vspace{-2mm}
\section{Experimental details}\label{app:experiments_details}

We aggregate the hyper-parameters of our  Algorithm~\ref{algorithm:otbary} for different experiments in Table~\ref{table:hyperparams}. For our experiments with manifold-constrained barycenter learning (Shape-Color \S\ref{subsec:colorshape}; Ave, celeba! \S\ref{subsec:ave_celeba}; MNIST 0/1 \S\ref{app:mnist_01}) we use StyleGAN2-ada model from the official repository:
\begin{center}
\url{https://github.com/NVlabs/stylegan2-ada-pytorch}.
\end{center}
For the details of the baseline solvers, see \citep[Appendix C]{kolesov2023energy}. We utilize exactly the same hyper-parameters. To implement EgBary solver \citep{kolesov2023energy}, we use their publicly available source code.

\textbf{Color preserving cost.} The operation $H_{c}$ which extracts the color from the image is designed as follows. We transform all the pixels in the given image to HSV scale. Then we pick only those whose value (V) is greater than $0.8$. We take the mean values of these pixels and obtain a vector in $[0,1]^{3}$.

\begin{table}[]
\centering
\small
\hspace*{-0mm}\begin{tabular}{|l|l|l|l|l|l|l|l|l|l|l|l|l|l|l|}
\hline
Experiment  & D     & K & $\epsilon$ & $\gamma$ & {\scriptsize \makecell{\textit{batch}\\ \textit{size}}} & $M_T$ & $\lambda_{1}$   & $\lambda_{2}$   & $\lambda_{3}$  & $f_{k,\theta}$   &  $T_{k,\phi}$   & $lr_{f_{k,\theta}}$  & $lr_{T_{k,\phi}}$  & {\scriptsize \makecell{\textit{\# of}\\ \textit{epochs}}}    \\ \hline
Toy 2D      & 2     & 3 & 1   & -      & $2^{10}$ & 3    & 1/3  & 1/3  & 1/3  & MLP    & MLP    & 1e-3 & 1e-3 & 1200  \\ \hline
Toy 2D      & 2     & 3 & -   & 1      & $2^{10}$ & 3    & 1/3  & 1/3  & 1/3  & MLP    & MLP    & 1e-3 & 1e-3 & 1200  \\ \hline
Gaussians   & 2$-$64  & 2 & -   & -      & $2^{10}$ & 3    & 0.25 & 0.25 & 0.5  & MLP    & MLP    & 1e-3 & 1e-3 & 1200  \\ \hline
MNIST 0/1   & 1x32x32  & 2 & -   & -      & 64   & 10   & 0.5  & 0.5  & -    & ResNet & UNET   & 2e-4 & 2e-4 & 20K \\ \hline
MNIST 0/1   & 512   & 2 & -   & -      & 64   & 10   & 0.5  & 0.5  & -    & ResNet & ResNet & 2e-4 & 2e-4 & 20K \\ \hline
MNIST 0/1   & 512   & 2 & 0.1 & -      & 64   & 10   & 0.5  & 0.5  & -    & ResNet & ResNet & 2e-4 & 2e-4 & 20K \\ \hline
MNIST 0/1   & 512   & 2 & -   & 10K & 64   & 10   & 0.5  & 0.5  & -    & ResNet & ResNet & 2e-4 & 2e-4 & 20K \\ \hline
Ave Celeba  & 3x64x64 & 3 & -   & -      & 64   & 10   & 0.25 & 0.5  & 0.25 & ResNet & UNET   & 2e-4 & 2e-4 & 40K \\ \hline
Ave Celeba  & 512   & 3 & -   & -      & 64   & 10   & 0.25 & 0.5  & 0.25 & ResNet & ResNet & 2e-4 & 2e-4 & 40K \\ \hline
Ave Celeba  & 512   & 3 & 0.1 & -      & 64   & 10   & 0.25 & 0.5  & 0.25 & ResNet & ResNet & 2e-4 & 2e-4 & 40K \\ \hline
Ave Celeba  & 512   & 3 & -   & 10K & 64   & 10   & 0.25 & 0.5  & 0.25 & ResNet & ResNet & 2e-4 & 2e-4 & 40K \\ \hline
Shape Color & 512   & 2 & 0.1 & -      & 64   & 10   & 0.5  & 0.5  & -    & ResNet & ResNet & 2e-4 & 2e-4 & 20K \\ \hline
\end{tabular}
\caption{Hyper-parameters of Algorithm~\ref{algorithm:otbary} for different experiments.}
\label{table:hyperparams}
\end{table}

\begin{figure*}[!t]
     \centering
    \begin{subfigure}[b]{0.3205\linewidth}  
          \centering
          \includegraphics[width=1.07\textwidth, height = 0.502\textheight]{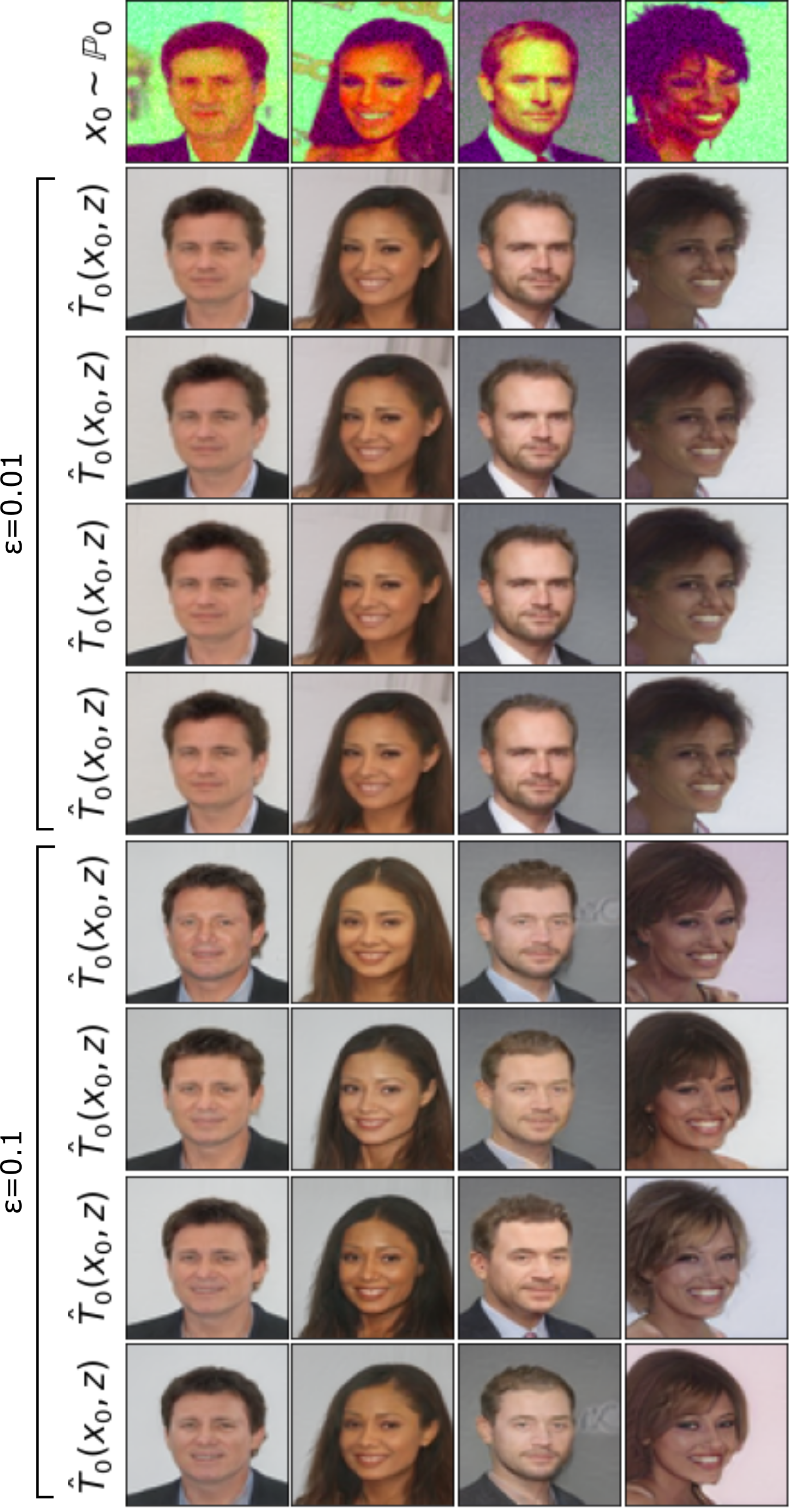}
          \caption{input, transformed samples ($\bbP_0$).}
          \label{fig:celeba-entropic-p0}
     \end{subfigure}\hfill
      \begin{subfigure}[b]{0.3205\linewidth}   
         \centering
         \includegraphics[width=0.995\textwidth]{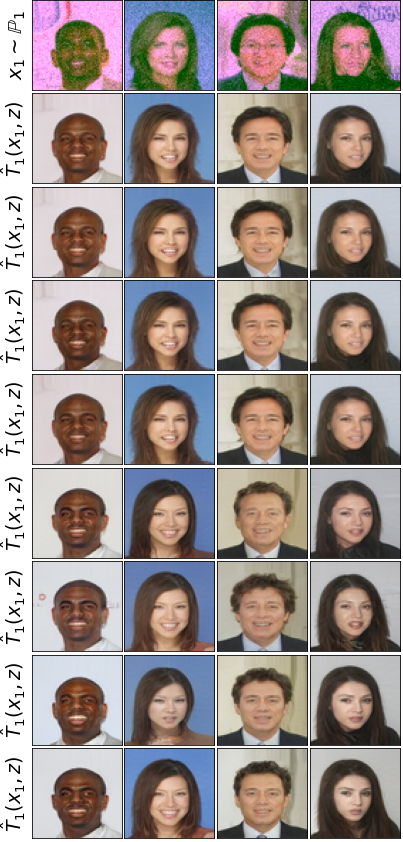}
         \caption{input, transformed samples ($\bbP_1$).}
         \label{fig:celeba-entropic-p1}
      \end{subfigure}
      \begin{subfigure}[b]{0.3205\linewidth} 
         \centering
         \includegraphics[width=0.995\textwidth]{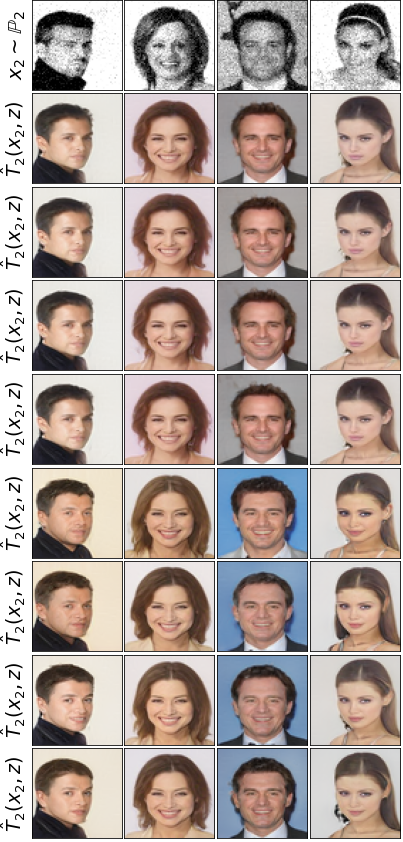}
         \caption{input, transformed samples ($\bbP_2$).}
         \label{fig:celeba-entropic-p3}
      \end{subfigure}
 \vspace{-2mm}
     \caption{Additional samples from \textbf{our} learned stochastic OT barycenter maps. Manifold-constrained data setup; $\epsilon$-$\KL$ weak cost functions with different regularization strengths $\epsilon$; Ave, Celeba! experiment (\S\ref{subsec:ave_celeba}).}
     \label{fig:ave-celeba-canvas-ent}
 \vspace{-1mm}
 \end{figure*}

\begin{figure*}[!t]
     \centering
    \begin{subfigure}[b]{0.3205\linewidth}
          \centering
          \includegraphics[width=1.07\textwidth,height=0.50215\textheight]{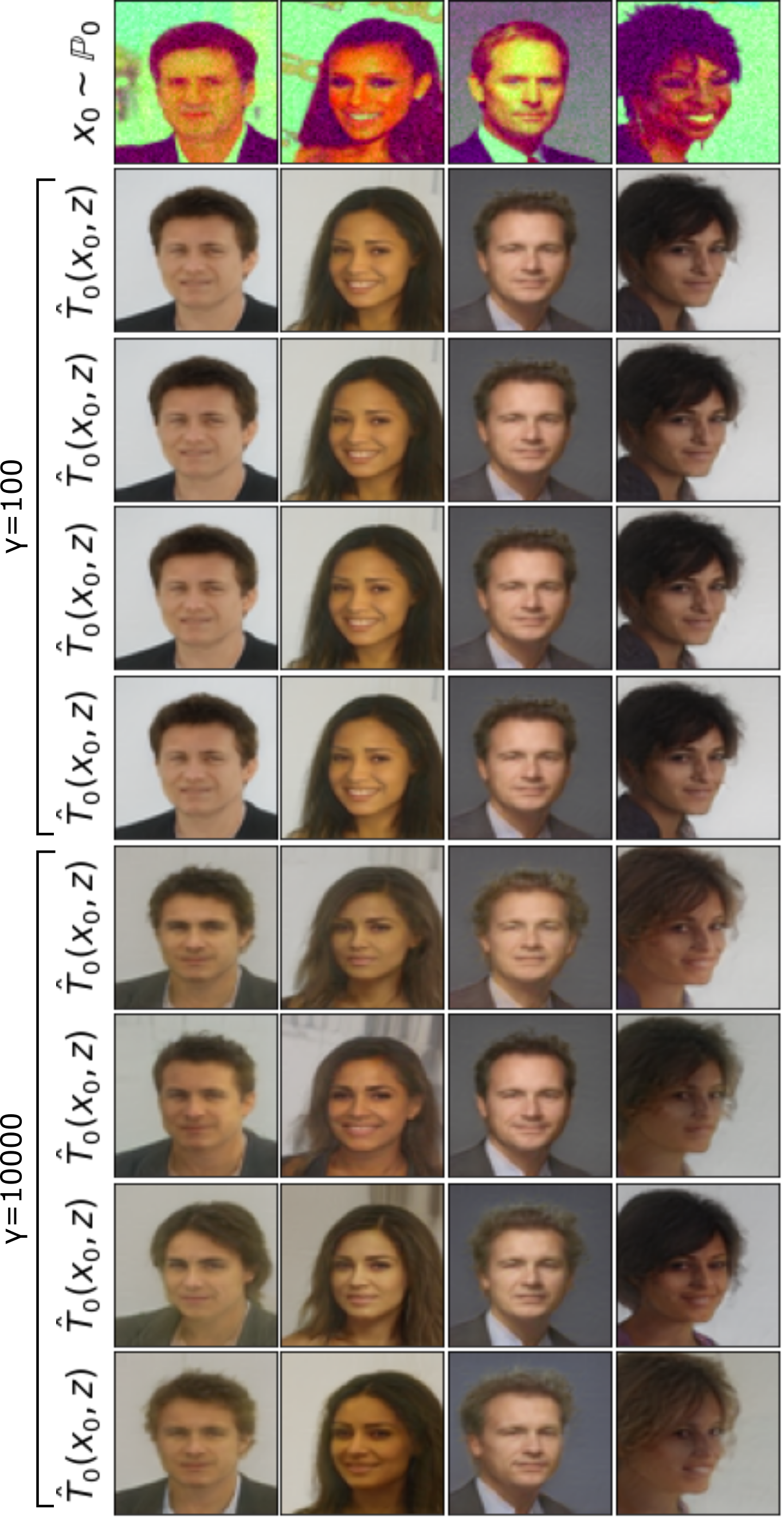}
          \caption{input, transformed samples ($\bbP_0$).}
          \label{fig:celeba-ker-p0}
     \end{subfigure}\hfill
      \begin{subfigure}[b]{0.3205\linewidth}  
         \centering
         \includegraphics[width=0.995\textwidth]{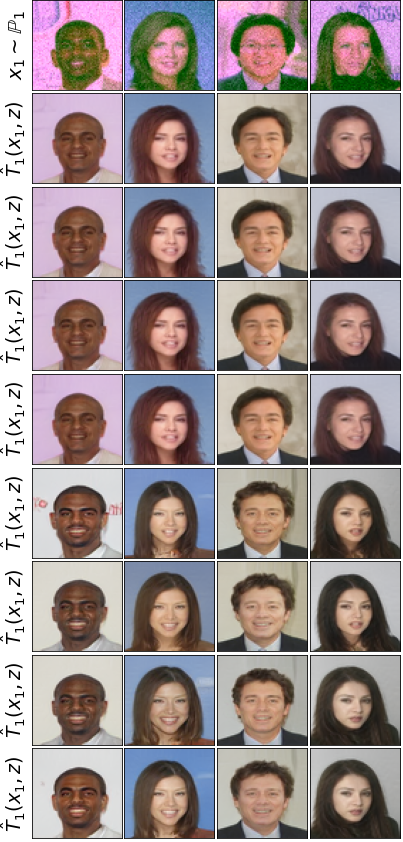}
         \caption{input, transformed samples ($\bbP_1$).}
         \label{fig:celeba-ker-p1}
      \end{subfigure}
      \begin{subfigure}[b]{0.3205\linewidth}
         \centering
         \includegraphics[width=0.995\textwidth]{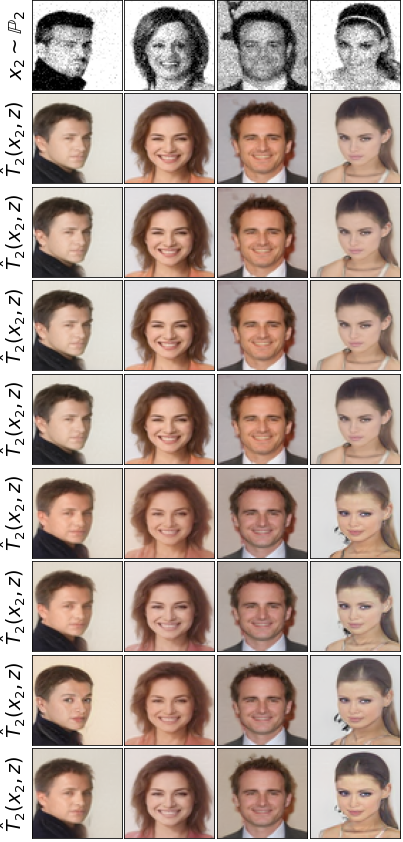}
         \caption{input, transformed samples ($\bbP_2$).}
         \label{fig:celeba-ker-p2}
      \end{subfigure}
 \vspace{-2mm}
     \caption{Additional samples from \textbf{our} learned stochastic OT barycenter maps. Manifold-constrained data setup; $\gamma$-Energy weak cost functions with different regularization strengths $\gamma$; Ave, Celeba! experiment (\S\ref{subsec:ave_celeba}).}
     \label{fig:ave-celeba-canvas-ker}
 \vspace{-3mm}
 \end{figure*}

\vspace{-2mm}
\section{Ave Celeba! experimental insights}\label{app:ave-celeba-ext}

In this section, we provide further details on Ave Celeba! experiment (\S \ref{subsec:ave_celeba}). In particular, we demonstrate training curves for our method, reveal time consumptions and carry out extended comparisons with the competing approaches.

\textbf{Comparison metrics details.} When comparing the methods, we utilize the following metrics: FID, $\mathcal{L}_2$-$\text{UVP}$ and transport cost ($W_2^2$).
The definitions of these metrics are as follows:
\begin{gather*}
    \mathcal{L}_2\text{-UVP}(\widehat{T}) = \frac{100 \%}{\text{Var}(\mathbb{Q}^*)} \cdot \mathbb{E}_{x \sim \mathbb{P}} \big[\Vert \widehat{T}(x) - T^*(x) \Vert^2\big] \, ; \quad \quad W_2^2(\widehat{T}) = \mathbb{E}_{x \sim \mathbb{P}} \big[\Vert \widehat{T}(x) - x \Vert^2\big].
\end{gather*}
In these formulas, $\mathbb{P}$ is a reference distribution ($\mathbb{P} \in \mathbb{P}_{1:3}$); $T^*$ is the ground truth squared Euclidean OT mapping between $\mathbb{P}$ and the GT barycenter $\mathbb{Q}^*$ (which are known by construction);  $\widehat{T}$ is a learned mapping. If the learned mapping $\widehat{T}$ is stochastic, i.e., it permits additional noise or it is represented by energy potential, $\mathcal{L}_2$-$\text{UVP}$ ($W_2^2$) are additionally averaged over this stochasticity. If $\widehat{T}$ maps to latent space, its output is additionally fed to StyleGAN encoder before computing the metrics.  Note that $\mathcal{L}_2$-$\text{UVP}$ directly compares the learned transport map with the true OT map to the barycenter.

\begin{figure*}[!t]
     \centering
    \begin{subfigure}[b]{1.0\linewidth} 
          \centering
          \hspace*{-18mm}\includegraphics[width=1.2\linewidth]{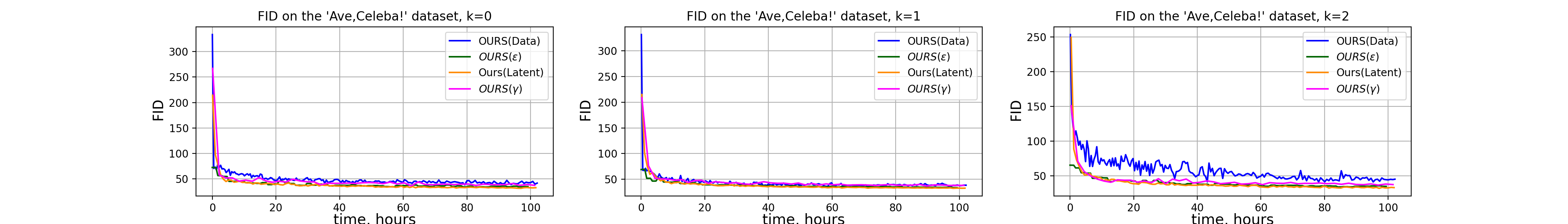}
          \caption{FID vs. time.}
         \label{fig:celeba-fid-vs-time}
     \end{subfigure}
\vspace*{-3mm}\newline
      \begin{subfigure}[b]{1.0\linewidth}  
         \centering
         \hspace*{-18mm}\includegraphics[width=1.2\linewidth]{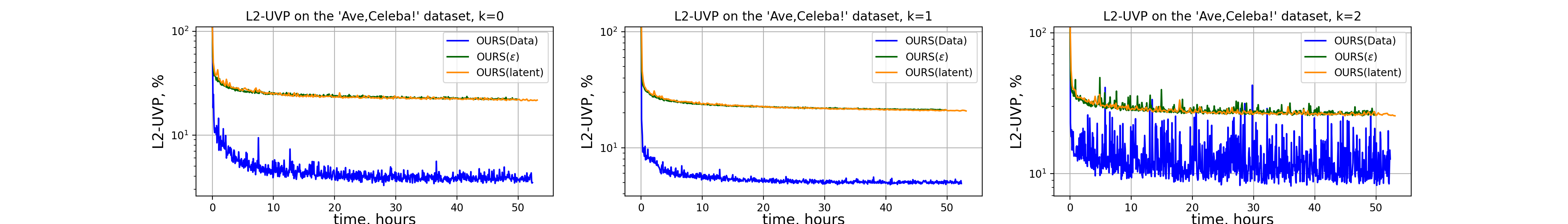}
         \caption{$\mathcal{L}_2$-$\text{UVP}$ vs. time.}
          \label{fig:celeba-l2uvp-vs-time}
      \end{subfigure}
\vspace*{-3mm}\newline
      \begin{subfigure}[b]{1.0\linewidth}  
         \centering
         \hspace*{-18mm}\includegraphics[width=1.2\linewidth]{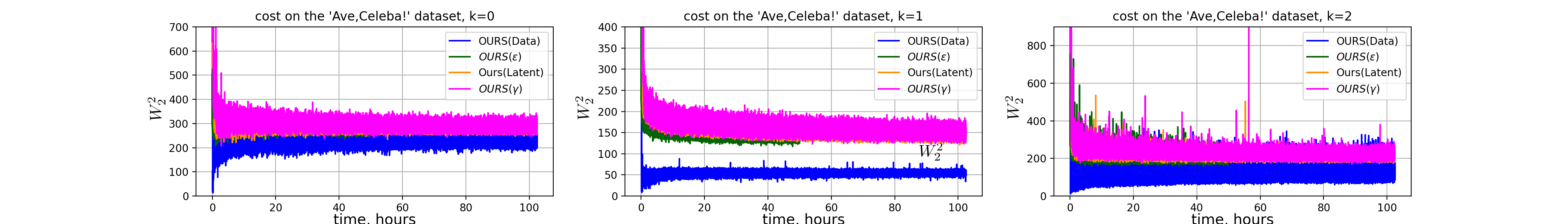}
         \caption{Transport cost vs. time.}
          \label{fig:celeba-cost-vs-time}
      \end{subfigure}
 \vspace*{-9.6mm}\newline
     \caption{Training curves for OUR proposed method (different costs, data/latent setups); Ave Celeba!}
     \label{fig:celeba-time}
     \vspace{-1mm}
 \end{figure*}

\textbf{Convergence curves for our method.}  We provide the detailed learning curves of our proposed method (in data/latent space, with different costs), to demonstrate its convergence capabilities, see Figure \ref{fig:celeba-time}. For the reasonable metrics which we track during the training, we choose: (a) FID \ref{fig:celeba-fid-vs-time}; (b) $\mathcal{L}_2$-$\text{UVP}$ \ref{fig:celeba-l2uvp-vs-time}; (c) Transport cost ($W_2^2$) \ref{fig:celeba-cost-vs-time}.

Interestingly, our approach experiences some local instabilities when learning the third barycenter mapping $\mathbb{P}_3 \rightarrow \mathbb{Q}^*$. This is because the third distribution represents \textit{grayscaled} images (see, e.g., Figure 4(c) from our paper), which makes the corresponding mapping much more difficult compared to the others.

\textbf{Training/inference times.} In Table \ref{tbl-ave-celeba-training-times}, we provide the \textbf{training} and \textbf{inference} times on Ave Celeba! setup ($k = 3$) for the considered competing solvers. 
\begin{table}[]
\centering
\begin{tabular}{|l|l|l|l|l|l|l|l|}
\hline
Method  & OURS(Data) & OURS(latent) & OURS($\epsilon$) & OURS($\gamma$) & WIN  & EgBary(Data) & EgBary(latent) \\ \hline
Training & 30 h        & 40 h         & 40 h      & 40 h         &$\sim$ 100 h  &   100 h          &   30 h             \\ \hline
Inference  & 0.01 s       & 0.05 s       & 0.05 s     & 0.08 s       & 0.02 s & 175 s          & 150 s            \\ \hline
\end{tabular}
\caption{Training/inference times of different barycenter solvers; Ave Celeba! experiment.}
\label{tbl-ave-celeba-training-times}
\end{table}
When training the competing solvers, we choose their recommended numbers of training iterations. The training times correspond to the checkpoints for which we provide FID metrics in Table \ref{table-fid-ave-celeba}. As we can see, while the training times are not very different, at the inference our approach is more competitive.

\begin{figure*}[!t]
     \centering
    \begin{subfigure}[b]{1.0\linewidth}
    \centering
    \hspace*{-18mm}\includegraphics[width=1.2\linewidth]{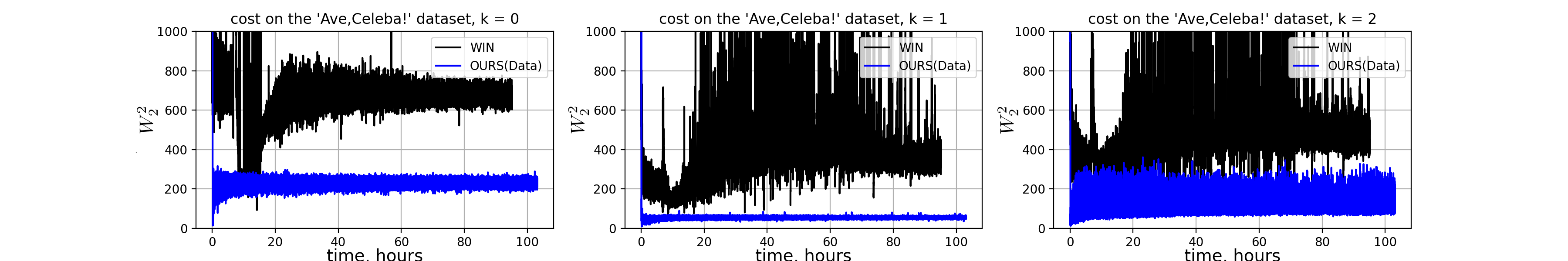}
     \end{subfigure}
 \vspace*{-9.6mm}\newline
     \caption{Transport cost w.r.t training time; our method (classical cost) vs. WIN; Data space.}
     \label{fig:winVSours-training}
     \vspace{-0mm}
 \end{figure*}

\textbf{Comparison of our method and WIN \cite{korotin2022wasserstein}.} We provide the detailed comparison of the training curves of our (bi-level) approach vs. WIN baseline in the data space setup, see Figure \ref{fig:winVSours-training}. Note that WIN represents the most promising tri-level OT barycenter solver. In the chart we demonstrate the evolution of the transport cost during the training. We emphasize that the \textbf{curve for WIN is much less stable} (we even trim the plots from the above) which is natural due to the reliance of WIN on tri-level adversariality.

\vspace{-2mm}
\subsection{Detailed comparison of Our method and EgBary \cite{kolesov2023energy}}\label{app:celeba-ext:our-vs_egbary}

In our experiments with Ave, celeba! Benchmark dataset (\S \ref{subsec:ave_celeba}) the competitive method, EgBary \cite{kolesov2023energy}, demonstrates better FID scores compared to ours. In what follows, we provide a detailed comparison of our approach with this baseline. As for the setup, we choose Ave Celeba! experiment (\S \ref{subsec:ave_celeba}). For the purpose of completeness and clearness of our comparison, we additionally train EgBary in the data space. For EgBary, we use the same Neural Network architectures as we do for our approach (discriminators $f_{k, \theta}$), see our Table \ref{table:hyperparams}. The hyperparameters for their method are chosen for quality reasons and following the guidelines from their paper: $\epsilon = 10^{-2}$, $lr = 10^{-4}$, $iter = 5000$, $\sqrt{\eta} = 0.1$, $L = 500$, $S = 64$.

\textbf{Comparison at inference.} In the provided tables below we demonstrate the inference times, achieved FID and $\mathcal{L}_2$-$\text{UVP}$ for
\vspace{-2mm}
\begin{itemize}[leftmargin=*]
    \item Our approach (classical cost) and EgBary in data space (Table \ref{tbl-egbaryVSours-data});
    \item Our approach with different costs (classical, $\epsilon$-KL and $\gamma$-Energy) and EgBary in latent StyleGAN space (Table \ref{tbl-egbaryVSours-latent}).
\end{itemize}

\begin{table}[h]
\centering
\begin{tabular}{|l|llp{1cm}|llp{1cm}|p{1.3cm}|p{1cm}|}
\hline
\multirow{2}{*}{Method} & \multicolumn{3}{|l|}{ FID}                                     & \multicolumn{3}{|l|}{   L2-UVP, \%}                               & \multirow{2}{*}{ \small \makecell{Langevin \\ steps}} & \multirow{2}{*}{ t, sec} \\ \cline{2-7}
                        & \multicolumn{1}{|p{1cm}|}{\bf{k=0}}  & \multicolumn{1}{p{1cm}|}{\bf{k=1}}  & \bf{k=2}  & \multicolumn{1}{p{1cm}|}{\bf{k=0}} & \multicolumn{1}{p{1cm}|}{\bf{k=1}} & \bf{k=2} &                    &                    \\ \hline
\multirow{5}{*}{EgBary} & \multicolumn{1}{|l|}{15.8} & \multicolumn{1}{l|}{15.3} & 18.3 & \multicolumn{1}{l|}{46}  & \multicolumn{1}{l|}{45}  & 48  & 50                 & 15                 \\
                        &\multicolumn{1}{|l|}{11.3} & \multicolumn{1}{l|}{11.2} & 14.3 & \multicolumn{1}{l|}{37}  & \multicolumn{1}{l|}{36}  & 40  & 150                & 45                 \\ 
                        &\multicolumn{1}{|l|}{8.4}  & \multicolumn{1}{l|}{8.7}  & 10.2 & \multicolumn{1}{l|}{35}  & \multicolumn{1}{l|}{33}  & 37  & 250                & 75                 \\ 
                        &\multicolumn{1}{|l|}{8.3}  & \multicolumn{1}{l|}{8.2}  & 9.9  & \multicolumn{1}{l|}{32}  & \multicolumn{1}{l|}{32}  & 34  & 500                & 150                \\
                        &\multicolumn{1}{|l|}{8.2}  & \multicolumn{1}{l|}{8.2}  & 9.8  & \multicolumn{1}{l|}{32}  & \multicolumn{1}{l|}{31}  & 33  & 1000               & 300                \\  \hline
OURS                    & \multicolumn{1}{l|}{30.7}    & \multicolumn{1}{l|}{31.0}    & \multicolumn{1}{l|}{31.7}   & \multicolumn{1}{l|}{21}       & \multicolumn{1}{l|}{21}       & \multicolumn{1}{l|}{26} & \parbox[t]{2mm}{\multirow{3}{*}{\hspace*{-1.2mm}\makecell{ Not \\ Applicable}}}      & \multicolumn{1}{l|}{0.05}                \\ 
OURS($\epsilon$)       & \multicolumn{1}{l|}{34.5}    & \multicolumn{1}{l|}{34.9}    & \multicolumn{1}{l|}{35.7}   & \multicolumn{1}{l|}{22}       & \multicolumn{1}{l|}{21}       & \multicolumn{1}{l|}{21}  &    & \multicolumn{1}{l|}{0.05}                \\
OURS($\gamma$)         & \multicolumn{1}{l|}{38.3}    & \multicolumn{1}{l|}{37.8}    & \multicolumn{1}{l|}{37.6}   & \multicolumn{1}{l|}{22}       & \multicolumn{1}{l|}{22}       & \multicolumn{1}{l|}{23} &     & \multicolumn{1}{l|}{0.08}        \\ \hline
\end{tabular}
\caption{OUR method vs. EgBary; Ave Celeba!; Latent space.}\label{tbl-egbaryVSours-latent}
\end{table}

\begin{table}[h]
\centering
\begin{tabular}{|l|llp{1cm}|llp{1cm}|p{1.3cm}|p{1cm}|}
\hline
\multirow{2}{*}{Method} &\multicolumn{3}{|l|}{FID}                                        & \multicolumn{3}{l|}{L2-UVP,\%}                               & \multirow{2}{*}{ \small \makecell{Langevin \\ steps}} & \multirow{2}{*}{t, sec} \\ \cline{2-7}
                        &\multicolumn{1}{|p{1cm}|}{\bf{k=0}}   & \multicolumn{1}{p{1cm}|}{\bf{k=1}}   & \bf{k=2}   & \multicolumn{1}{p{1cm}|}{\bf{k=0}} & \multicolumn{1}{p{1cm}|}{\bf{k=1}} & \bf{k=2} &                    &                    \\ \hline
\multirow{5}{*}{EgBary} & \multicolumn{1}{|l|}{125.3} & \multicolumn{1}{l|}{122.3} & 187.6 & \multicolumn{1}{l|}{53}  & \multicolumn{1}{l|}{39}  & 46  & 10                 & 7                 \\ 
                        &\multicolumn{1}{|l|}{119.4} & \multicolumn{1}{l|}{120.0} & 169.7 & \multicolumn{1}{l|}{28}  & \multicolumn{1}{l|}{27}  & 31  & 150                & 105                 \\ 
                        &\multicolumn{1}{|l|}{118.5} & \multicolumn{1}{l|}{120.4} & 168.8 & \multicolumn{1}{l|}{26}  & \multicolumn{1}{l|}{26}  & 31  & 250                & 175                \\ 
                        &\multicolumn{1}{|l|}{118.3} & \multicolumn{1}{l|}{120.1} & 168.9 & \multicolumn{1}{l|}{25}  & \multicolumn{1}{l|}{24}  & 30  & 500                & 350               \\ 
                        &\multicolumn{1}{|l|}{118.8} & \multicolumn{1}{l|}{121.2} & 170.2 & \multicolumn{1}{l|}{25}  & \multicolumn{1}{l|}{24}  & 30  & 1000               & 700               \\ \hline
OURS                    &  \multicolumn{1}{l|}{39.0} & \multicolumn{1}{l|}{38.6}  & \multicolumn{1}{l|}{39.8}  & \multicolumn{1}{l|}{3}       & \multicolumn{1}{l|}{4}       & \multicolumn{1}{l|}{9} & N/A    & 0.01 \\ \hline
\end{tabular}
\caption{OUR method vs. EgBary; Ave Celeba!; Data space.}\label{tbl-egbaryVSours-data}
\end{table}
\vspace{-2mm}
\underline{Some comments}:
\vspace{-2mm}
\begin{enumerate}[leftmargin=*]
\item EgBary relies on the Langevin sampling. In order to provide further details on EgBary performance, in the tables we report the metrics for different numbers of utilized Langevin steps. Fewer steps - the inference is faster, while the quality metrics are not that good and vice versa. 
\item The distribution represented by StyleGAN (learned on CelebA) and the original CelebA distribution are similar but slightly different, which introduces additional biases for the computed metrics. That is why \textit{latent} methods and \textit{data} methods should be compared independently.
\end{enumerate}
\vspace{-3mm}
\underline{Conclusions}. At first, we can see that EgBary method is much more time-consuming at the inference stage. 
This is expected since it uses MCMC sampling. Secondly, in the data space their results (FID) are not that good. In particular, in the data space, our approach demonstrates the best FID. Thirdly, even in the latent space our method demonstrates better results according to the $\mathcal{L}_2$-$\text{UVP}$. At the same time, we acknowledge that in the latent space, their FID score is better than ours.

\begin{figure*}[!t]
     \centering
    \begin{subfigure}[b]{1.0\linewidth} 
          \centering
          \hspace*{-18mm}\includegraphics[width=1.2\linewidth]{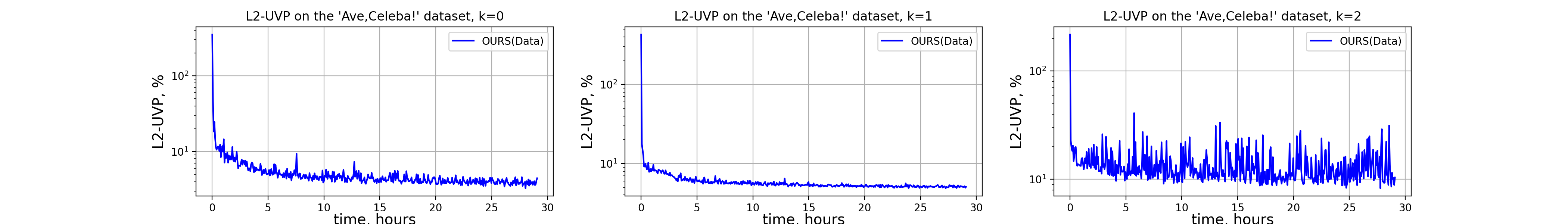}
          \caption{Data space; OURS (classical cost).}
         \label{fig:egbaryVSours-training-data}
     \end{subfigure}
\vspace*{-3mm}\newline
      \begin{subfigure}[b]{1.0\linewidth}  
         \centering
         \hspace*{-18mm}\includegraphics[width=1.2\linewidth]{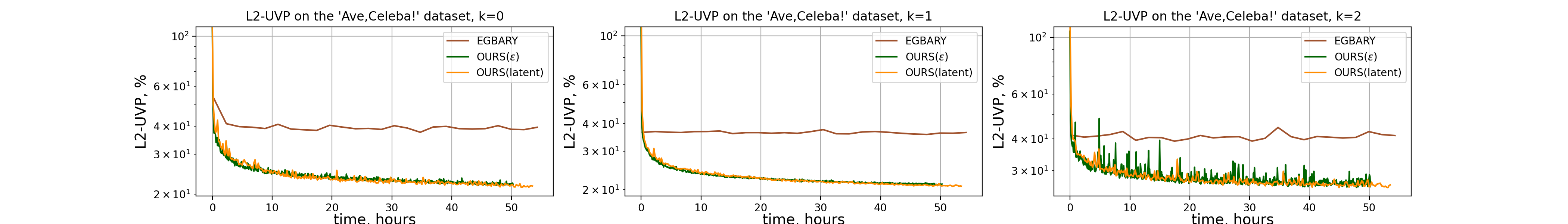}
         \caption{Latent space; OURS (classical, $\epsilon$-KL costs), EgBary.}
          \label{fig:egbaryVSours-training-latent}
      \end{subfigure}
 \vspace*{-9.6mm}\newline
     \caption{Behaviour of $\mathcal{L}_2$-$\text{UVP}$ (w.r.t. to elapsed time) for our approach and EgBary in different setups; Ave Celeba!}
     \label{fig:egbaryVSours-training}
     \vspace{-0mm}
 \end{figure*}

\textbf{Comparison in training.} We provide the behaviour of $\mathcal{L}_2$-$\text{UVP}$ (w.r.t.\ to elapsed time) when training our approach and EgBary in (i) StyleGAN latent space, see Figure \ref{fig:egbaryVSours-training-latent}; (ii) data space, see Figure \ref{fig:egbaryVSours-training-data}. For the latter, we provide training statistics only for our method. 

\begin{figure*}[!t]
     \centering
    \begin{subfigure}[b]{0.3205\linewidth}  
          \centering
          \includegraphics[width=0.995\textwidth]{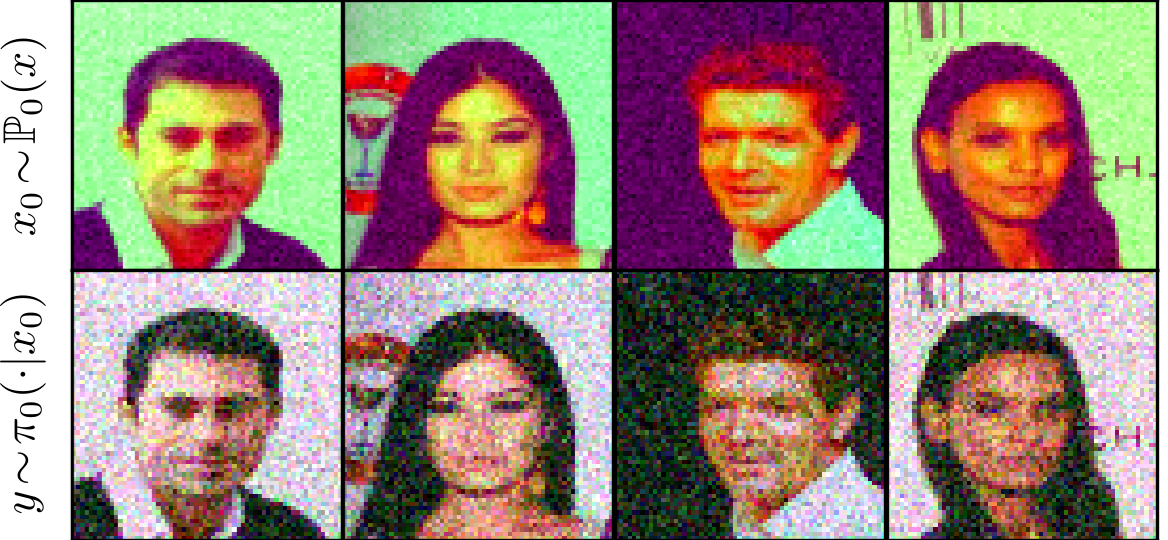}
          \caption{input, transformed samples ($\bbP_0$).}
          \label{fig:egbary-data-celeba-samples-p0}
     \end{subfigure}\hfill
      \begin{subfigure}[b]{0.3205\linewidth}   
         \centering
         \includegraphics[width=0.995\textwidth]{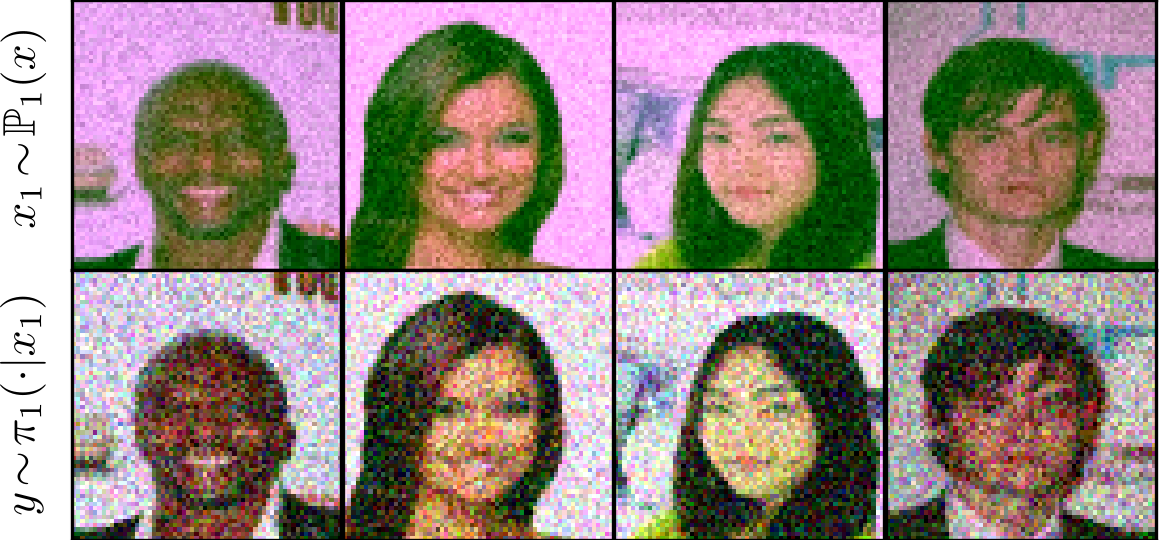}
         \caption{input, transformed samples ($\bbP_1$).}
         \label{fig:egbary-data-celeba-samples-p1}
      \end{subfigure}\hfill
      \begin{subfigure}[b]{0.3205\linewidth} 
         \centering
         \includegraphics[width=0.995\textwidth]{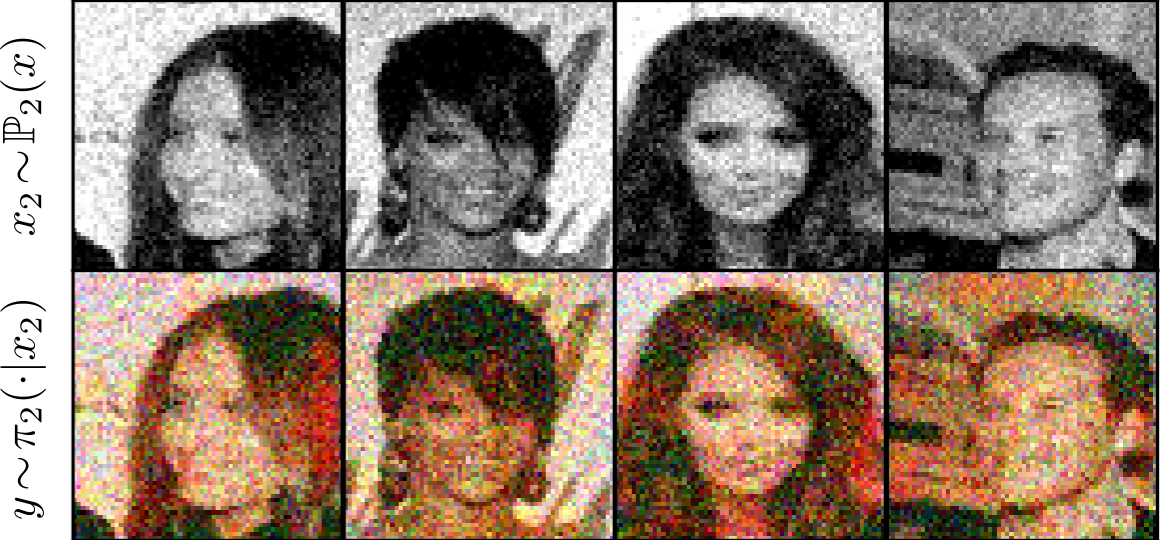}
         \caption{input, transformed samples ($\bbP_2$).}
         \label{fig:egbary-data-celeba-samples-p2}
      \end{subfigure}
 \vspace{-2mm}
     \caption{\centering Samples from (EgBary) learned stochastic OT barycenter plans; Data space; Ave, Celeba! experiment (\S\ref{subsec:ave_celeba}).}
     \label{fig:egbary-data-celeba-samples}
 \vspace{-1mm}
 \end{figure*}

\underline{Conclusions (latent space).} EgBary's latent learning curve achieves its optimum (gets stuck) rather quickly and stops improving. Notably, overall \textbf{our $\mathcal{L}^{2}$-UVP score is always much better than their score}. This is probably due to the fact that Entropy regularization ``blows up'' the barycenter distribution and introduces extra bias. Besides, MCMC sampling adds inconvenient noise to the learning process. In particular, in data space, this is clearly seen from the additional noise on the resulting transported samples, see Figure \ref{fig:egbary-data-celeba-samples}.

\underline{Conclusions (data space).}  We provide qualitative examples of EgBary's data space performance in Figure \ref{fig:egbary-data-celeba-samples}. This EgBary's behaviour is analogous to the one that we have already shown in MNIST 0/1 experiment in the data space (Figure \ref{fig:MNIST-0-1} in the Appendix). In contrast, our method in the data space works incomparably better both in terms of FID ($\sim 30$ vs. $\geq 100$) and $\mathcal{L}^{2}$-UVP ($\sim 5\%$ vs. $\sim 25\%$).

\vspace{-2mm}
\section{Extended related works. Detailed comparison of our method with \citep{chi2023variational}}\label{app:ext-chi-discussion}

Similar to ours, \citep{chi2023variational} utilizes a bi-level optimization objective. To contextualize the novelty of our proposed method, in this section we provide a detailed discussion of this research, compare their approach with ours, and distinguish our contributions compared to their work. 
\begin{enumerate}[leftmargin=5mm]
    \item \underline{Origin of bi-level objective}. While both, our objective and the objective by \citep{chi2023variational}, are bi-level, the origins of these objectives are rather different. Our considered optimization problem \eqref{eq:weakbary_maxmin} stems from the semi-dual formulation of the weak OT problem \eqref{eq:weakot_dual}. This formulation can be used with a diverse set of admissible weak cost functions, in particular, with \textit{unregularized} classical cost function, as we demonstrate in our paper. In turn, the work \citep{chi2023variational} is based on the conventional dual OT problem introduced in \citep{cuturi2013sinkhorn}. One limitation of this formulation is that it deals exclusively with regularized OT problems, i.e., seeks for barycenter w.r.t.\ regularized OT cost functions. Moreover, only two reasonable choices of regularization  is known for this formulation: entropic and quadratic. For some attempts to consider more general regularizations coupled with conventional dual OT, see \citep{blondel2018smooth}.
    \item \underline{What is optimized in the bi-level objective?} The entities that are obtained in the output of our method and the method by \citep{chi2023variational} are different. In our algorithm, we optimize w.r.t.\ to plans $\pi_k$, parameterized as stochastic or deterministic mappings from the reference distributions $\mathbb{P}_k$ to the (implicitly) learned barycenter distribution $\mathbb{Q}$. In contrast, the approach from \citep{chi2023variational} explicitly learns the generative distribution for $\mathbb{Q}$ but \textbf{does not} recover the mappings from reference distributions to the recovered barycenter. At the same time, the knowledge of these mappings is important in some applications of barycenter problem, see, e.g., our Shape-Color experiment, \S \ref{subsec:colorshape}.
    \item \underline{Is it true that the objective by \citep{chi2023variational} is bi-level?} One of the ``levels'' in the bi-level objective from \citep{chi2023variational} is double $\sup_{\phi} \sup_{\psi}$ optimization w.r.t.\ pair of dual OT potentials. While we acknowledge that no adversariality appears here, we note that typically such $\sup$-$\sup$ problems are optimized via an \textit{alternating} procedure, see \citep{seguy2018large, daniels2021score}.
    \item \underline{Barycenter distribution parameterization in \citep{chi2023variational}}. When (explicitly) parameterizing the barycenter distribution, the authors of \citep{chi2023variational} propose the use of Gaussians or Mixtures of Gaussians (MoG) with learnable parameters, see their ``Introducing a Variational Distribution'' section. While MoG is known to satisfy some universal approximation properties, the performance of such models in high dimensions is questionable.
    \item \underline{Practical considerations}. The authors of \citep{chi2023variational} \textbf{only} demonstrate the applicability of their approach in low dimensions ($< 9$) and in moderate dimensions ($D = 128, 256$) for \textbf{Gaussian case}. We are not sure if their approach scales to high dimensions, e.g., to the image domain. In contrast, for our approach, we demonstrate its scalability to the image domain. Furthermore, our proposed procedure is based on well-established architectural, technical and numerical practices from existing research, e.g., \citep{korotin2023neural, choi2023generative}, proven to scale well with dimensions.
\end{enumerate}

\end{document}